\documentclass{article}
\usepackage{arxiv}

\usepackage[utf8]{inputenc} 
\usepackage[T1]{fontenc}    
\usepackage{hyperref}       
\usepackage{url}            
\usepackage{booktabs}       
\usepackage{amsfonts}       
\usepackage{nicefrac}       
\usepackage{microtype}      
\usepackage{lipsum}		
\usepackage{graphicx}
\usepackage{natbib}
\usepackage{doi}
\usepackage{xcolor}
\usepackage{amsmath}
\usepackage{algorithm}
\usepackage{algpseudocode}
\usepackage{cleveref}       
\usepackage{subfigure}

\title{Improving Adversarial Robustness for Free with Snapshot Ensemble}

\author{ {\hspace{0mm}Yihao Wang} \\
	Department of Statistics\\
	University of Illinois at Urbana-Champaign\\
	Champaign, IL 61820 \\
	\texttt{yihaow2@illinois.edu}
}

\hypersetup{
pdftitle={Improving Adversarial Robustness for Free with Snapshot Ensemble}
}

\begin{document}
\maketitle

\begin{abstract}
	Adversarial training, as one of the few certified defenses against adversarial attacks, can be quite complicated and time-consuming, while the results might not be robust enough. To address the issue of lack of robustness, ensemble methods were proposed, aiming to get the final output by weighting the selected results from repeatedly trained processes. It is proved to be very useful in achieving robust and accurate results, but the computational and memory costs are even higher. Snapshot ensemble, a new ensemble method that combines several local minima in a single training process to make the final prediction, was proposed recently, which reduces the time spent on training multiple networks and the memory to store the results. Based on the snapshot ensemble, we present a new method that is easier to implement: unlike original snapshot ensemble that seeks for local minima, our snapshot ensemble focuses on the last few iterations of a training and stores the sets of parameters from them. Our algorithm is much simpler but the results are no less accurate than the original ones: based on different hyperparameters and datasets, our snapshot ensemble has shown a 5\% to 30\% increase in accuracy when compared to the traditional adversarial training.
\end{abstract}

\section{Introduction}
Deep learning has demonstrated considerable problem-solving abilities in broad computer engineering and visual analysis tasks since its emergence \citep{hinton2006fast}, accompanied by long-term discussions of model performances of clean examples in various examinations and experiments. In recent years, however, discoveries on adversarial examples have lead to further evaluations on models' vulnerable adversarial robustness and resulting security issues \citep{szegedy2013intriguing, yuan2019adversarial, goodfellow2014explaining, meng2017magnet, ma2018characterizing}. What often happens to adversarial examples is that when images are slightly perturbed to fool classifiers, outputs will show a large fluctuation, further indicating the lack of robustness of models \citep{goodfellow2014explaining}.

Usually two methods are applied to process adversarial perturbations: the \textit{correct classification} \citep{biggio2010multiple, guo2017calibration, moosavi2017universal, sankaranarayanan2018regularizing} and the \textit{selection of unperturbed images} \citep{meng2017magnet, xu2017feature, athalye2018obfuscated}. The former relies on robust classifiers to make correct predictions on all the images, benign or adversarial, while the latter detects and filters the unperturbed images from adversarial examples, only on which it predicts. In particular, as the backbone of the correct classification, the \textit{adversarial training} has been proved effective in numerous deep learning models \citep{goodfellow2014explaining, kurakin2016adversarial, huang2015learning, shrivastava2017learning, tramer2017ensemble}, along with its apparently unstable, expensive, and time-consuming drawbacks \citep{madry2017towards, sun2018domain}. For stability issue, since slightly perturbed images can result in significant fluctuations in the outputs, there exist large variances within and between trials. Furthermore, for the computational complexity, the running time is usually three to thirty times that of the non-robust networks \citep{shafahi2019adversarial}. This is mainly due to the time-consuming generation of adversarial examples which alone requires an optimization procedure, e.g. via fast gradient sign method (FGSM) \citep{goodfellow2014explaining}, projected gradient descent (PGD) \citep{goodfellow2014explaining, madry2017towards, kurakin2016adversarial}, One Pixel attack \citep{su2019one}, CW attack \citep{carlini2016towards}, or DeepFool \citep{moosavi2015deepfool}.

In the meanwhile, due to such defects of adversarial training, alternative methods have been proposed to improve its performances or to replace it. The Adversarial Robustness based Adaptive Label Smoothing (AR-AdaLS) proposed by \cite{qin2020improving} aims to improve the smoothness of adversarial robustness in order to solve the instability: by training the model to distinguish the training data of varied adversarial robustness and by giving different supervision to the training data, their methods promotes label smoothing \citep{szegedy2013intriguing} and leads to better calibration and stability. \cite{jakubovitz2018improving} and \cite{hoffman2019robust} studied Jacobian regularization to regularize the training loss after the regular training, aiming to provide another way of enhancing robustness other than adversarial training. \cite{mopuri2018nag}, inspired by the architecture of GANs, attempted to capture the distribution of adversarial perturbation; their method exhibited extraordinary fooling rates, variety, and cross model generalizability.

There has been very few certified defenses \citep{lecuyer2019certified, goodfellow2014explaining, cohen2019certified, weng2018towards, wong2018provable} that are both applicable and functional in large scale problems such as ImageSet in recent years however, making adversarial training still remain the most trusted defense \citep{shafahi2019adversarial}. Yet its troublesome features and the lack of outstanding performances on complex datasets make general tasks time-consuming, and large tasks almost unprocessable under ordinary circumstances, thus urging the need of finding other fast and for free methods that would raise adversarial robustness to the same level. 

\subsection*{Contributions}
Instead of improving performances of adversarial training, we study a training algorithm in which images are not perturbed in the dataset (either randomly as in certified robustness or adversarially as in adversarial training). The algorithm, while keeping the time spent on robust training almost equal to the non-robust ones, also produces robust models. Instead of perturbing the dataset, we apply the \textit{Snapshot Ensemble} \citep{huang2017snapshot, smith2017cyclical, loshchilov2016sgdr} along the training process to store multiple historical weights so as to defend against adversarial attacks such as FGSM or PGD, in a \textit{Bayesian Neural Network} \citep{blundell2015weight, kingma2015variational, ru2019bayesopt,zhang2021differentially} manner. The proposed method produces results as fast as the non-robust networks, with only a few seconds difference when trained on CIFAR-10 \citep{krizhevsky2009learning} and MNIST \citep{lecun1998mnist} datasets. The accuracy after attacks (i.e. the \textit{robust accuracy}) with the ensemble during training is shown to be 5\% to 30\% better than the accuracy with regular training, depending on the dataset and \textit{perturbation magnitude} $\epsilon$.

\section{Background Knowledge}
\subsection*{Adversarial Examples \& Robustness} As previously introduced, the adversarial examples are the perturbed images or samples to fool the classifier in order to generate inaccurate outputs. Usually, the adversarial examples are defined within a range or \textit{perturbation set}, for example, $l_2$ or $l_\infty$ spaces with a radius $\epsilon$. Here $\epsilon$ is also known as the perturbation magnitude. Therefore the adversarial example is
\begin{align}
x+\delta^*
\text{ where }\delta^*=\text{argmax}_{\delta}(\ell(f_\theta(x+\delta),y) \text{ s.t. } \|\delta\|_p\leq \epsilon
\label{eq:adv examples}
\end{align}
in which $\ell$ is the loss function, $f$ is the neural network governed by its parameters $\theta$, $x$ is the input sample, $y$ is the corresponding label and $\delta$ is the perturbation.

From the viewpoint of adversarial examples, we say a model is \textit{adversarially robust} if a small perturbation $\delta$ does not change the output:
\begin{equation}
f_\theta(x+\delta) = f_\theta(x)
\end{equation}
Numerous speculative hypotheses such as insufficient model averaging and regularizing had been proposed to address the causes of the existence of adversarial examples and model vulnerability before \cite{goodfellow2014explaining} found the determinate factor of the local linearity property of deep neural networks that even when non-linear activation functions are used, most of the training are manually operated in linear regions \citep{pascanu2013difficulty}.

\subsection*{Adversarial Attack } Adversarial attack is a growing threat in the world of machine learning and deep learning, especially in applied fields such as computer vision and natural language processing. Based on the knowledge and goals of attackers, white-box and black-box attacks \citep{nidhra2012black, beizer1995black} are commonly implemented to fulfill targeted \citep{szegedy2013intriguing} or non-targeted goals \citep{moosavi2017universal, athalye2018synthesizing}. The aim of attackers is to insert a least amount of perturbation to the input to obtain desired misclassification \citep{huang2017adversarial}. An adversarial attack is, therefore, an optimization procedure to solve the constrained maximization problem \eqref{eq:adv examples}.


Perturbation sets are what determine the constraints of the maximization problem \eqref{eq:adv examples}, e.g. $l_p$ space such that:
\begin{equation}
    \left \| \delta \right \|_p = \sqrt[p]{\sum_{i=1}^{k} \left | \delta_i\right | ^p} \leq \epsilon
\end{equation}
Among the $l_p$ spaces, $l_0$ (the number of non-zero elements in the vector), $l_2$, and $l_\infty$ are three commonly used norms in adversarial attacks, i.e.
\begin{equation}
    \left \| \delta \right \|_0 = \#\left (i\mid \delta_i \ne 0 \right )\leq \epsilon,\quad     \left \| \delta \right \|_2 = \sqrt{\sum_{i=1}^{k} \left | \delta_i\right | ^2}\leq \epsilon,\quad
    \left \| \delta \right \|_\infty =  \max_{i}\left ( \left | \delta_i\right |\right )\leq \epsilon
\end{equation}



To $\textup{maximize}_{\left \| \delta \right \|_p \le \epsilon} \ell(f_{\theta}(x+\delta), y)
$, one usually performs a standard gradient ascent method over $\delta$, followed by a projection to ensure $\|\delta\|_p\leq \epsilon$. Many adversarial attack methods have been proposed. Here, we list and experiment on some prevalent attack methods: 

\begin{itemize}
    \item $l_0$ attack: OnePixel \citep{su2019one}, SparseFool \citep{modas2019sparsefool}
    \item $l_2$ attack: Projected Gradient Descent-$l_2$ (PGDL2) \citep{goodfellow2014explaining, madry2017towards}, DeepFool \citep{moosavi2015deepfool}, CW attack \citep{carlini2016towards}, AutoAttack-$l_2$ \citep{wong2020fast}
    \item $l_\infty$ attack: Fast Gradient Sign Method (FGSM) \citep{goodfellow2014explaining}, Projected Gradient Descent (PGD) \citep{goodfellow2014explaining, madry2017towards}, AutoAttack-$l_\infty$ \citep{wong2020fast}
\end{itemize}

\subsection*{FGSM } As one of the earliest and most popular adversarial attacks described by \cite{goodfellow2014explaining}, Fast Gradient Sign Method (FGSM) serves as a baseline attack in our training. As notified previously, to optimize the parameter in trained models is to maximize the loss function over $\delta$. We can increase the loss by moving constantly in the direction of the gradient by some step size \citep{goodfellow2014explaining, wiyatno2019adversarial}, but because FGSM is a $l_\infty$ attack, the magnitude of gradients is restricted within the square threshold of $l_\infty$, so the direction is the only thing we need to care about. For gradients that exceed the threshold, we clip them back to the exact extrema of the perturbation set $\left[-\epsilon, \epsilon\right]$; for those within the threshold, because there are only two directions $+$ or $-$, we only need to adjust them also to the boundaries $\pm \epsilon$ accordingly to maximize the loss. Simply speaking, by taking the sign of the gradient, we obtain the optimal max-norm perturbation to be either $-\epsilon$ or $\epsilon$
\begin{equation}
	\delta^* = \epsilon\cdot\mathrm{sign}(\nabla_\delta \ell(f_{\theta}(x+\delta), y))
\end{equation}

\subsection*{PGD } Projected Gradient Descent (PGD) is the more careful iteration of updates of FGSM with smaller step sizes
\begin{equation}
	\delta_{t+1} = \mathcal{P}(\delta_t+\alpha\nabla_{\delta_t} \ell(f_{\theta}(x+\delta_t), y))
\end{equation}

where $\mathcal{P}$ stands for the projection back to the $l_\infty$ norm (clipping or truncating within range $\left[-\epsilon, \epsilon\right]$) \citep{wiyatno2019adversarial, madry2017towards}. Note that PGD also has the form of $l_2$ version and it is the workhorse in adversarial attack nowadays.

\subsection*{Adversarial Training}

Normally when training a non-robust classifier we want to optimize the parameter $\theta$ by minimizing the average loss
\begin{align}\mathop{\textup{minimize}}\limits_{\theta}\frac{1}{n} \sum_{i=1}^{n} \ell(f_{\theta}(x_{i}), y_{i}) 
\label{eq:loss min}
\end{align}
where \{$x_{i} \in \mathcal{X}, y_{i} \in \mathcal{Y}\}, i = 1,..., n$, and $f_{\theta}$ the function that $\mathcal{X} \to \mathcal{Y}$. In adversarial training, we instead want to optimize $\theta$ by the minimax problem:
\begin{align}
    \mathop{\textup{minimize}}\limits_{\theta}\frac{1}{n} \sum_{i=1}^{n} \mathop{\textup{maximize}}\limits_{\delta: \|\delta\|_p\leq \ \epsilon}\ell(f_{\theta}(x_{i}+\delta), y_{i})
\end{align}
Algorithically speaking, at each iteration, one needs to solve the inner constrained maximization, e.g. by PGD or FGSM, and then the outer minimizaiton, e.g. by standard SGD or Adam.

\begin{algorithm}
\caption{Adversarial Training with FGSM}\label{ADV Train}
\begin{algorithmic}[1]
\Statex \textbf{Input:} Training Sample $\mathcal{X}$; Neural Network $f_\theta$; Perturbation bound $\epsilon$; Learning Rate $\eta$; Total Epoch $T$; 
\Statex \textbf{Process:} 
\State Initialize $\theta$ randomly and set $\delta \gets 0$
\For{epoch = $1, \cdots, T$}
    \For{minibatch $B\subset \mathcal{X}$}
        \State $g_{adv} = \nabla_\delta \ell(f_{\theta}(x+\delta), y)$
        \State Maximization: Update $\delta$ by $\delta + \epsilon\cdot$sign($g_{adv}$) and then clip to $\left[-\epsilon, \epsilon\right]$
        \State $g_\theta = \nabla_\theta \ell(f_{\theta}(x+\delta), y)$ where x, y $\in$ B
        \State Minimization: Update $\theta$ by $\theta-\eta\cdot g_\theta$
    \EndFor
\EndFor
\end{algorithmic}
\end{algorithm}

\subsection*{Snapshot Ensemble}

Regular training, or as \cite{gawlikowski2021survey} refer to as \textit{deterministic methods}, is an one-to-one process that one input after classification predicts one output, but it is not sufficient to produce accurate and robust results. To improve on this, the ensemble methods and the Bayesian Neural Networks (BNN) were proposed, both at the cost of higher computation and memory cost \citep{zhou2021ensemble}. 
Recently, a new method of ensemble learning called \textit{snapshot ensemble learning} (SEL) was proposed \citep{huang2017snapshot}. Unlike Stochastic Gradient Descent (SGD), which avoids saddle points and local minima \citep{bottou2010large, dauphin2014identifying}, SEL stores and ensembles the local minima to improve model performances: by dividing the training process into multiple cycles, the small learning rate encourages the model to converge towards the local minima, and then SEL combines these local minima \citep{huang2017snapshot, wen2019new}.

In this work, we propose a new ensemble similar to SEL in spirit, i.e. collecting and ensembling network parameters along the single training procedure. However, our ensemble does not require local minima and treats the parameters at each iteration as a random draw from the limiting distribution of parameters, which is detailed in \Cref{sec: our SEL}.

\begin{itemize}
    \item Regular ensemble: initialize $M$ neural networks with parameters $\theta_1(0),\cdots,\theta_M(0)$ and train separately to get $\theta_1(t),\cdots,\theta_M(t)$ at the $t$-th iteration; the final prediction is $\sum_i f_{\theta_i(t)}(x)$.
    \item Original snapshot ensemble: initialize 1 neural network with $\theta_1(0)$ and use $M$ local minima $\theta_1(t_j)$ where $j\in[M]$ during the training process; the final prediction is $\sum_j f_{\theta_1(t_j)}(x)$.
    \item Our snapshot ensemble: initialize 1 neural network with $\theta_1(0)$ and use the last $M$ iterations in the epoch $\theta_1(t-M+1),\cdots,\theta_1(t)$; the final prediction is $\sum_i f_{\theta_1(t-i+1)}(x)$.
\end{itemize}



\section{Snapshot Ensemble Improves Adversarial Robustness}\label{sec: our SEL}

\begin{algorithm}[!htb]
\caption{Our Snapshot Ensemble}\label{Our SEL}
\begin{algorithmic}[1]
\Statex \textbf{Input:} Dataset $\mathcal{D} = \left \{\left (x_1, y_1 \right), \cdots, \left (x_n, y_n \right) \right \}$; network model $f$ with trainable parameters $\theta_t$; number of iterations $T$; number of copies $M$; optimizer $A$
\For{$t = 1, \cdots, T$}
    \State $D_t$ = sample(Bootstrap) from $\mathcal{D}$ (Batch)
    \State $\theta_{t}=A(D_t,\theta_{t-1},f)$
\EndFor
\Statex \textbf{Output:} $f\left(x\right)$ = $\sum_{k=T-M+1}^{T}{f}\left (x;\theta_k \right)$
\end{algorithmic}
\end{algorithm}

Instead of training multiple networks (like regular ensemble method), both original SEL and ours only need to train a single network, therefore saving the computational complexity by $M$ times. However, both SEL needs to store $M$ sets of historical parameters, hence requiring higher memory cost than the regular ensemble.

We further distinguish our SEL with the traditional SEL. The original SEL seeks multiple local minima via the so-called cosine annealing learning rate \citep{loshchilov2016sgdr, loshchilov2017decoupled} that adjusts the magnitude of learning rate during training. As the epoch increases, cosine annealing learning rate first decreases and then rises rapidly at the end of each cosine cycle to escape from the local minima \citep{huang2017snapshot}. The process is then repeated several times until all the epochs are finished. By recording the local minima, we can use a combination strategy (voting, averaging, second-level learning, etc.) to compute the weighted result that exceeds the original accuracy rate and shows an steady increase at all the recorded spots when compared with the base model \citep{brownlee2018better, huang2017snapshot, izmailov2018averaging, loshchilov2017decoupled}.

Our snapshot ensemble, however, does not concern the cosine annealing and local minima. Unlike the regular snapshot ensemble that uses a complicated process to collect sample points at local minima \citep{loshchilov2016sgdr, loshchilov2017decoupled, huang2017snapshot} or the BNN that regards the weight as a Gaussian distribution with a mean and a variance and optimizes them for each weights \citep{anzai2012pattern, gawlikowski2021survey}, we are looking for a simpler snapshot ensemble method that extracts weights from the last few iterations and compute the weighted output accordingly. Through experiments on different datasets and attack methods, our SEL consistently shows significant better robust accuracy, with 5\% to 30\%increase from the regular accuracy.

\section{Experiments}
Our experiments are implemented with two commonly used datasets: MNIST \citep{lecun1998mnist} and CIFAR-10 \citep{krizhevsky2009learning}. The MNIST set is a 28 $\times$ 28 grey scale dataset with 60000 training and 10000 testing figures of hand written numeric figure. The CIFAR-10 set consists of 6000 examples of 10 classes, with 50000 training figures and 10000 testing figures. Being the 3-channel (RGB) 32 $\times$ 32 pixels color dataset, CIFAR-10 contains objects in real world that have not only a higher level of noises but very diverse features and scales, making the CIFAR-10 set less resistant to adversarial attacks and tougher to reflect the performances of numerous models \citep{pang2019rethinking, yin2019fourier, peck2017lower, sen2020empir}. 

The optimizers I have chosen are Stochastic Gradient Descent (SGD), \textit{Heavy Ball} (HB, i.e. SGD with a momentum), and \textit{Nesterov Accelerated Gradient} (NAG), with more concentration on Heavy Ball. Gradient Descent (GD), as one of the oldest and the most prevalent optimizers, is a method that finds the minimum loss step by step. While GD takes in all the data in each step, SGD stochastically picks a single point and updates for the parameters. Every step of SGD is weaker than GD, but because it also takes less total steps with a faster pace, SGD saves a lot more time. Therefore, SGD has become the most welcomed optimizer for its simplicity and time-saving. 

Heavy ball method is a modified form of gradient descent: in each step of gradient descent, the optimizer generates a momentum vector in the direction of this step and a given magnitude so that in the next step, this momentum is composed with the new gradient descent that nudges the new parameters in the magnitude and direction of the previous step \citep{ghadimi2015global, gadat2018stochastic, sutskever2013importance, saunders2018notes, ruder2016overview, botev2017nesterov}. It is like the movement of a heavy ball in the real world: when the ball's movement direction changes, it will be affected by the previous movement trend and will present a result slightly deviating from the expectation. Unlike SGD that stresses the global minimum and becomes invalid at local extrema or non-convex points, the momentum of heavy ball could bring the steps out from local extrema with no extra time spends, yet there have been less focuses on it since its proposal by Polyak in 1964. Therefore, we have chosen to do our experiments with more concentrations on heavy ball.

NAG is the smarter version of heavy ball, that is, it takes in two successive steps to decide the momentum. Through the intermediate point we can determine from the previous two steps what the momentum in the current step should be like and make a correction accordingly \citep{ruder2016overview, saunders2018notes, lin2019nesterov, botev2017nesterov}. 

The neural network architecture for CIFAR-10 is taken from Pytorch tutorial\footnote{See \url{https://pytorch.org/tutorials/beginner/blitz/cifar10_tutorial.html}}, and that for MNIST is adapted by changing the second hidden layer from 400 to 256 units.

Some default hyperparameters: learning rate = 0.02; momentum = 0.9; snapshot = 20; epoch = 20; epsilons = [0.0, 0.01, 0.02, 0.03, 0.04]; weight decay = 0.0; alpha (PGD) = 0.02; steps (PGD) = 2; random start (PGD) = False. One or more of these hyperparameters are adjusted in different experiments accordingly; other hyperparameters, otherwise notified, are used as default in the experiments.

\subsection{CIFAR-10}

Different colors represent different attack magnitude $\epsilon$ of FGSM. Dashed line is our snapshot ensemble with four different snapshots. Solid line is regular training. Default number of snapshots is 10. Notice that when $\epsilon=0$, the accuracy is indeed the clean accuracy without suffering the attack.

\subsubsection{Outer minimization optimizers}

\begin{figure}[!htb]
\centering
\subfigure[SGD]{
\begin{minipage}[t]{0.333\linewidth}
\centering
\includegraphics[width=2.in]{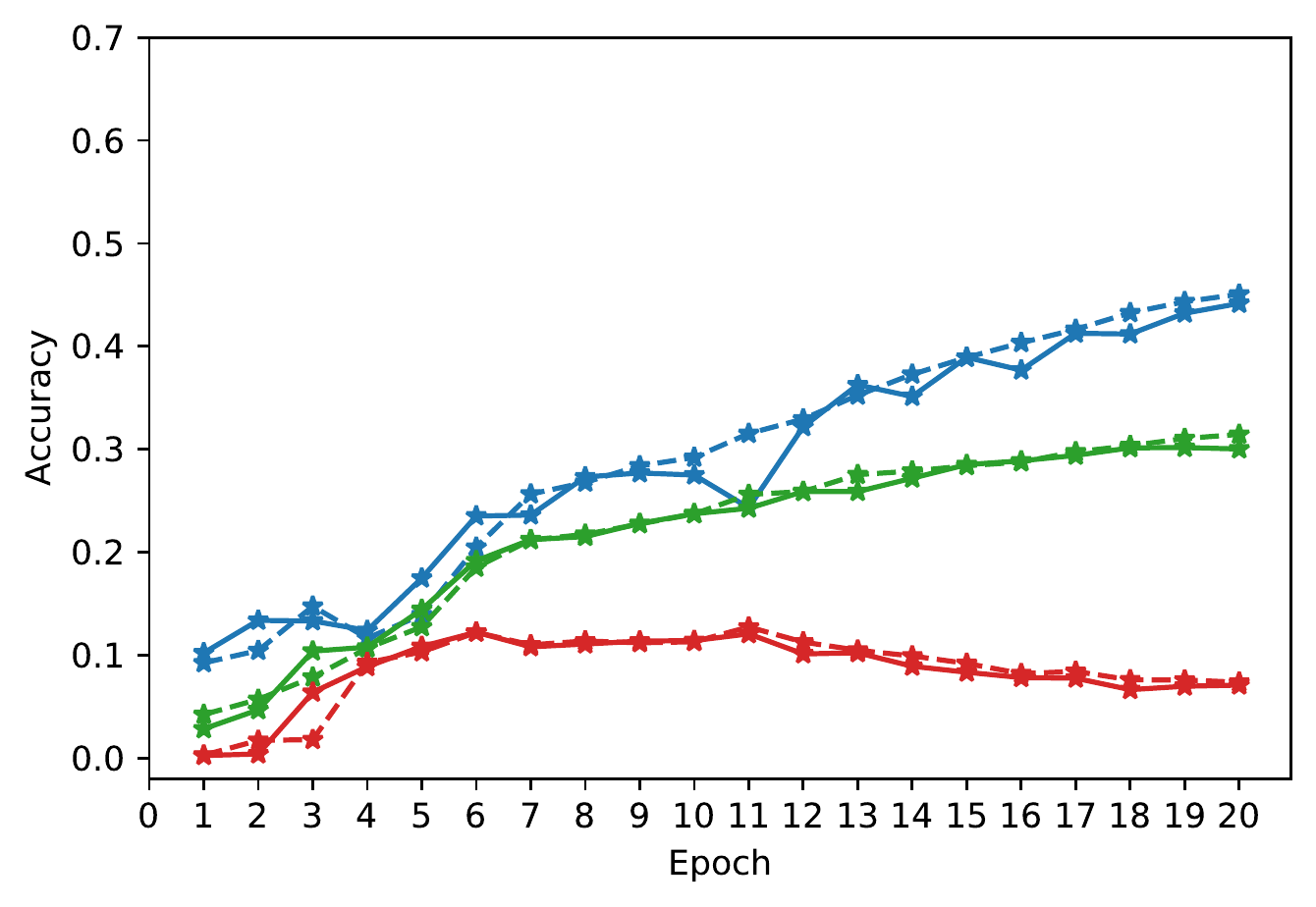}
\end{minipage}%
}%
\hspace{-0.5cm}
\subfigure[NAG]{
\begin{minipage}[t]{0.333\linewidth}
\centering
\includegraphics[width=2.in]{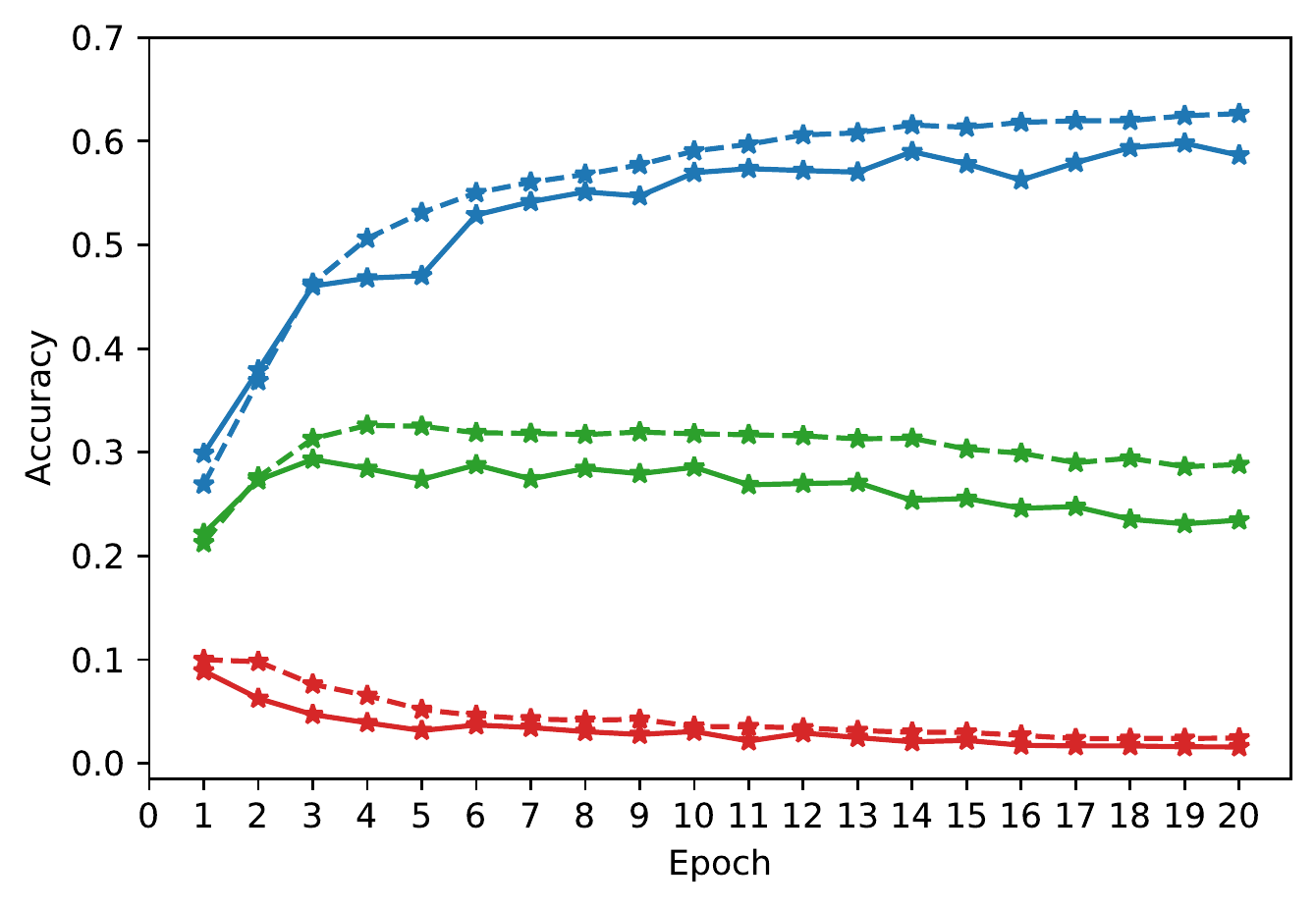}
\end{minipage}%
}%
\hspace{-0.4cm}
\subfigure[HB]{
\begin{minipage}[t]{0.333\linewidth}
\centering
\includegraphics[width=2.3in]{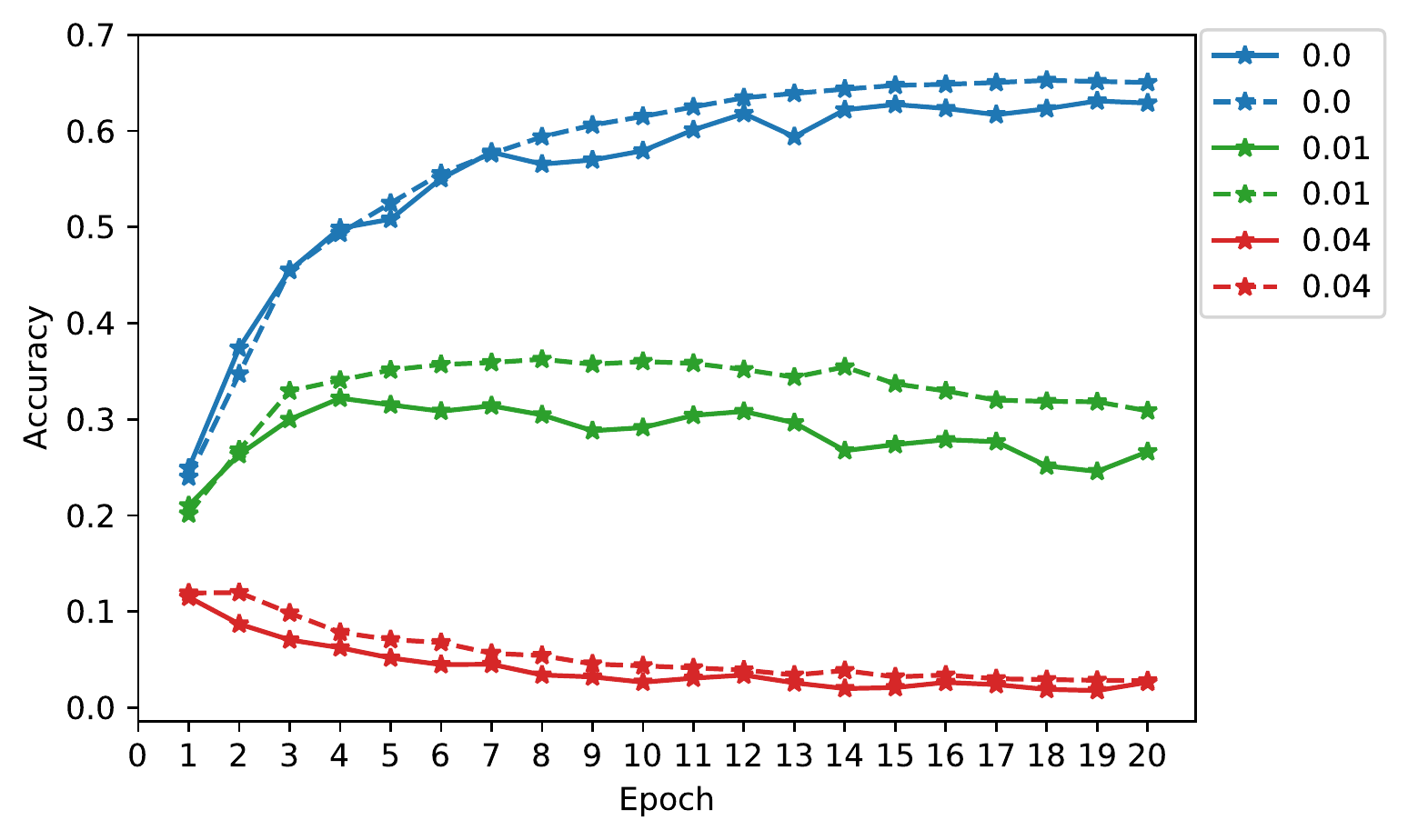}
\end{minipage}
}%
\caption{Adversarial accuracy of neural networks on CIFAR-10 under FGSM, trained with different optimizers.}
\label{fig:CIFAR FGSM optimizer}
\end{figure}

\begin{figure}[!htb]
\centering
\subfigure[SGD]{
\begin{minipage}[t]{0.333\linewidth}
\centering
\includegraphics[width=2.in]{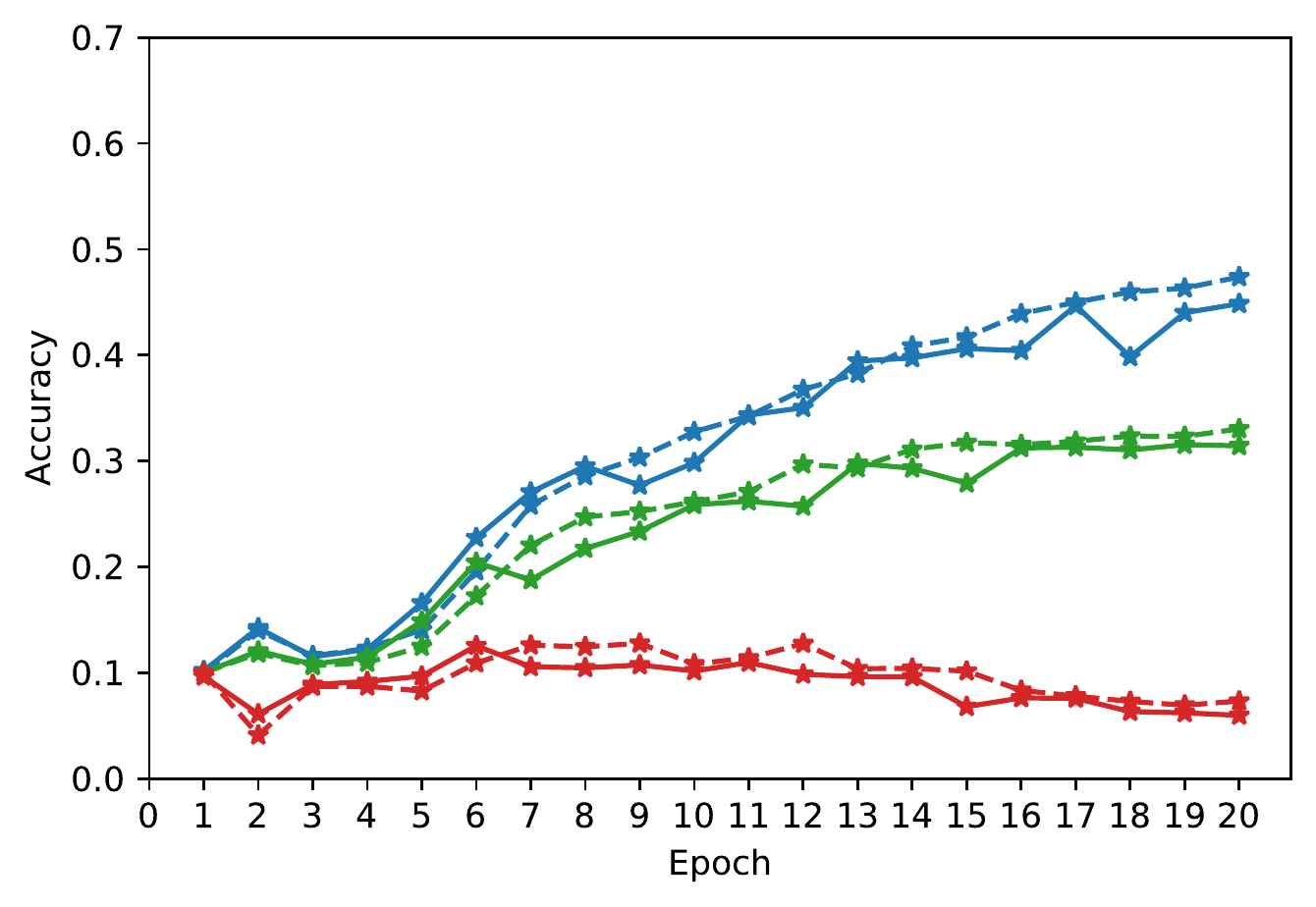}
\end{minipage}%
}%
\hspace{-0.5cm}
\subfigure[NAG]{
\begin{minipage}[t]{0.333\linewidth}
\centering
\includegraphics[width=2.in]{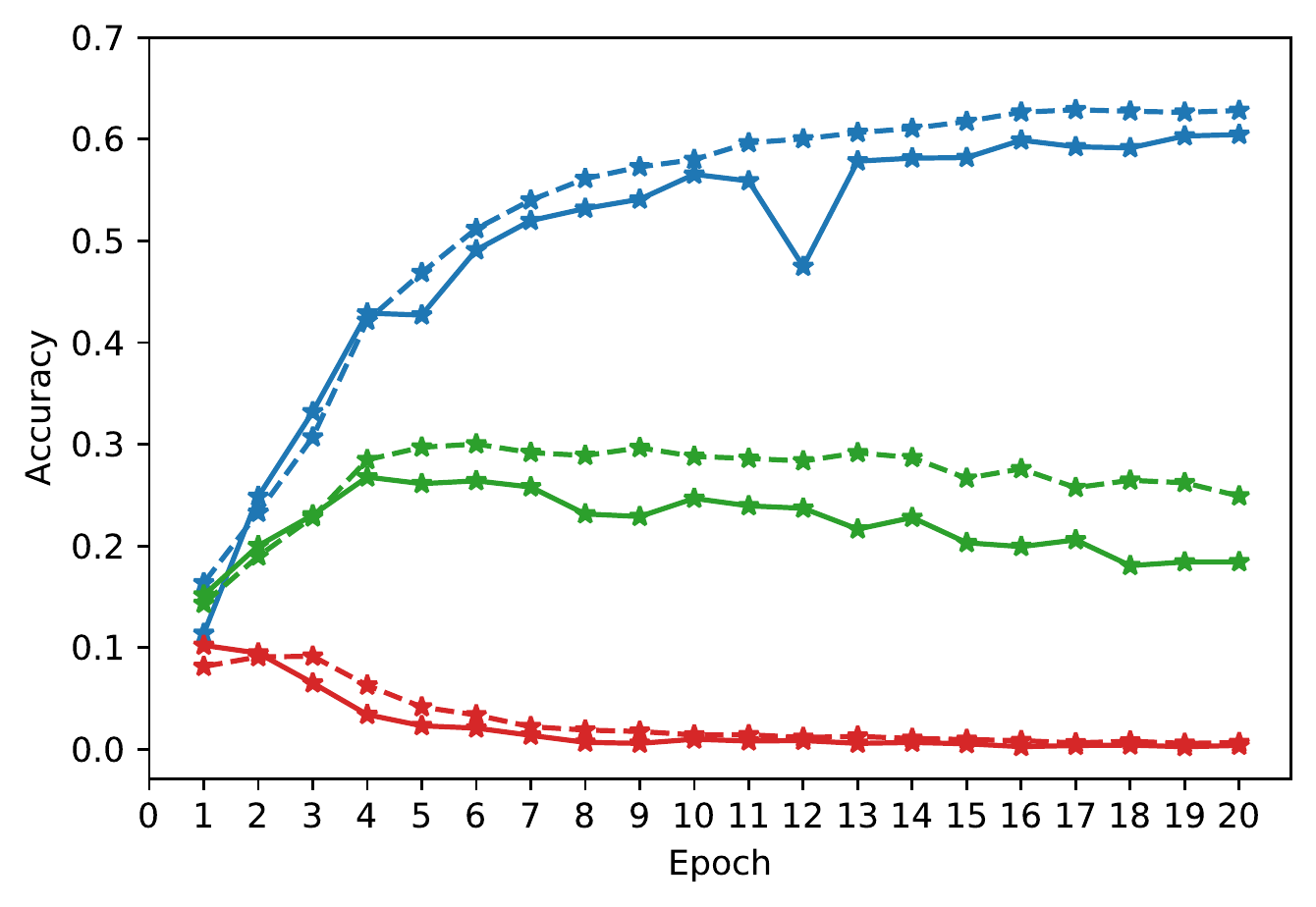}
\end{minipage}%
}%
\hspace{-0.4cm}
\subfigure[HB]{
\begin{minipage}[t]{0.333\linewidth}
\centering
\includegraphics[width=2.3in]{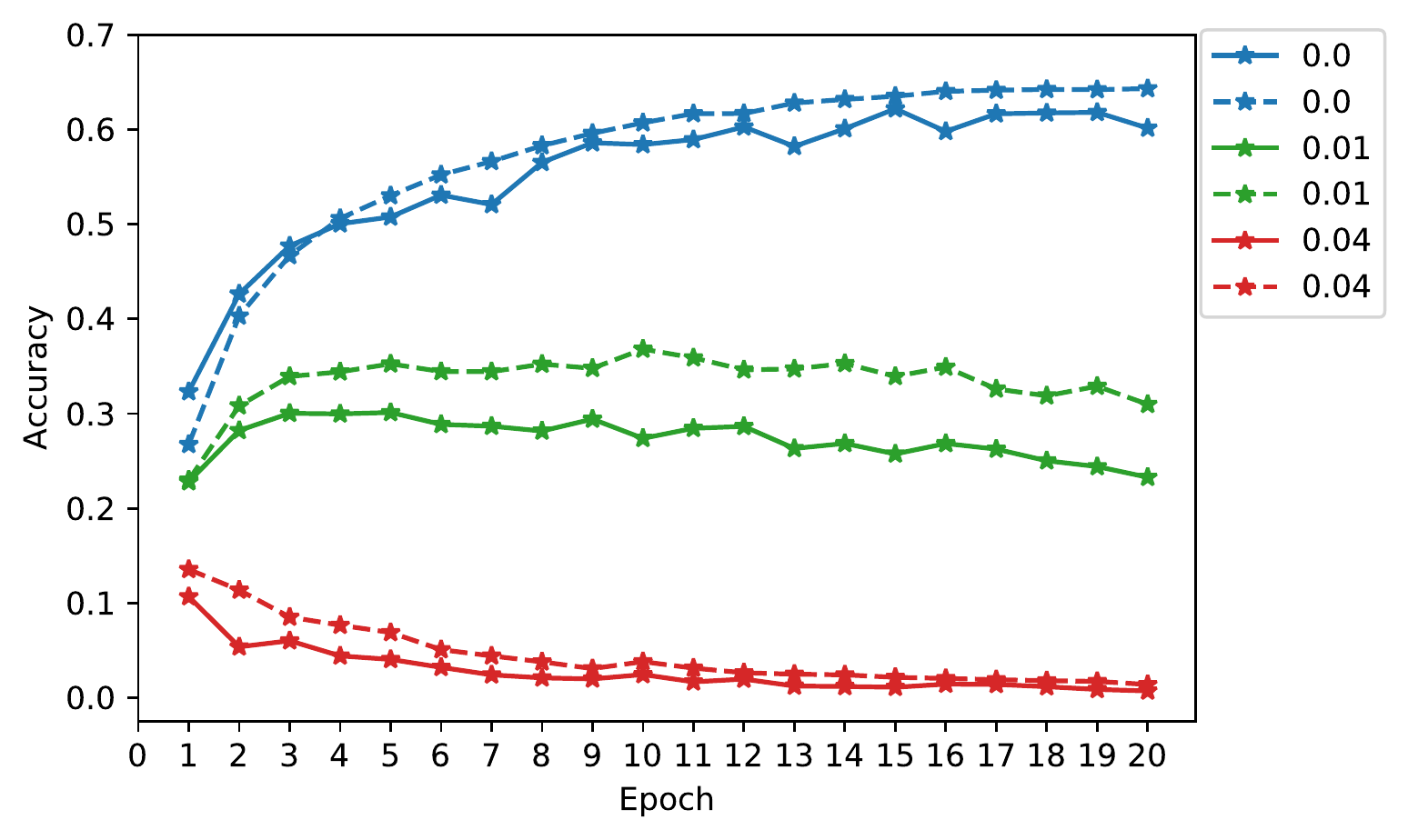}
\end{minipage}
}%
\caption{Adversarial accuracy of neural networks on CIFAR-10 under PGD, trained with different optimizers.}
\label{fig:CIFAR PGD optimizer}
\end{figure}

From \Cref{fig:CIFAR FGSM optimizer} and \Cref{fig:CIFAR PGD optimizer} we observe that, fixing an attack, the outer minimization optimizer does affect on the adversarial robustness of neural networks. For example, when attacked by FGSM or PGD with $\epsilon=0.01$, the neural network learned by SGD has around 30\% accuracy, while momentum methods achieve around 20\%. On the other hand, our snapshot ensemble improves much more on momentum methods (roughly 10\%) than on SGD. In addition, different attacks seem to have small effects on the performance under various $\epsilon$.

\subsubsection{Number of snapshots}

\begin{figure}[!htb]
\centering
\subfigure[98 snapshots]{
\begin{minipage}[t]{0.25\linewidth}
\centering
\includegraphics[width=1.5in]{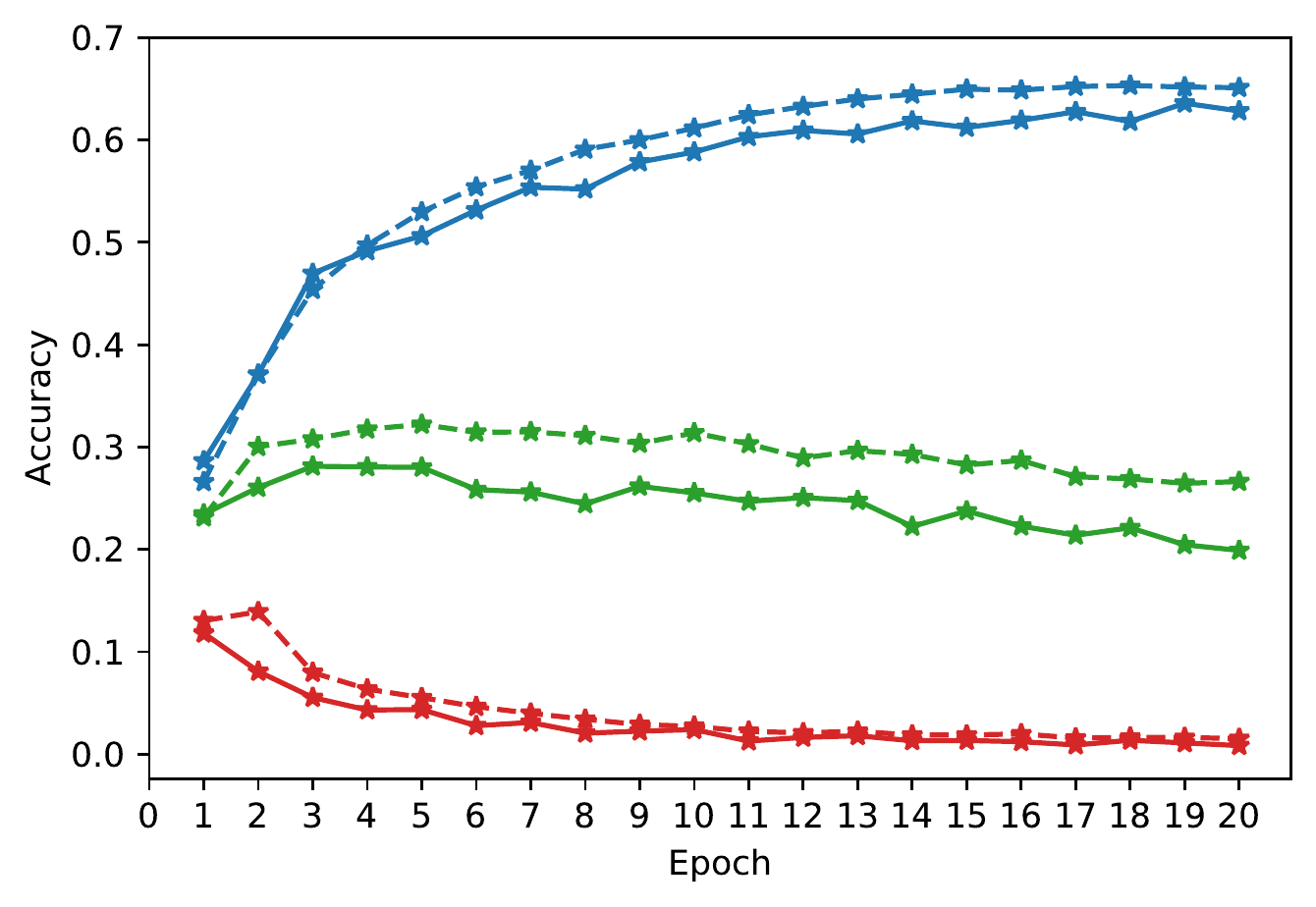}
\end{minipage}
}%
\hspace{-0.5cm}
\subfigure[40 snapshots]{
\begin{minipage}[t]{0.25\linewidth}
\centering
\includegraphics[width=1.5in]{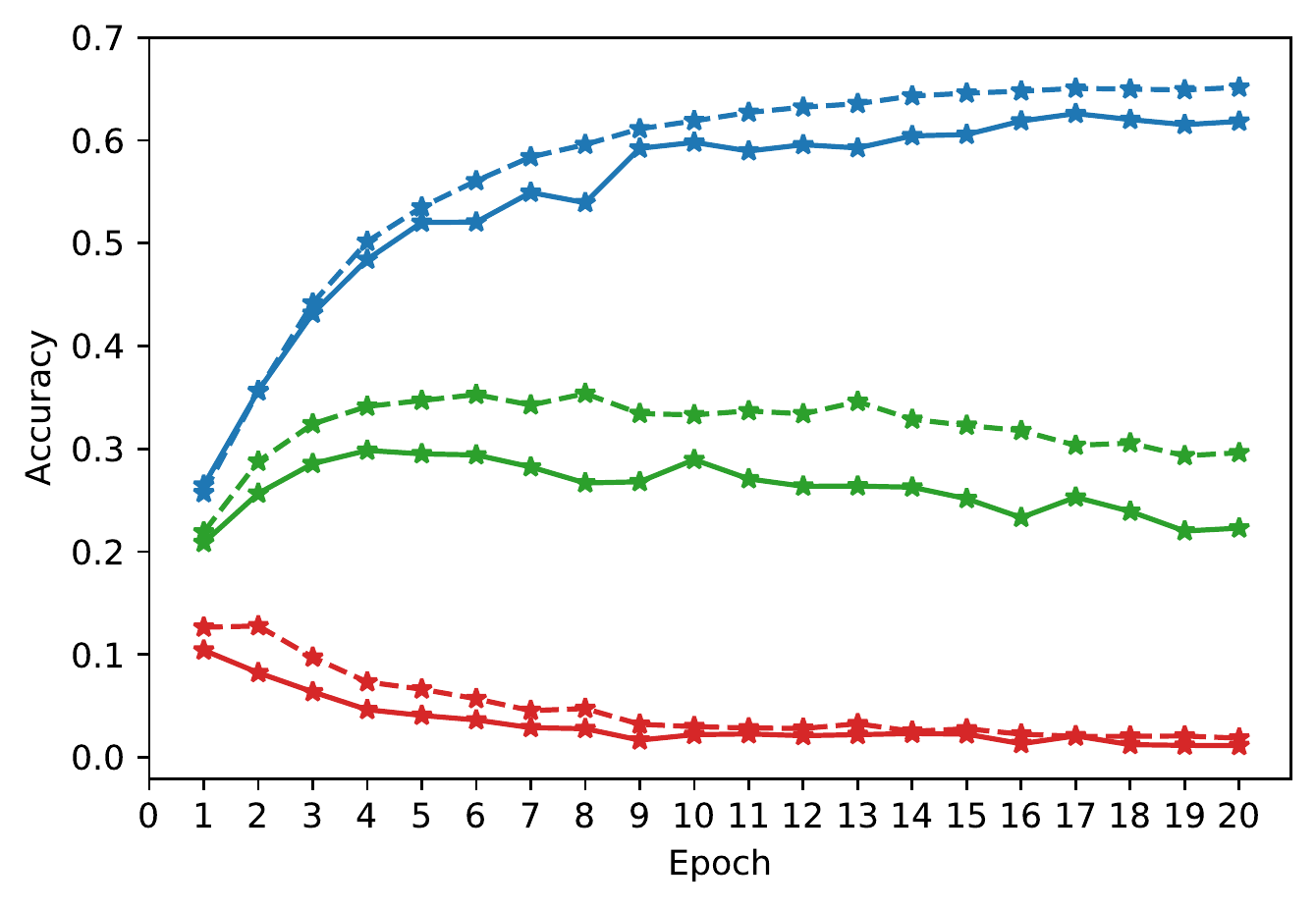}
\end{minipage}
}%
\hspace{-0.5cm}
\subfigure[20 snapshots]{
\begin{minipage}[t]{0.25\linewidth}
\centering
\includegraphics[width=1.5in]{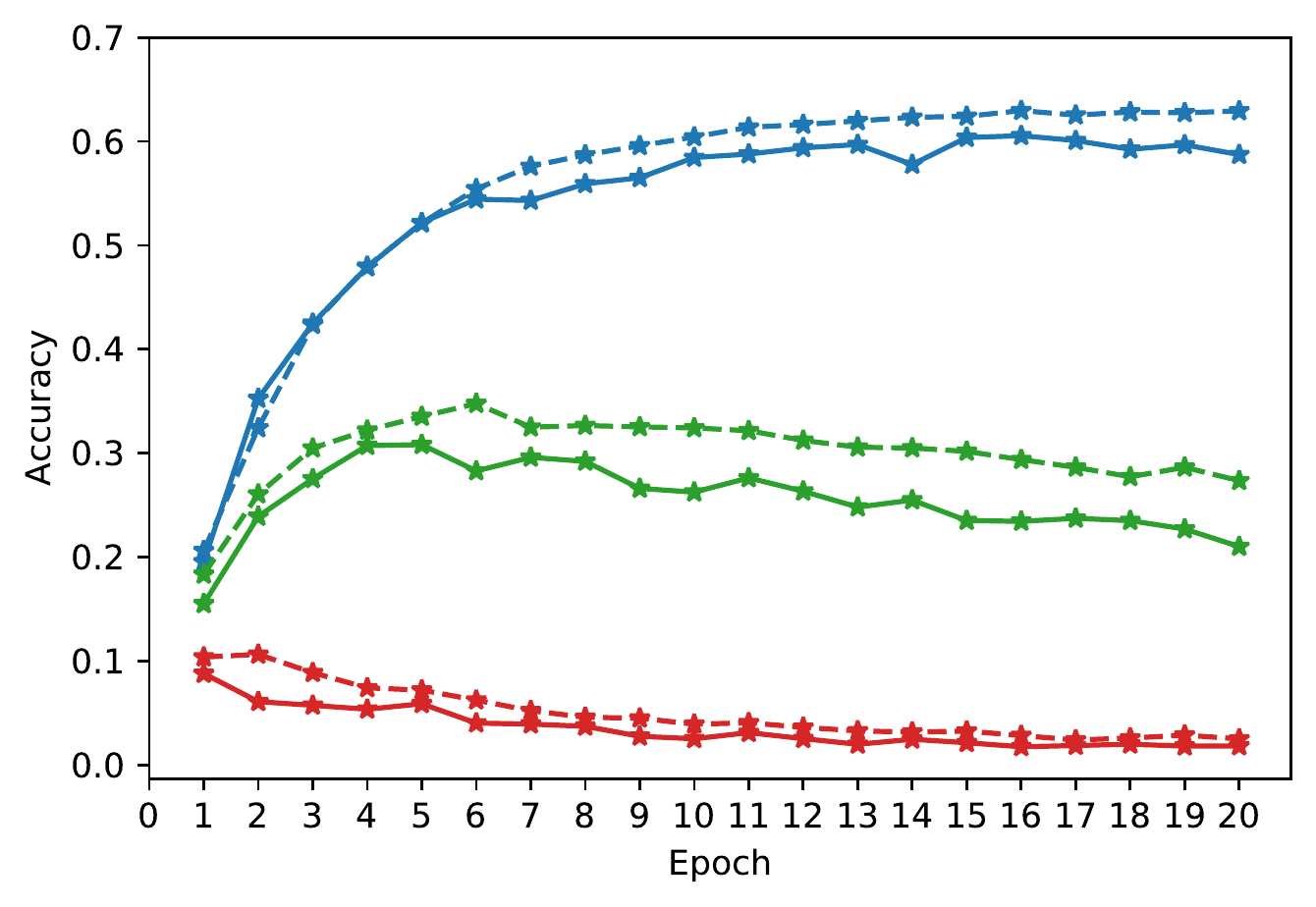}
\end{minipage}%
}%
\hspace{-0.2cm}
\subfigure[10 snapshots]{
\begin{minipage}[t]{0.25\linewidth}
\centering
\includegraphics[width=1.7in]{HBCIFAR10.pdf}
\end{minipage}%
}%
\caption{ Adversarial accuracy of neural networks on CIFAR-10 under FGSM, ensembled with different snapshots.}
\label{fig:CIFAR FGSM snapshots}
\end{figure}

\begin{figure}[!htb]
\centering
\subfigure[98 snapshots]{
\begin{minipage}[t]{0.25\linewidth}
\centering
\includegraphics[width=1.5in]{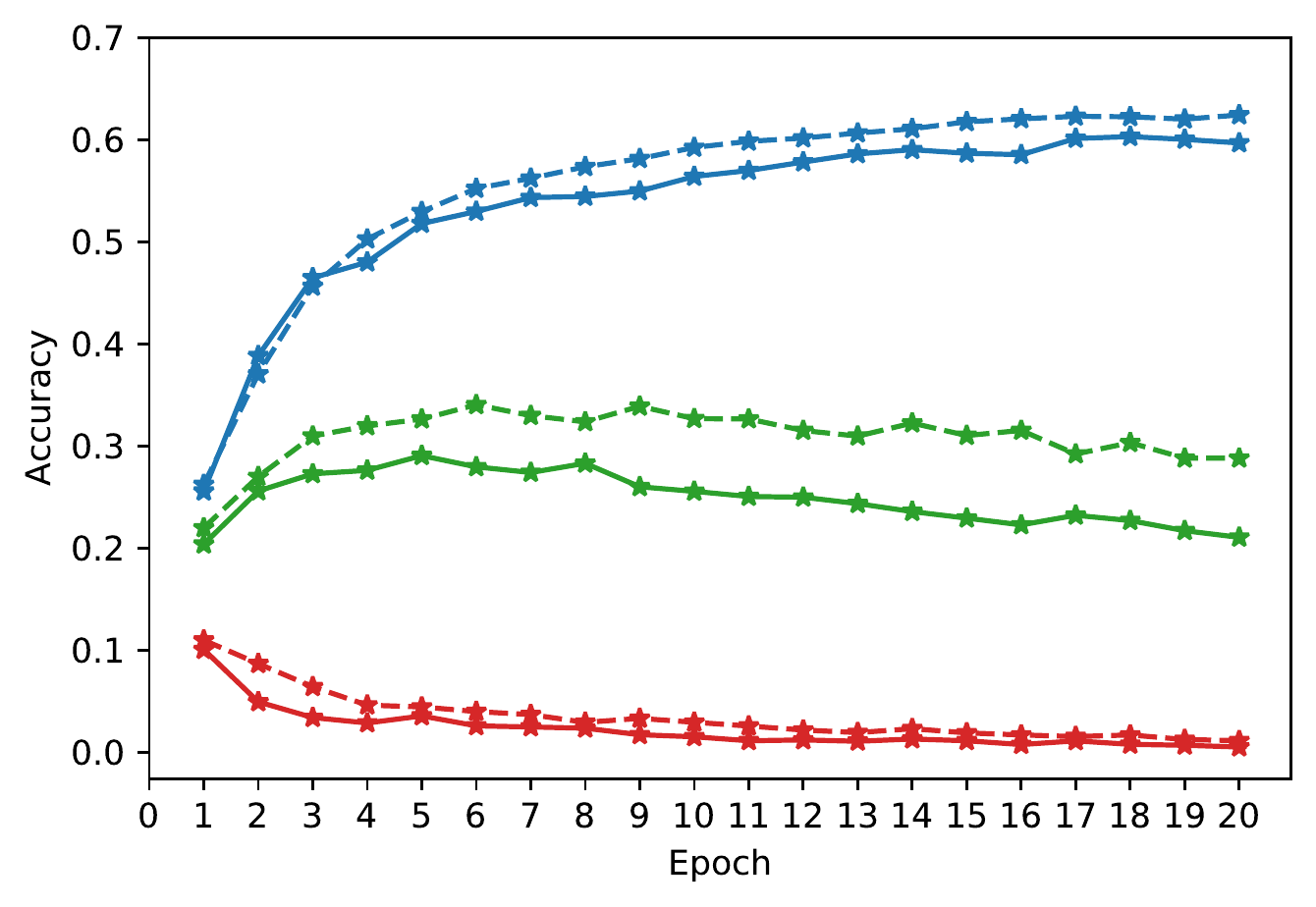}
\end{minipage}
}%
\hspace{-0.5cm}
\subfigure[40 snapshots]{
\begin{minipage}[t]{0.25\linewidth}
\centering
\includegraphics[width=1.5in]{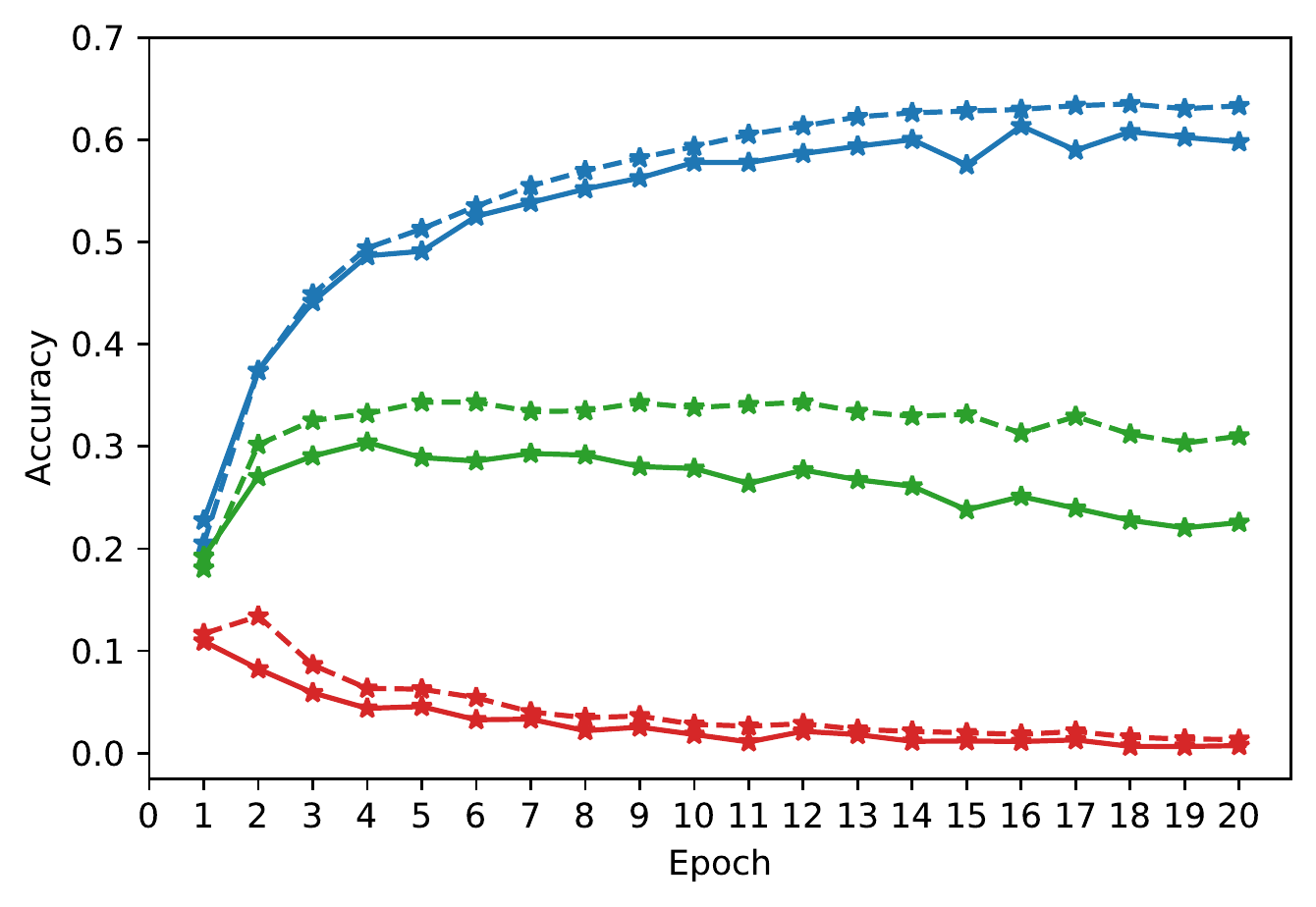}
\end{minipage}
}%
\hspace{-0.5cm}
\subfigure[20 snapshots]{
\begin{minipage}[t]{0.25\linewidth}
\centering
\includegraphics[width=1.5in]{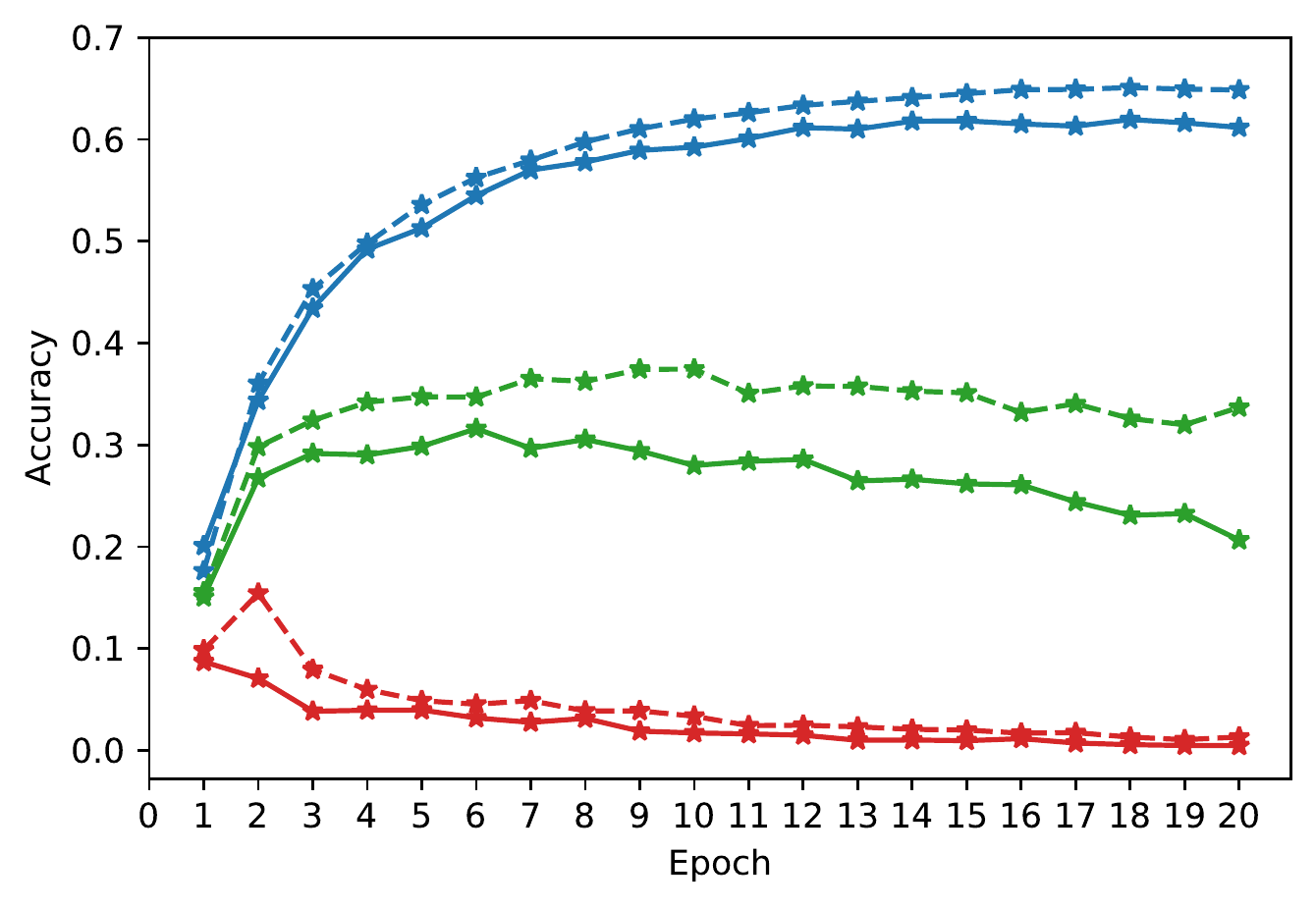}
\end{minipage}%
}%
\hspace{-0.4cm}
\subfigure[10 snapshots]{
\begin{minipage}[t]{0.25\linewidth}
\centering
\includegraphics[width=1.75in]{PGDHBCIFAR10.pdf}
\end{minipage}%
}%
\caption{ Adversarial accuracy of neural networks on CIFAR-10 under PGD, ensembled with different snapshots.}
\label{fig:CIFAR PGD snapshots}
\end{figure}
 From \Cref{fig:CIFAR FGSM snapshots} and \Cref{fig:CIFAR PGD snapshots} we see that the number of snapshots has ignorable influence on the adversarial accuracy. Because more snapshots consume more time spent on ensembling (in other words, prediction time) and more memory to store the historical weights from past iterations, our experiments suggest that using a small number of snapshots is sufficient in practice.

\subsubsection{Learning rate}

\begin{figure}[!htb]
\centering
\subfigure[0.002 learning rate]{
\begin{minipage}[t]{0.25\linewidth}
\centering
\includegraphics[width=1.5in]{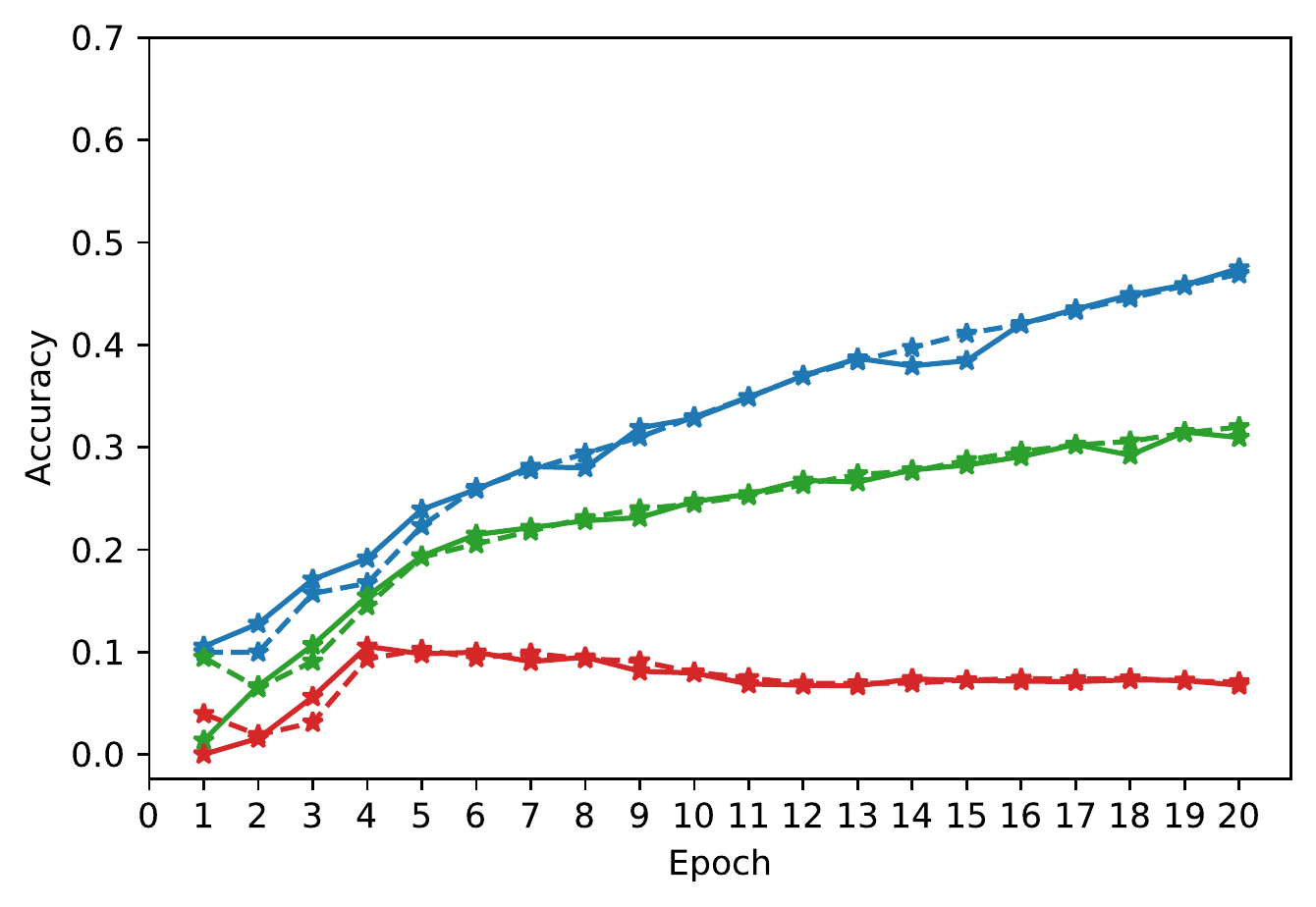}
\end{minipage}%
}%
\hspace{-0.5cm}
\subfigure[0.005 learning rate]{
\begin{minipage}[t]{0.25\linewidth}
\centering
\includegraphics[width=1.5in]{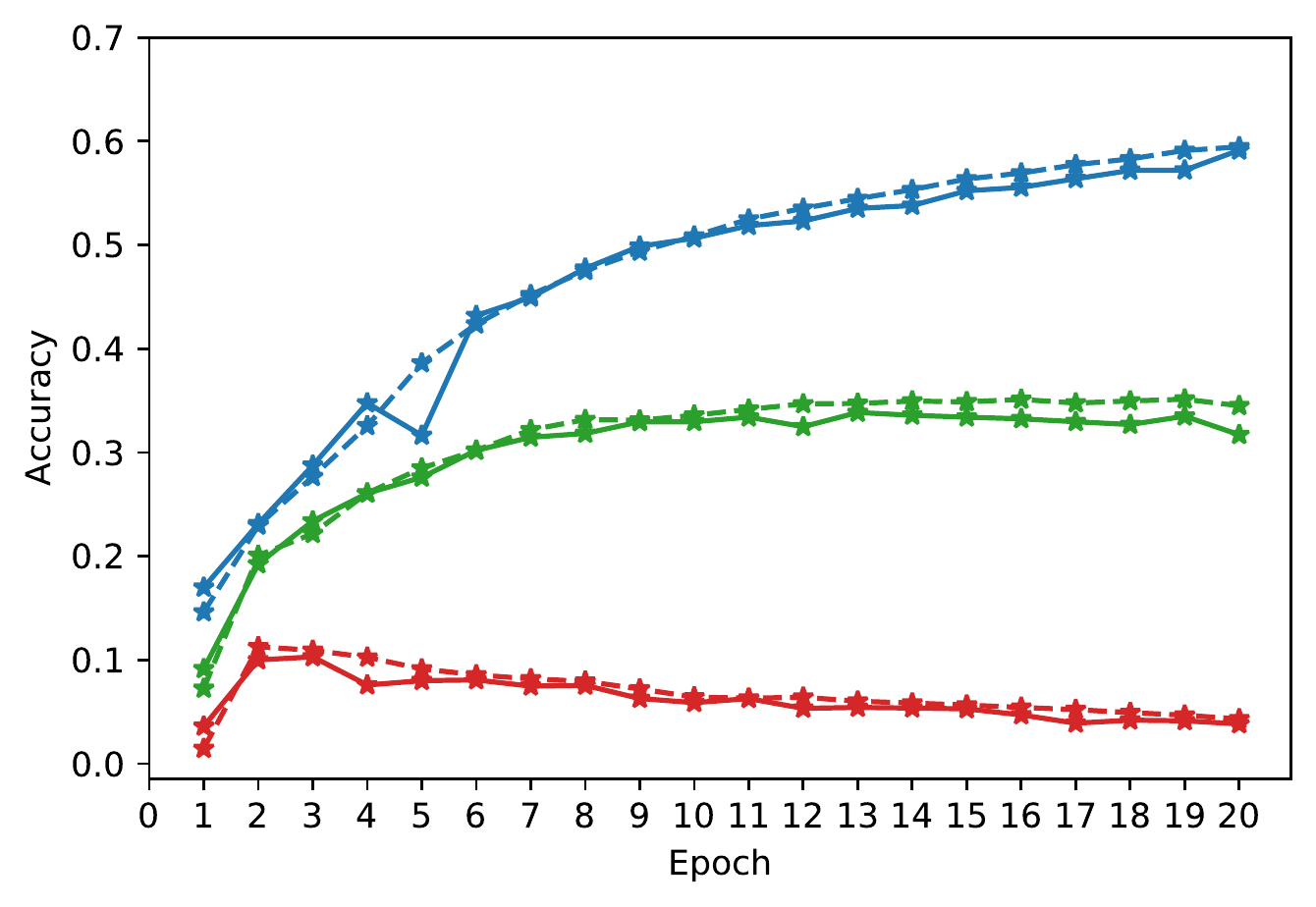}
\end{minipage}%
}%
\hspace{-0.5cm}
\subfigure[0.01 learning rate]{
\begin{minipage}[t]{0.25\linewidth}
\centering
\includegraphics[width=1.5in]{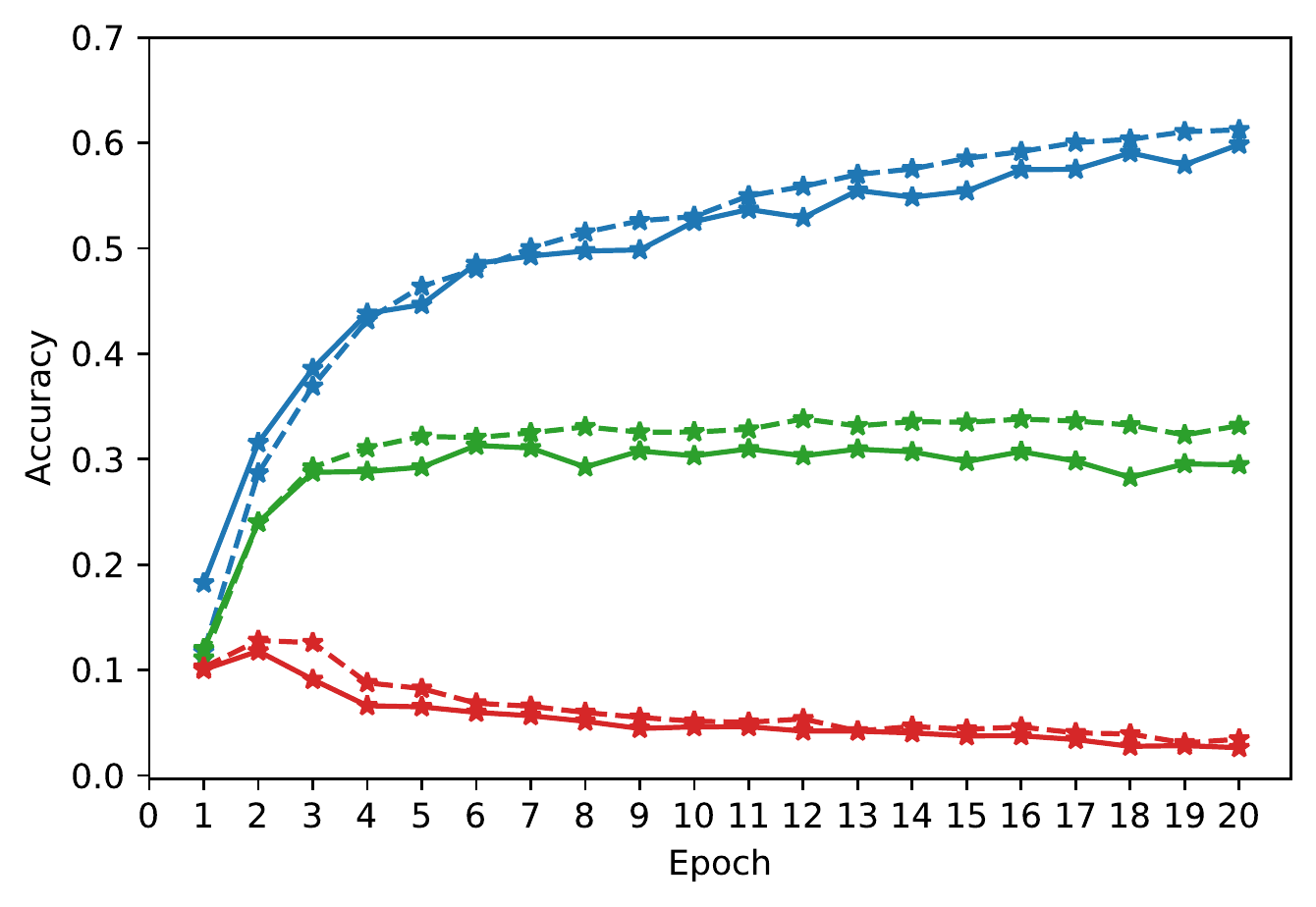}
\end{minipage}
}%
\hspace{-0.4cm}
\subfigure[0.02 learning rate]{
\begin{minipage}[t]{0.25\linewidth}
\centering
\includegraphics[width=1.75in]{HBCIFAR10.pdf}
\end{minipage}
}%
\centering
\caption{Adversarial accuracy of neural networks on CIFAR-10 under FGSM, trained with different learning rates.}
\label{fig:CIFAR FGSM lr}
\end{figure}

\begin{figure}[!htb]
\centering
\subfigure[0.002 learning rate]{
\begin{minipage}[t]{0.25\linewidth}
\centering
\includegraphics[width=1.5in]{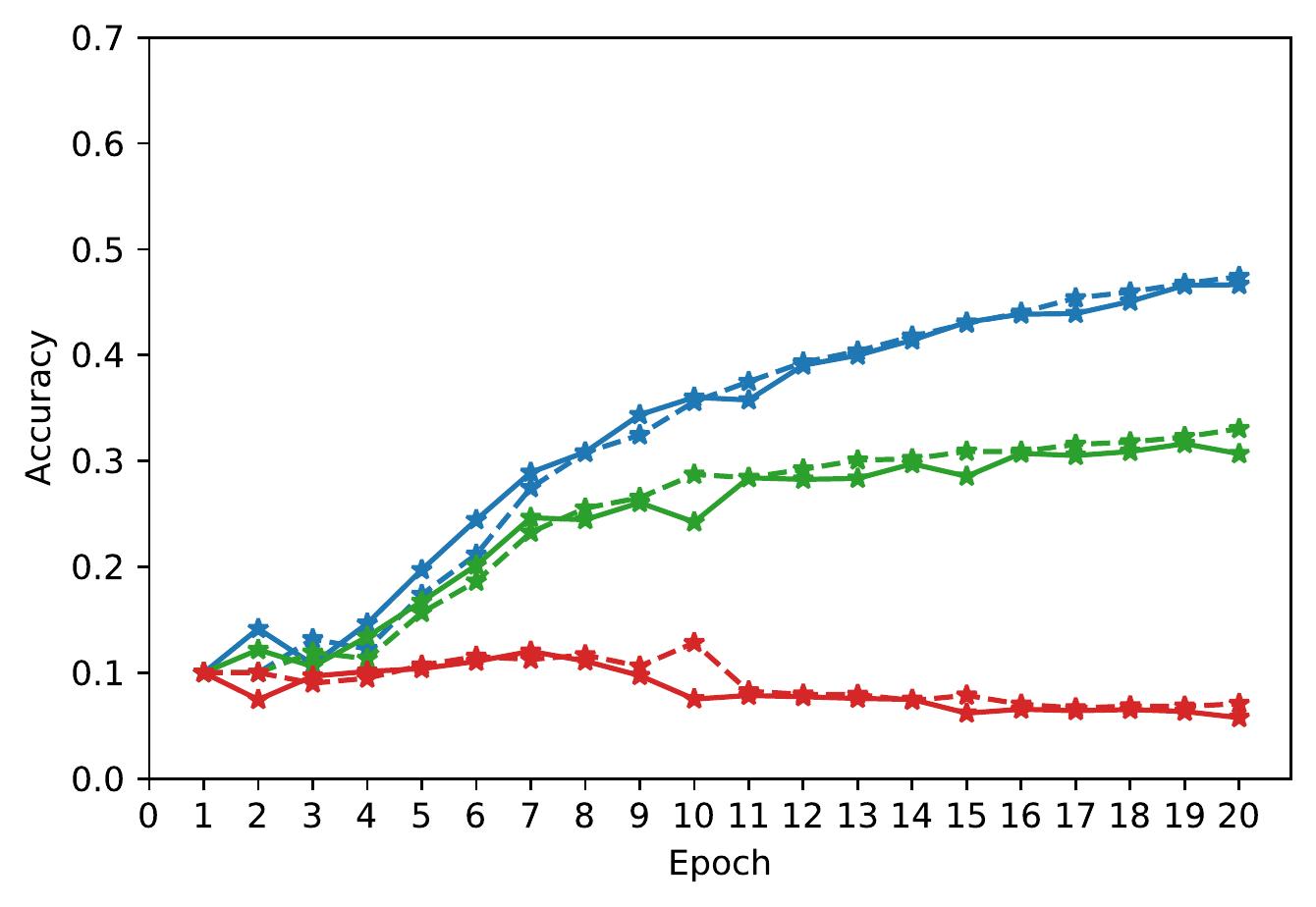}
\end{minipage}
}%
\hspace{-0.55cm}
\subfigure[0.005 learning rate]{
\begin{minipage}[t]{0.25\linewidth}
\centering
\includegraphics[width=1.5in]{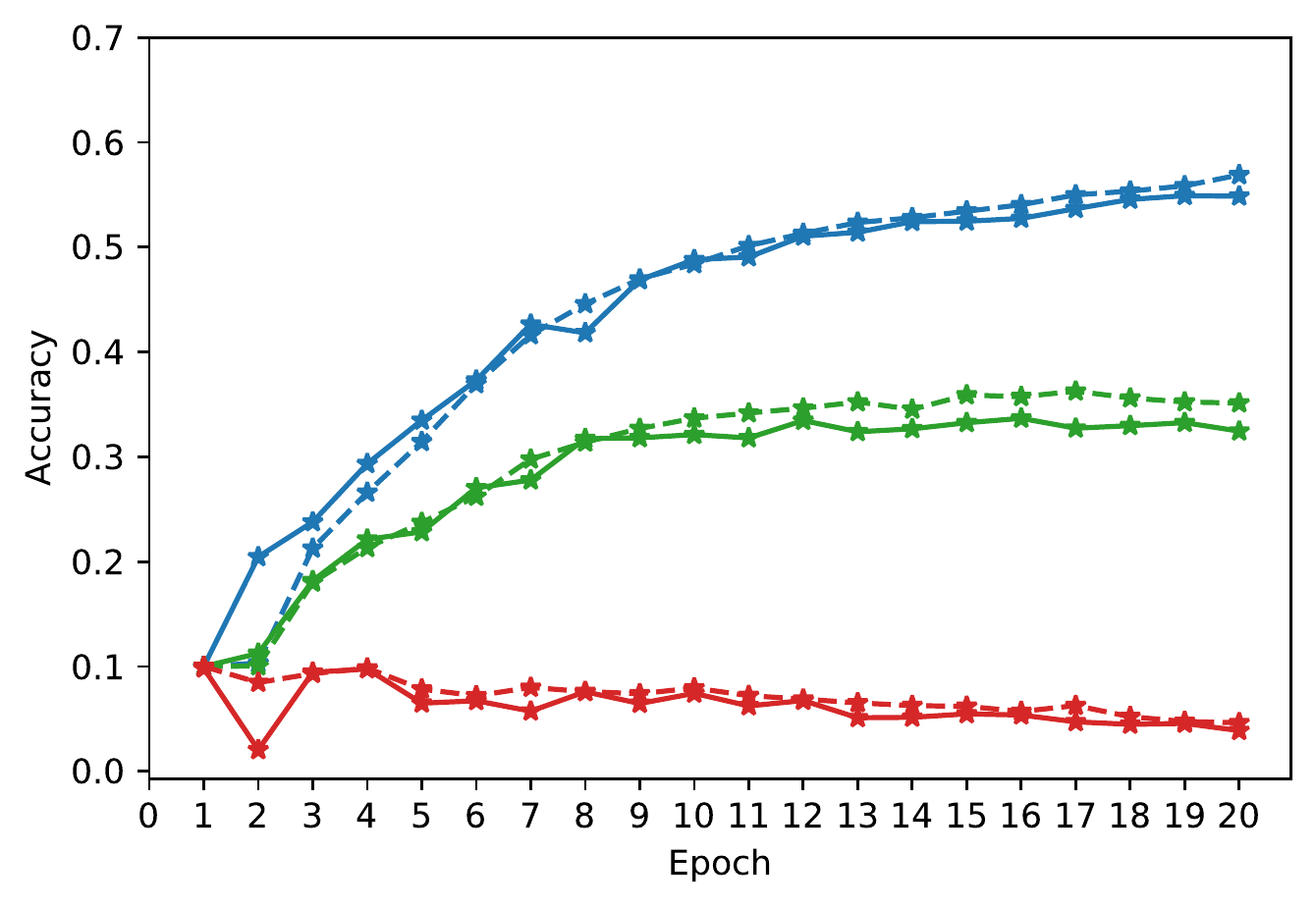}
\end{minipage}
}%
\hspace{-0.55cm}
\subfigure[0.01 learning rate]{
\begin{minipage}[t]{0.25\linewidth}
\centering
\includegraphics[width=1.5in]{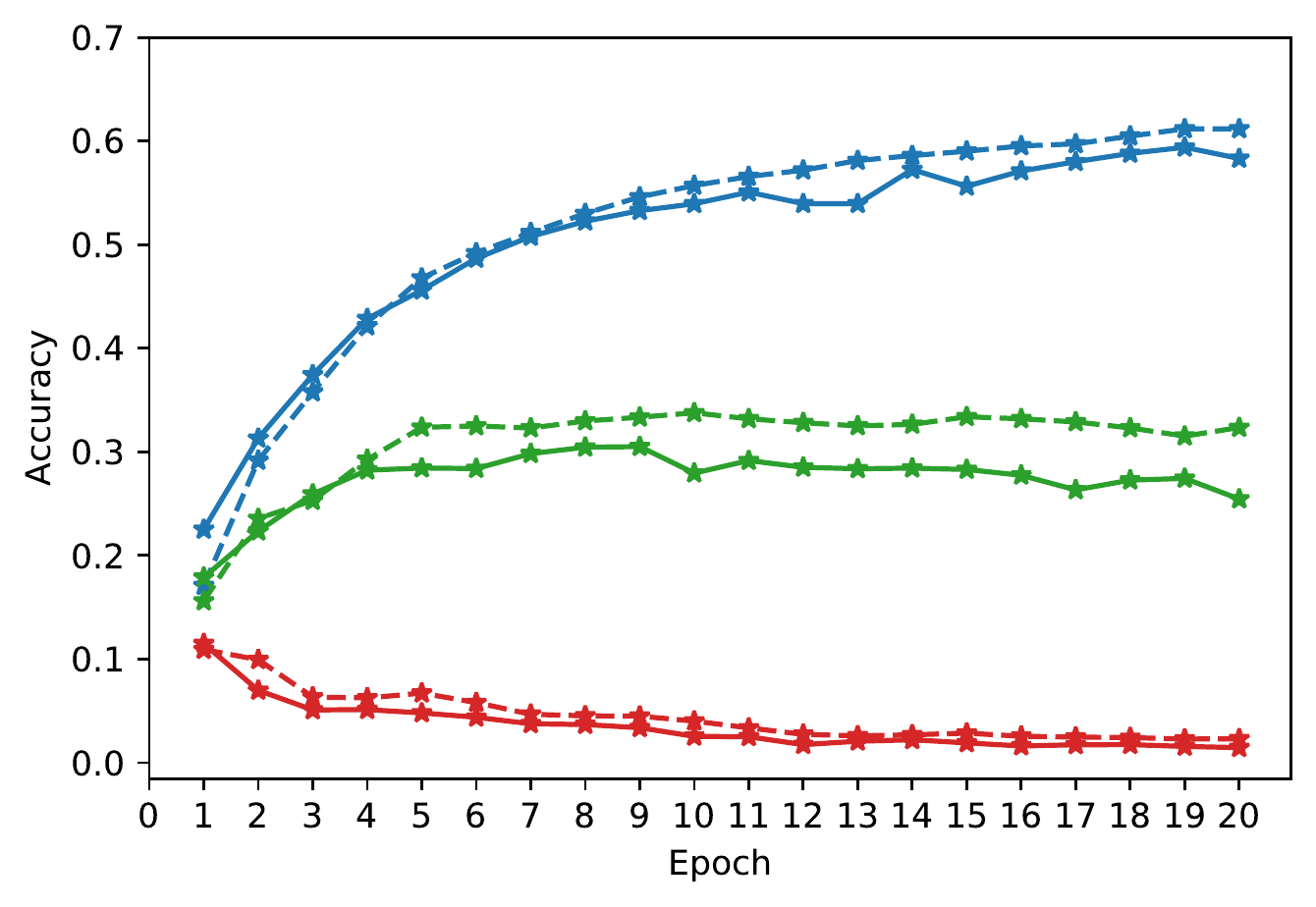}
\end{minipage}
}%
\hspace{-0.4cm}
\subfigure[0.02 learning rate]{
\begin{minipage}[t]{0.25\linewidth}
\centering
\includegraphics[width=1.75in]{PGDHBCIFAR10.pdf}
\end{minipage}
}%
\centering
\caption{Adversarial accuracy of neural networks on CIFAR-10 under PGD, trained with different learning rates.}
\label{fig:CIFAR PGD lr}
\end{figure}

 From \Cref{fig:CIFAR FGSM lr} and \Cref{fig:CIFAR PGD lr} we can see that a higher learning rate corresponds to a higher adversarial accuracy. Especially, larger learning rate has a more significant effect on the performance of our snapshot ensemble. It is clear that a small magnitude of learning rate produces non-robust results, but an excessively large learning rate could have the same bad effect since the models may not converge.

\subsubsection{Momentum coefficient}

\begin{figure}[!htb]
\centering
\subfigure[0.5 momentum]{
\begin{minipage}[t]{0.333\linewidth}
\centering
\includegraphics[width=2in]{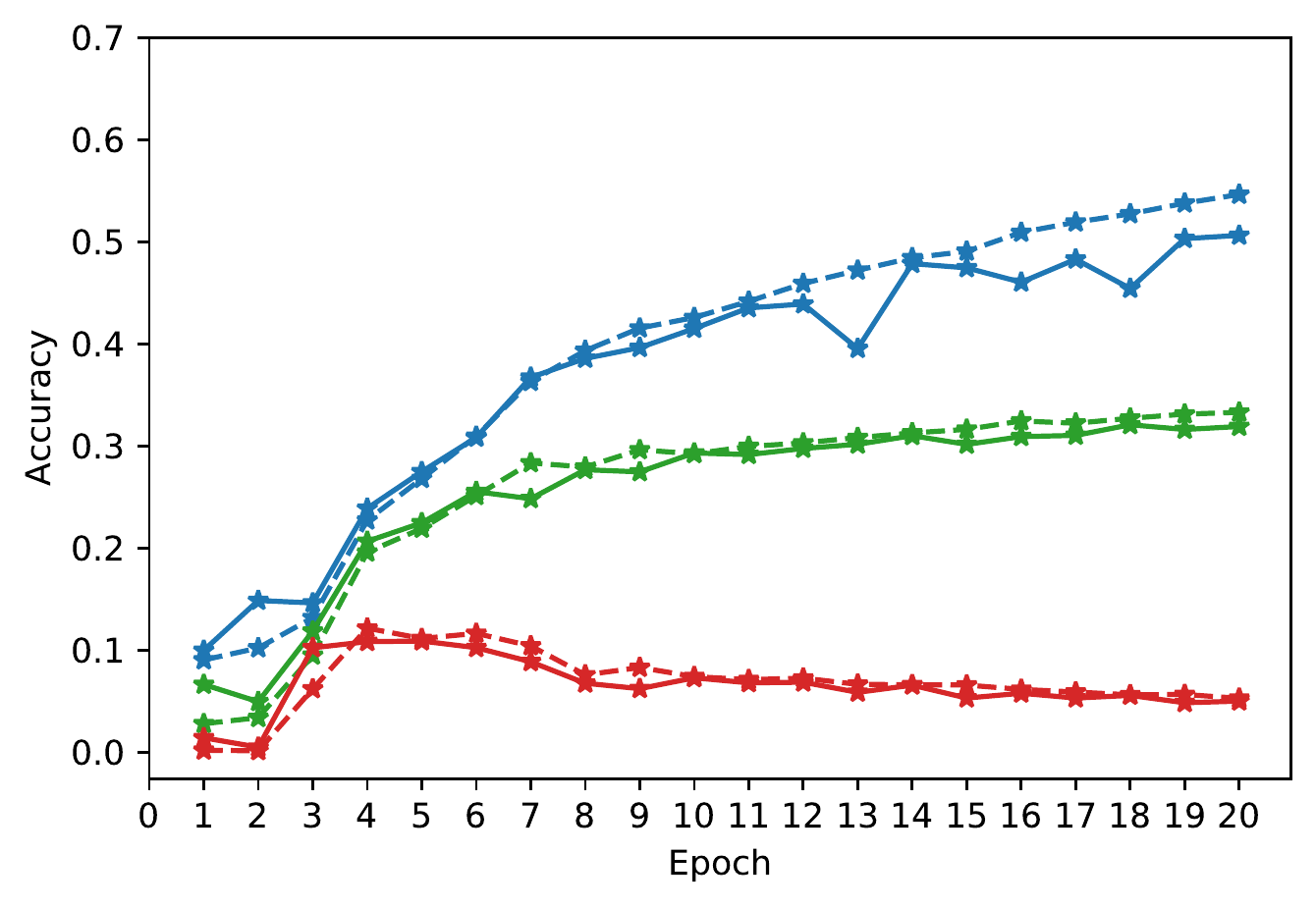}
\end{minipage}%
}%
\hspace{-0.6cm}
\subfigure[0.7 momentum]{
\begin{minipage}[t]{0.333\linewidth}
\centering
\includegraphics[width=2in]{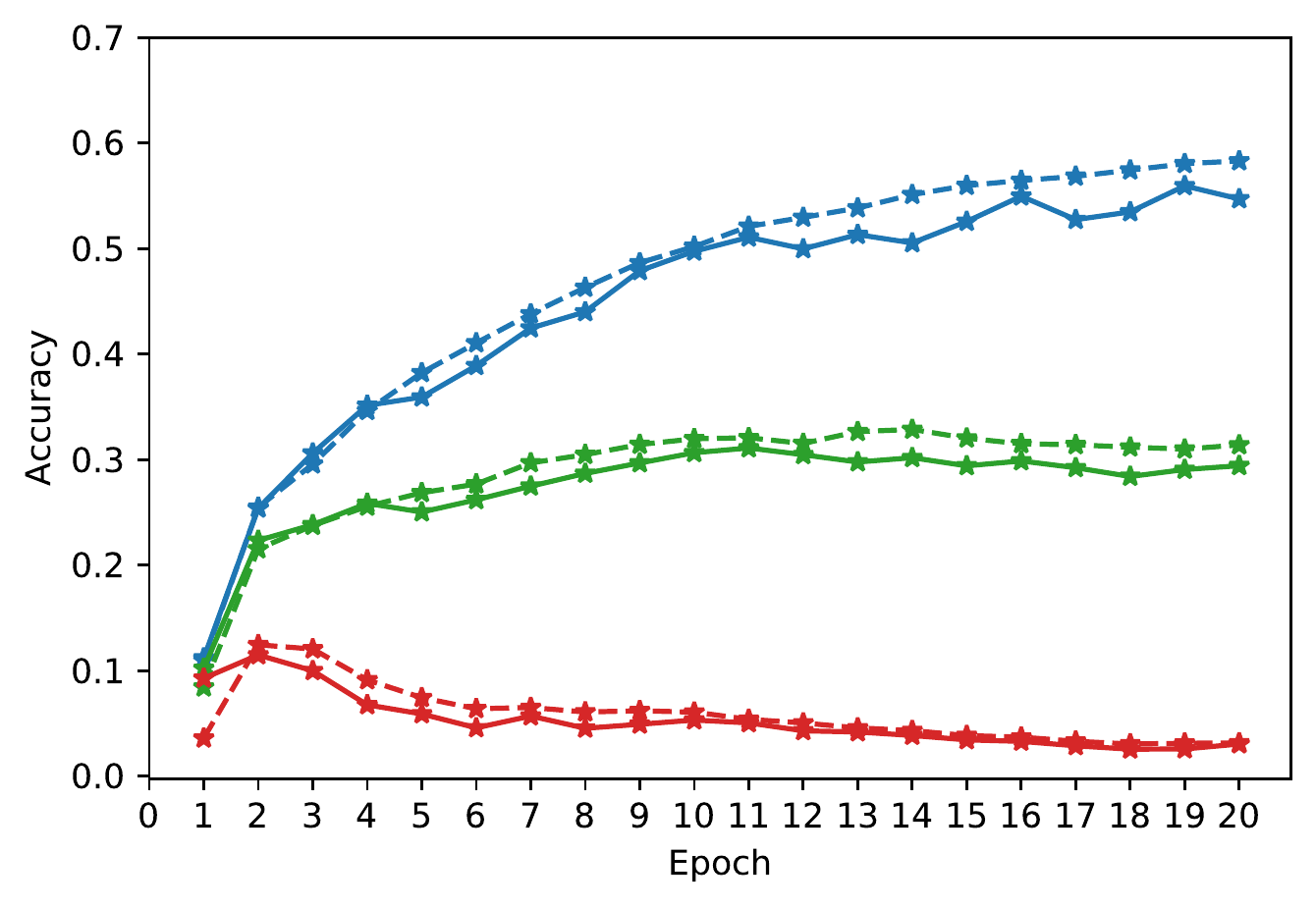}
\end{minipage}%
}%
\hspace{-0.4cm}
\subfigure[0.9 momentum]{
\begin{minipage}[t]{0.333\linewidth}
\centering
\includegraphics[width=2.3in]{HBCIFAR10.pdf}
\end{minipage}
}%
\centering
\caption{ Adversarial accuracy of neural networks on CIFAR-10 under FGSM, trained with different momentums.}
\label{fig:CIFAR FGSM mm}
\end{figure}

\begin{figure}[!htb]
\centering
\subfigure[0.5 momentum]{
\begin{minipage}[t]{0.333\linewidth}
\centering
\includegraphics[width=2in]{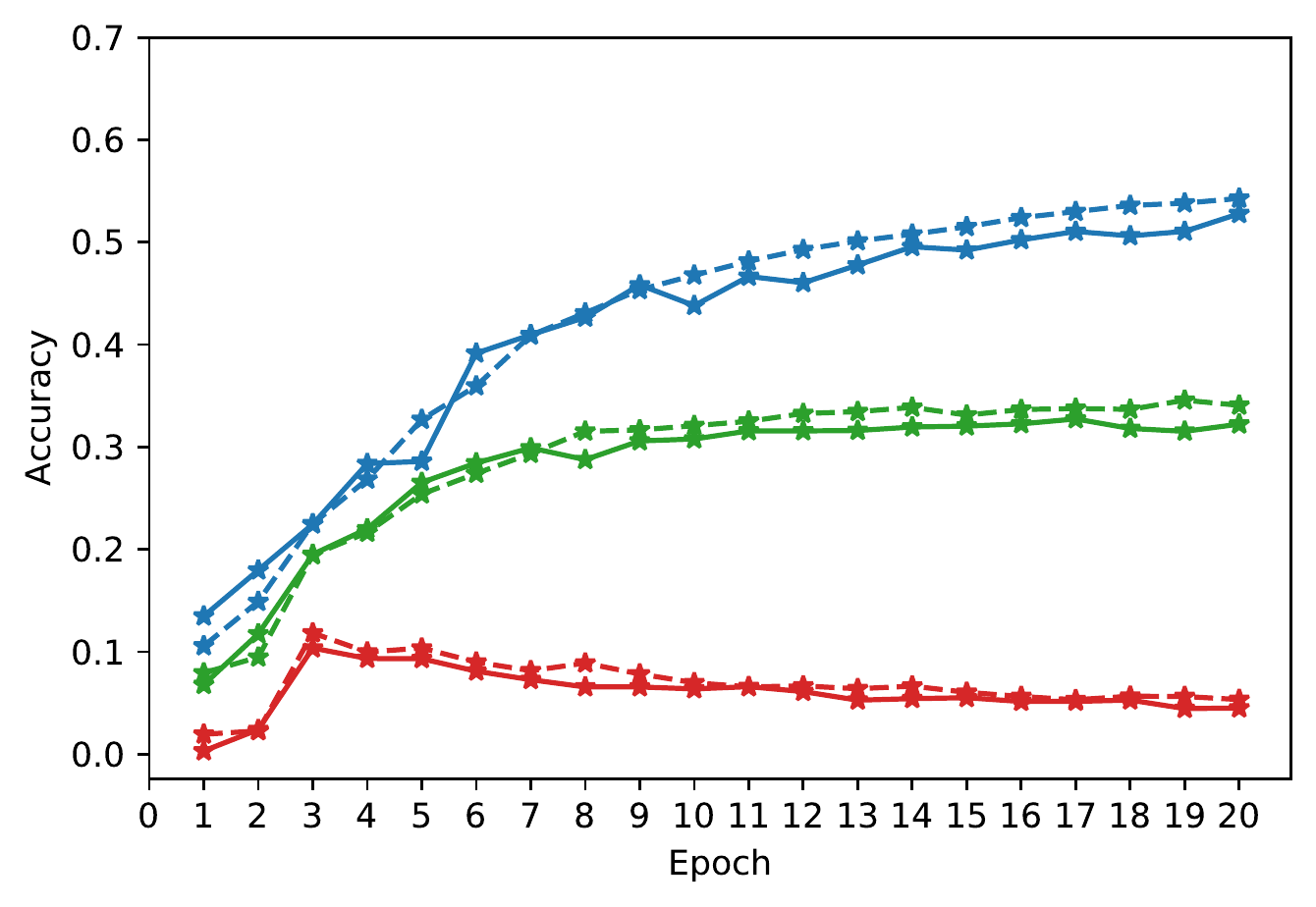}
\end{minipage}%
}%
\hspace{-0.6cm}
\subfigure[0.7 momentum]{
\begin{minipage}[t]{0.333\linewidth}
\centering
\includegraphics[width=2in]{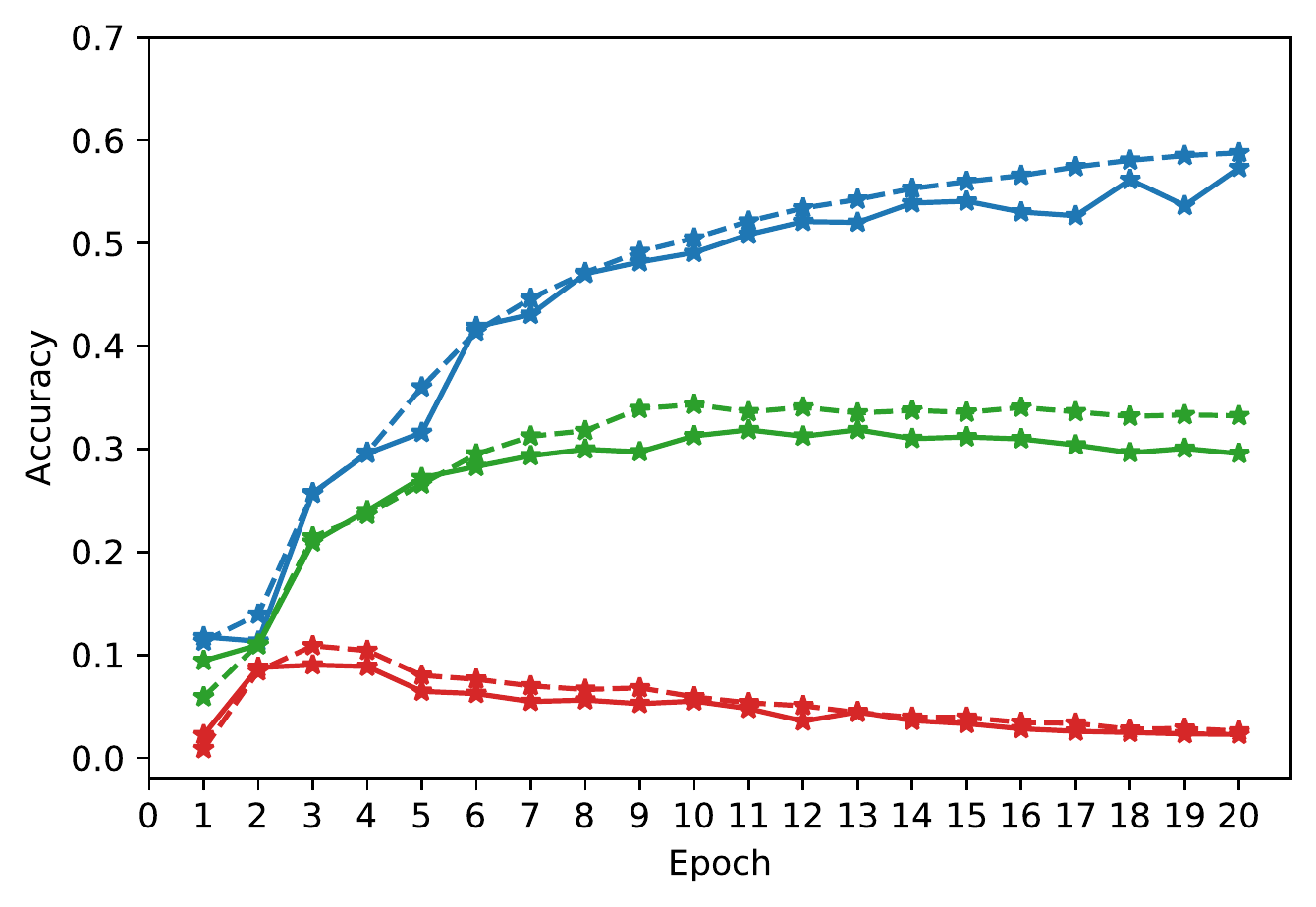}
\end{minipage}%
}%
\hspace{-0.4cm}
\subfigure[0.9 momentum]{
\begin{minipage}[t]{0.333\linewidth}
\centering
\includegraphics[width=2.3in]{PGDHBCIFAR10.pdf}
\end{minipage}
}%
\centering
\caption{Adversarial accuracy of neural networks on CIFAR-10 under PGD, trained with different momentums.}
\label{fig:CIFAR PGD mm}
\end{figure}

From \Cref{fig:CIFAR FGSM mm} and \Cref{fig:CIFAR PGD mm} we also observe that a higher momentum corresponds to a higher adversarial accuracy and again that higher momentum with our snapshot ensemble improves the performance more significantly.

\subsection{MNIST}

The above conclusions for the CIFAR-10 dataset are also applicable for the MNIST dataset, but due to the characteristics of the MNIST images, the adversarial robustness is very strong, and the graphically slight increases cannot be clearly displayed, we do not repeatedly show them here. You may find the section in more details in the appendix.

\section{Conclusion}
In summary, we propose a new adversarial training procedure that does not require any changes to the training but incorporates the ensemble method during iterations. Our method takes the same computational complexity as the traditional adversarial training and improves the accuracy consistently. In fact, our ensemble is similar to the snapshot ensemble, but easier to implement, since we stores sets of parameters from last iterations instead of the local minima.

At the core of our method is the randomness of stored parameters. It naturally begs the question of what types of noises are present and what are their effects. For example, using standard SGD or most gradient methods with minibatch, there exists the sampling noise which empirically improves the robust accuracy. In gradient methods like stochastic gradient Langevin dynamics, the noise is isotropic (independent of data) and may have different effects than the sampling noise. Another example could be the random data augmentation, especially in image tasks. Training (either regularly or adversarially) on such tasks involve the transformation noise into the process, leading to uncertain effect on robustness. It would be desirable to conduct further experiments on these various sources of noise and on larger scale of datasets.

\bibliographystyle{unsrtnat}
\bibliography{references}  





\clearpage
\appendix
\section{Neural Network Architectures}
The neural network architectures for MNIST and CIFAR-10 datasets are as follows: 
\begin{figure}[!htb]
    \centering
    \includegraphics[width=16cm]{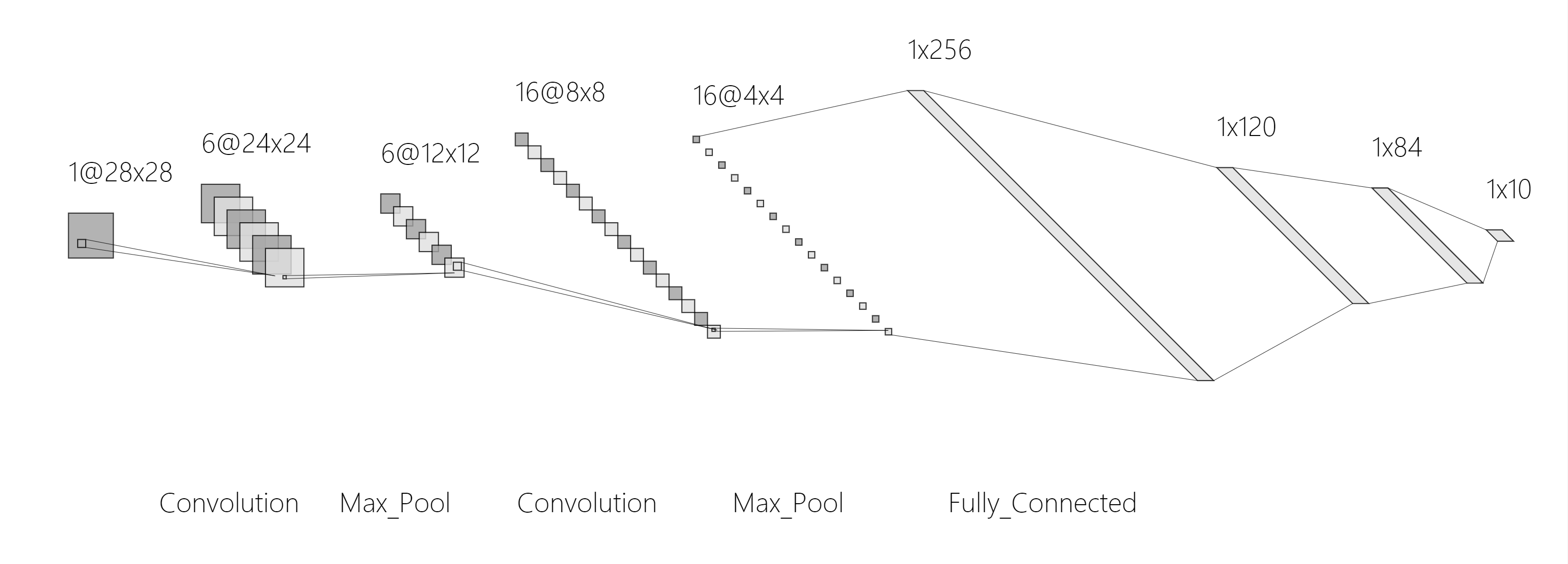}
    \caption{MNIST Neural Network Architecture}
    \label{fig:mnist}
\end{figure}
\begin{figure}[!htb]
    \centering
    \includegraphics[width=15.5cm]{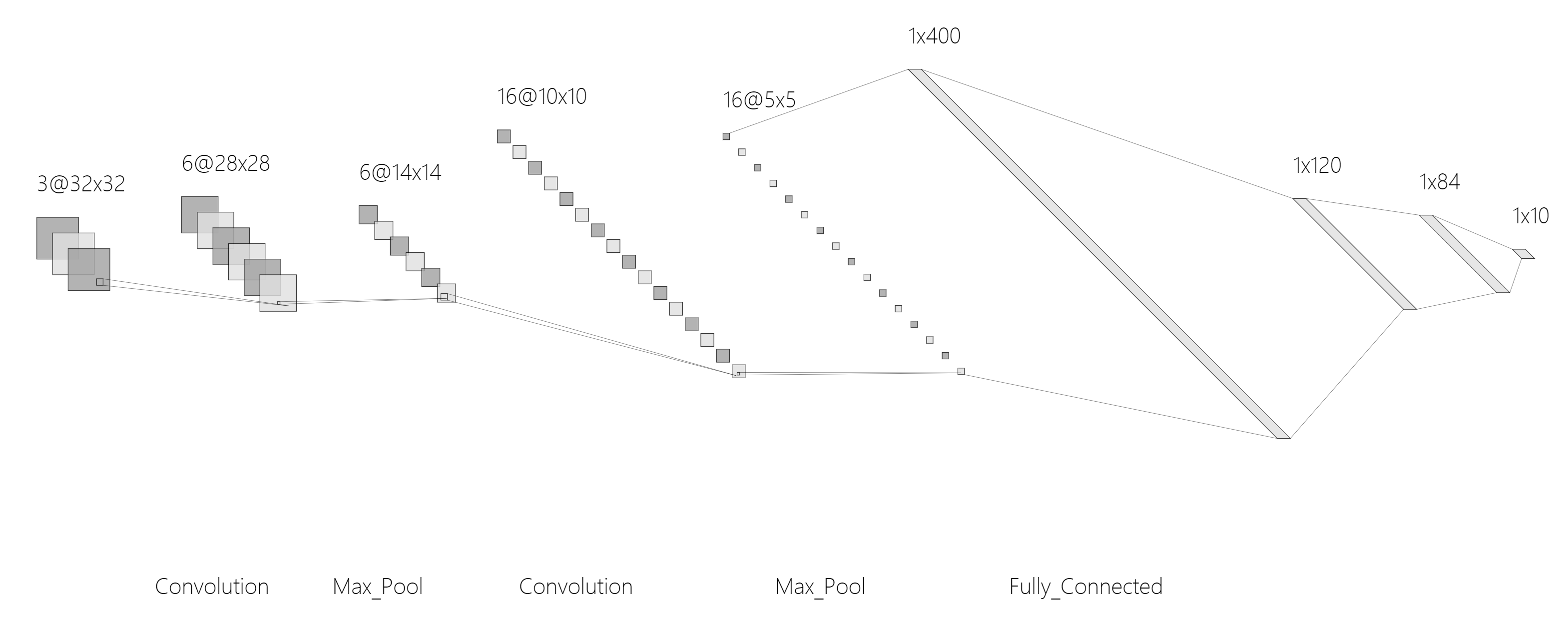}
    \caption{CIFAR-10 Neural Network Architecture}
    \label{fig:cifar10}
\end{figure}

\section{MNIST}
Due to MNIST's extreme robustness, we have enlarged some graphs and have removed several beginning epochs from some of the graphs to amplify the differences.
\subsubsection{Outer minimization optimizers}

\begin{figure}[!htb]
\centering
\subfigure[SGD]{
\begin{minipage}[t]{0.333\linewidth}
\centering
\includegraphics[width=2.in]{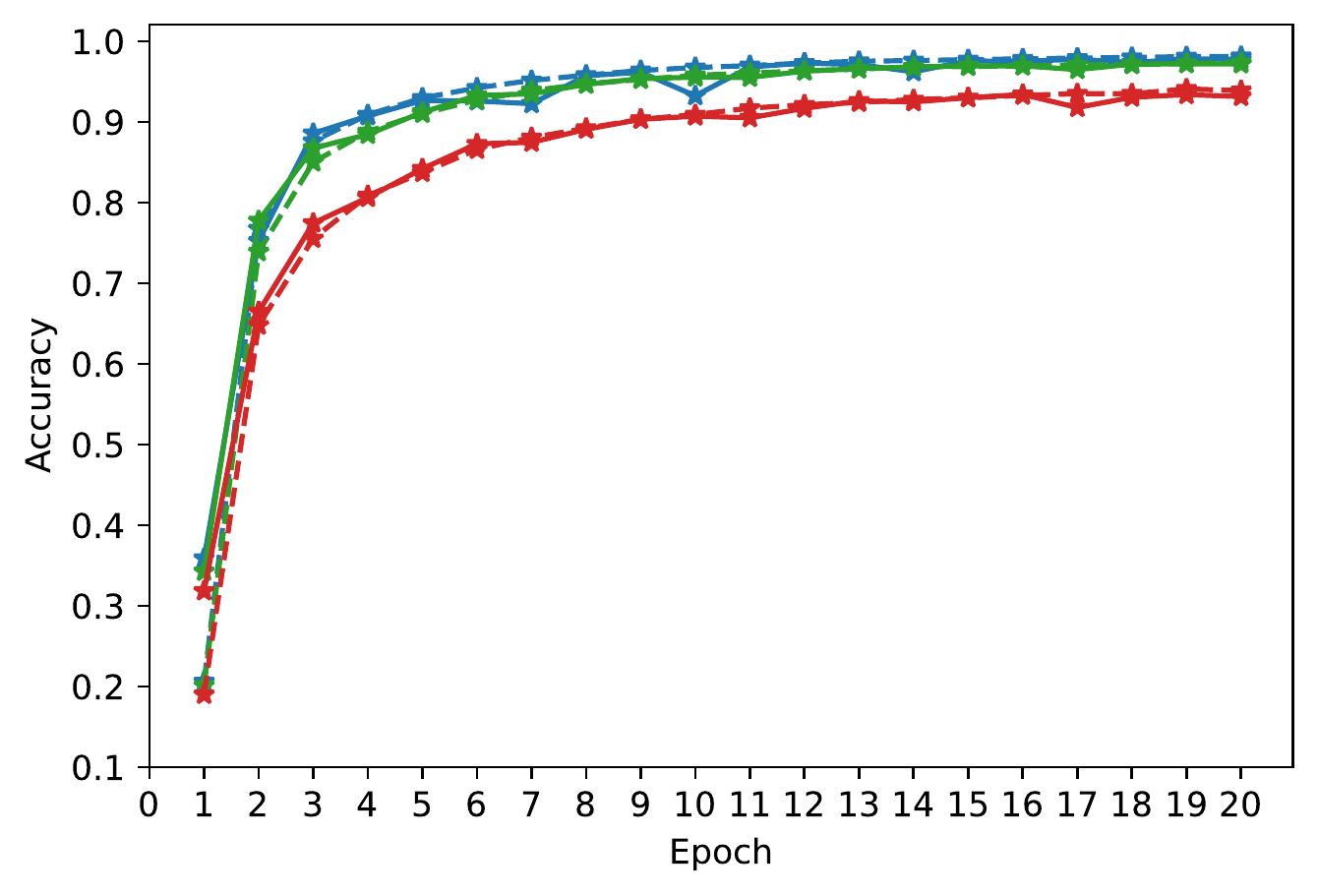}
\end{minipage}%
}%
\hspace{-0.5cm}
\subfigure[NAG]{
\begin{minipage}[t]{0.333\linewidth}
\centering
\includegraphics[width=2.in]{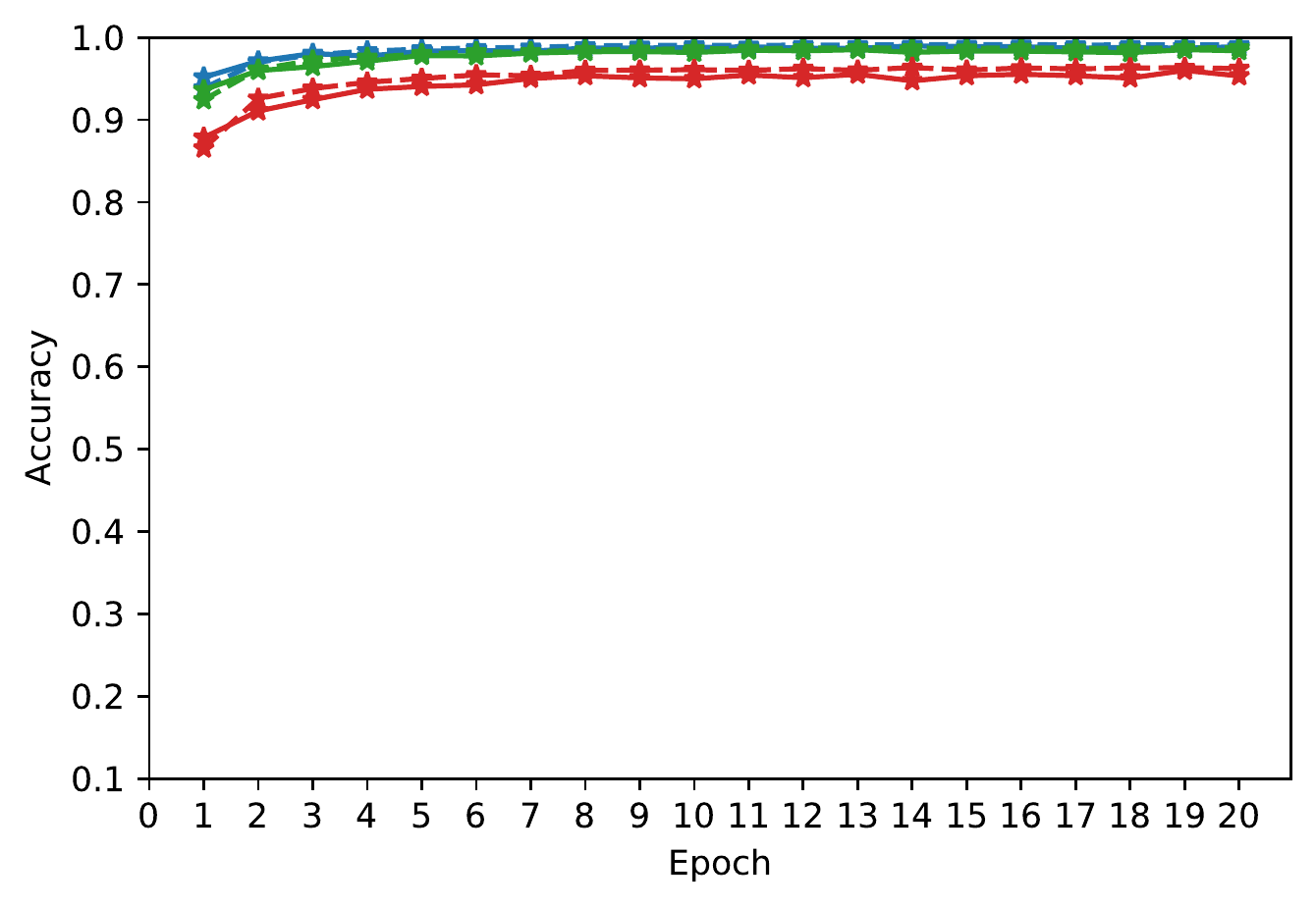}
\end{minipage}%
}%
\hspace{-0.4cm}
\subfigure[HB]{
\begin{minipage}[t]{0.333\linewidth}
\centering
\includegraphics[width=2.3in]{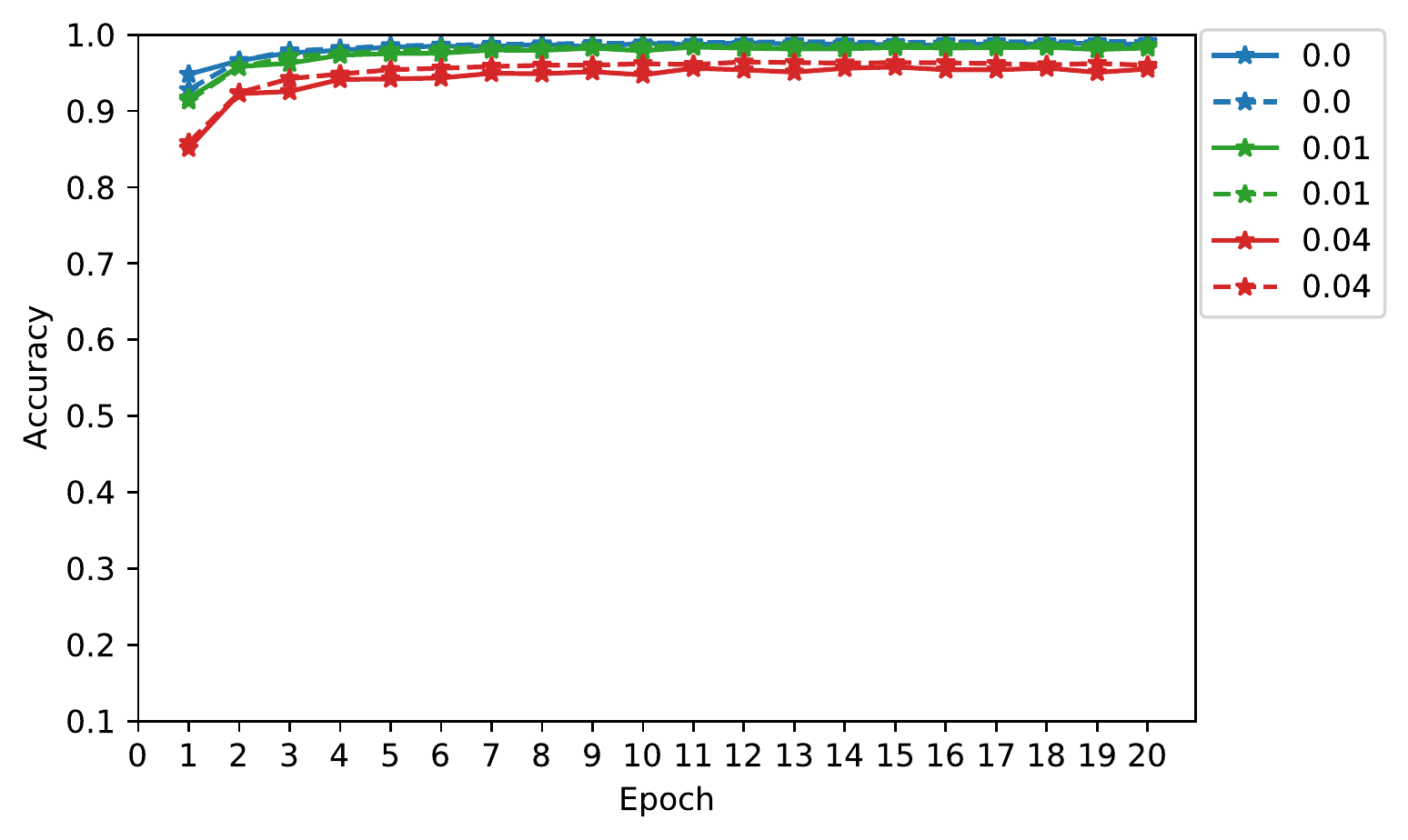}
\end{minipage}
}%

\subfigure[SGD (zoomed in)]{
\begin{minipage}[t]{0.333\linewidth}
\centering
\includegraphics[width=2.in]{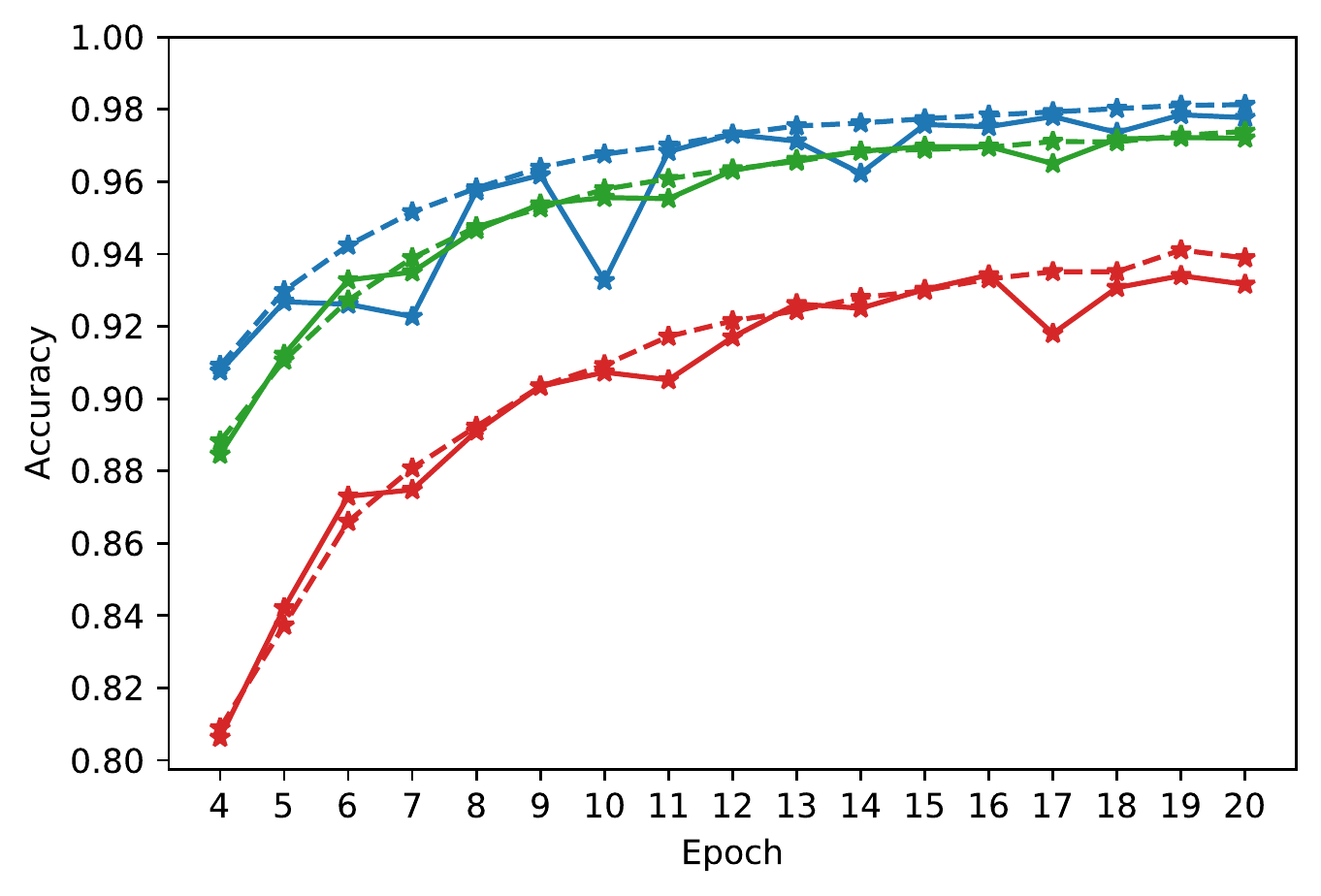}
\end{minipage}%
}%
\hspace{-0.5cm}
\subfigure[NAG (zoomed in)]{
\begin{minipage}[t]{0.333\linewidth}
\centering
\includegraphics[width=2.in]{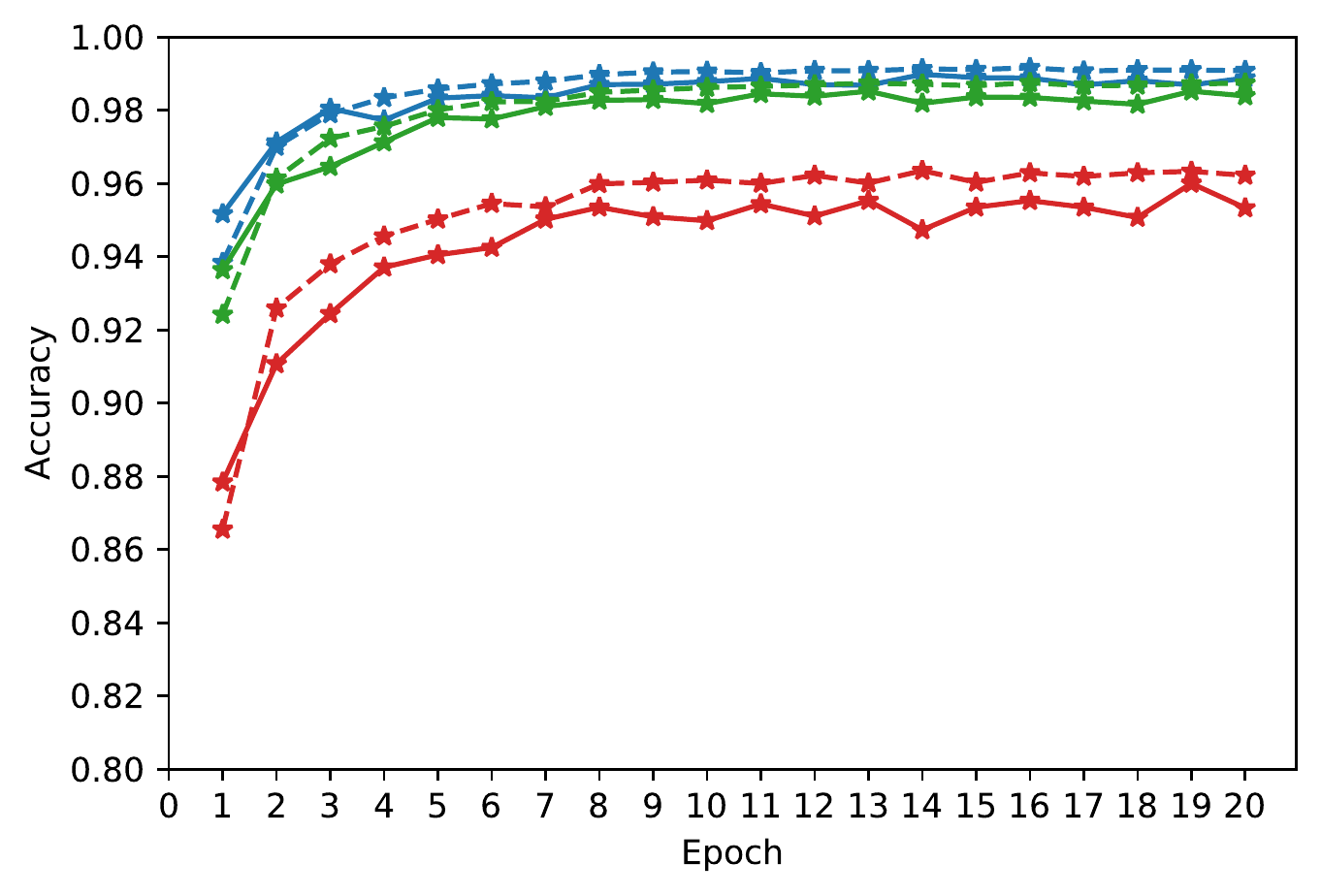}
\end{minipage}%
}%
\hspace{-0.4cm}
\subfigure[HB (zoomed in)]{
\begin{minipage}[t]{0.333\linewidth}
\centering
\includegraphics[width=2.3in]{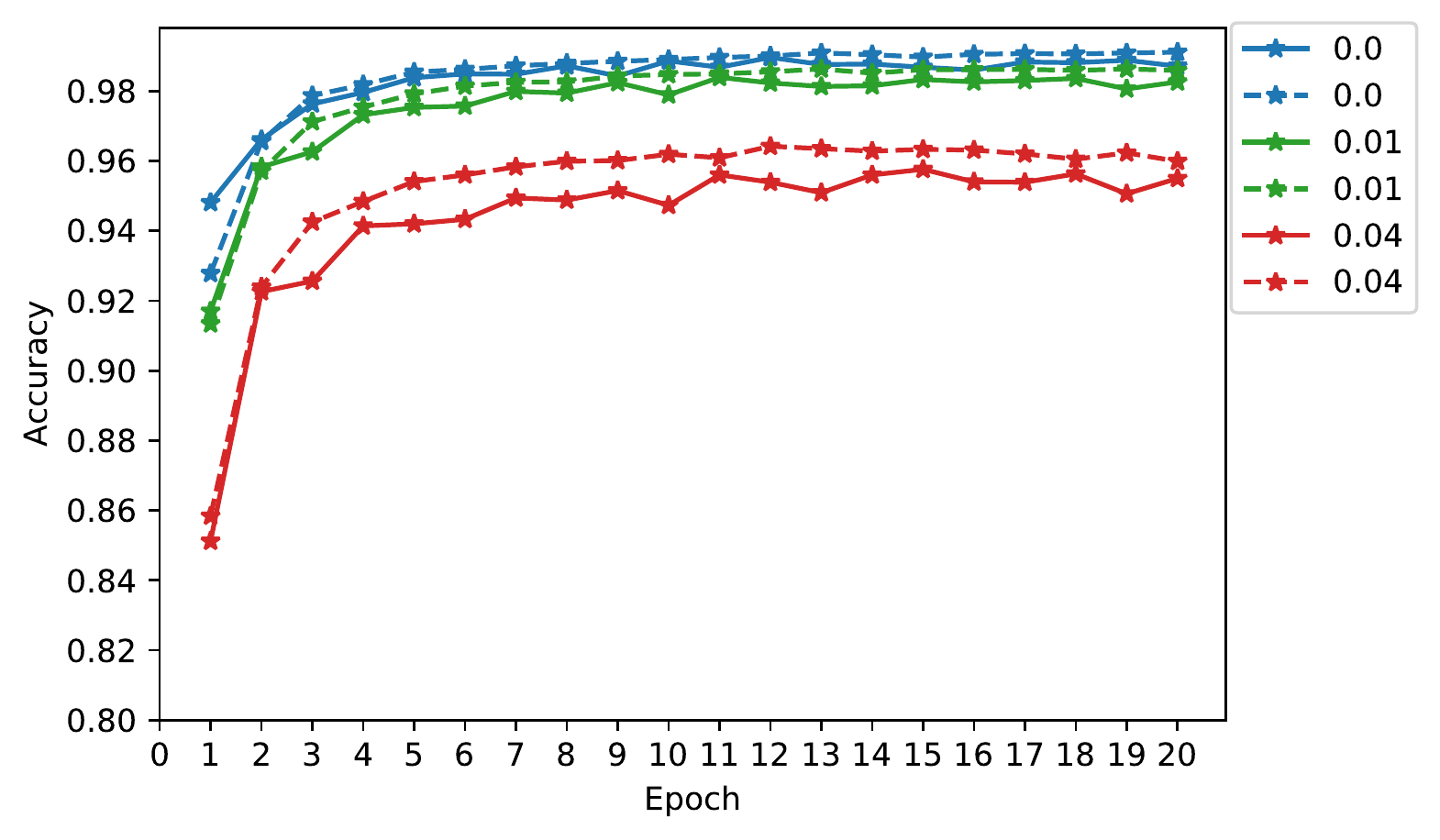}
\end{minipage}
}%
\caption{Adversarial accuracy of neural networks on MNIST under FGSM, trained with different optimizers.}
\label{fig:MNIST FGSM optimizer}
\end{figure}

\begin{figure}[!htb]
\centering
\subfigure[SGD]{
\begin{minipage}[t]{0.333\linewidth}
\centering
\includegraphics[width=2.in]{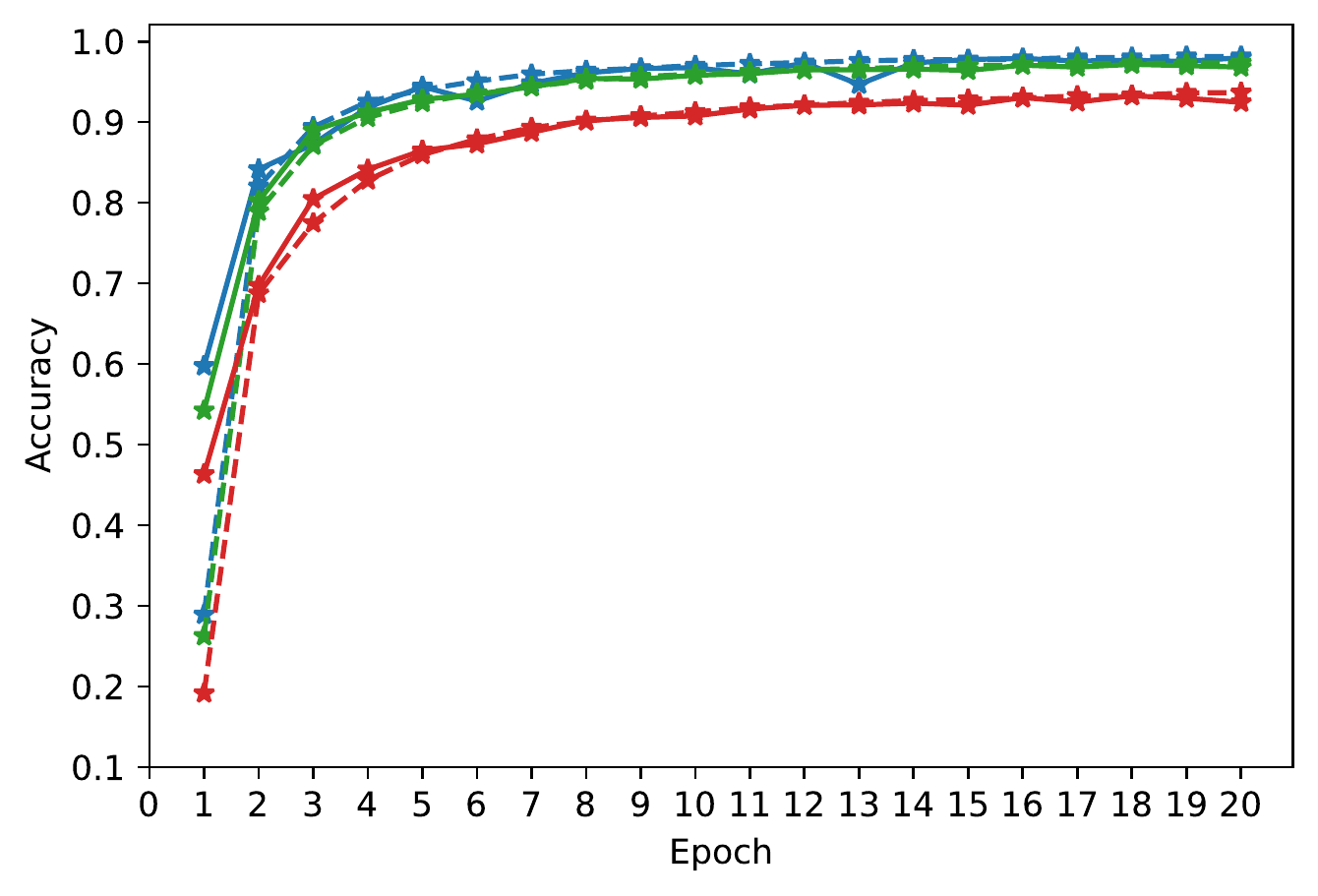}
\end{minipage}%
}%
\hspace{-0.5cm}
\subfigure[NAG]{
\begin{minipage}[t]{0.333\linewidth}
\centering
\includegraphics[width=2.in]{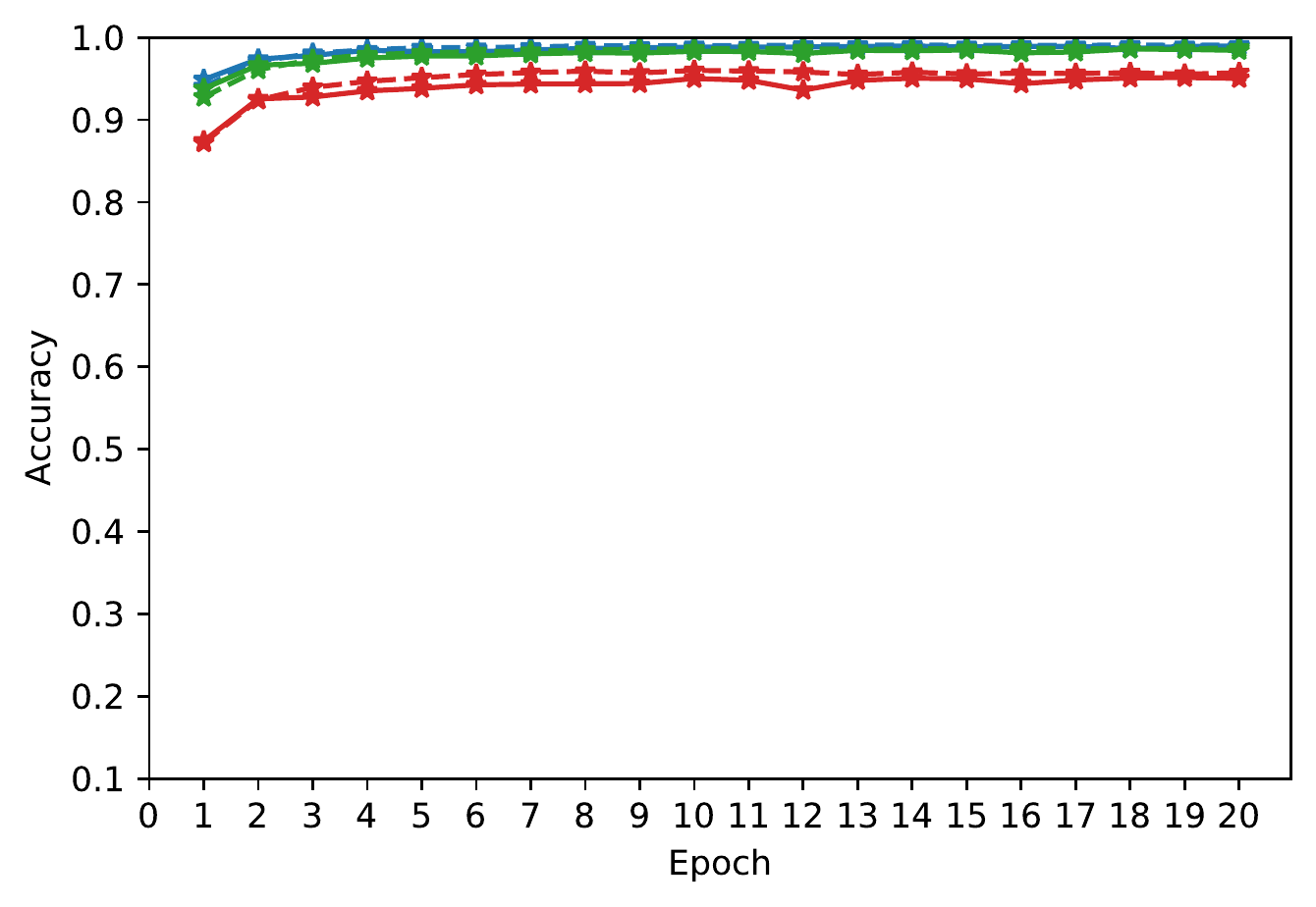}
\end{minipage}%
}%
\hspace{-0.4cm}
\subfigure[HB]{
\begin{minipage}[t]{0.333\linewidth}
\centering
\includegraphics[width=2.3in]{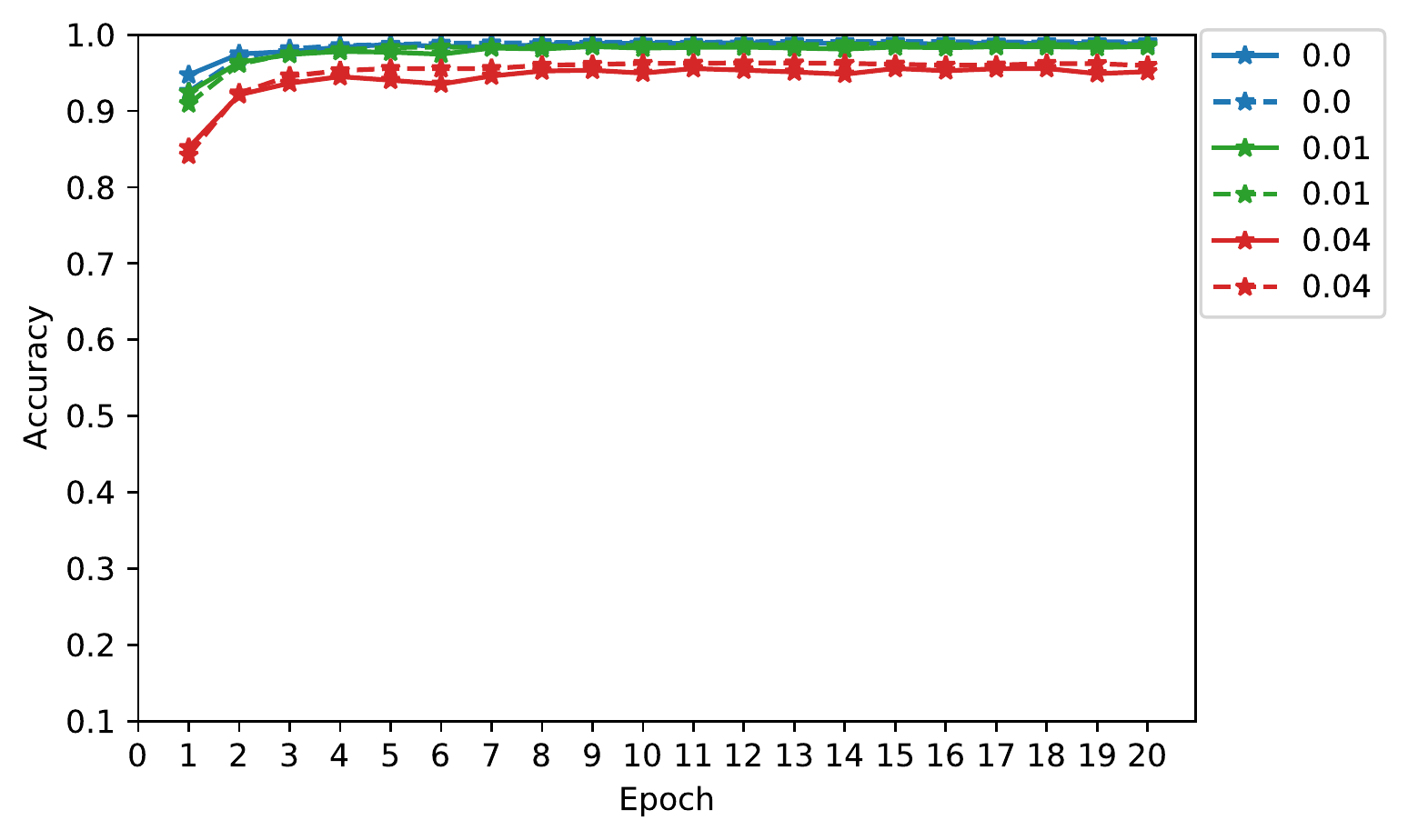}
\end{minipage}
}%

\subfigure[SGD (zoomed in)]{
\begin{minipage}[t]{0.333\linewidth}
\centering
\includegraphics[width=2.in]{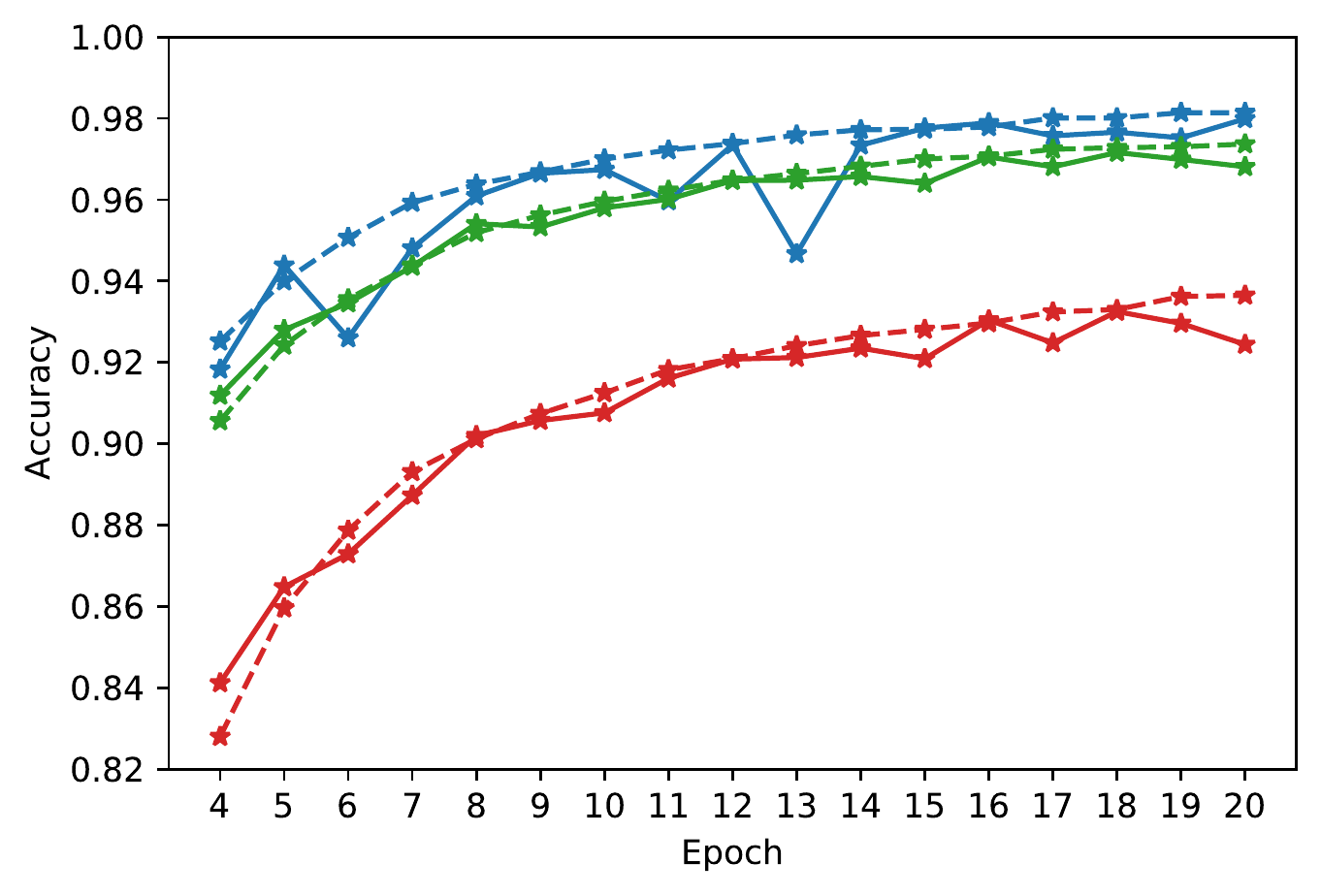}
\end{minipage}%
}%
\hspace{-0.5cm}
\subfigure[NAG (zoomed in)]{
\begin{minipage}[t]{0.333\linewidth}
\centering
\includegraphics[width=2.in]{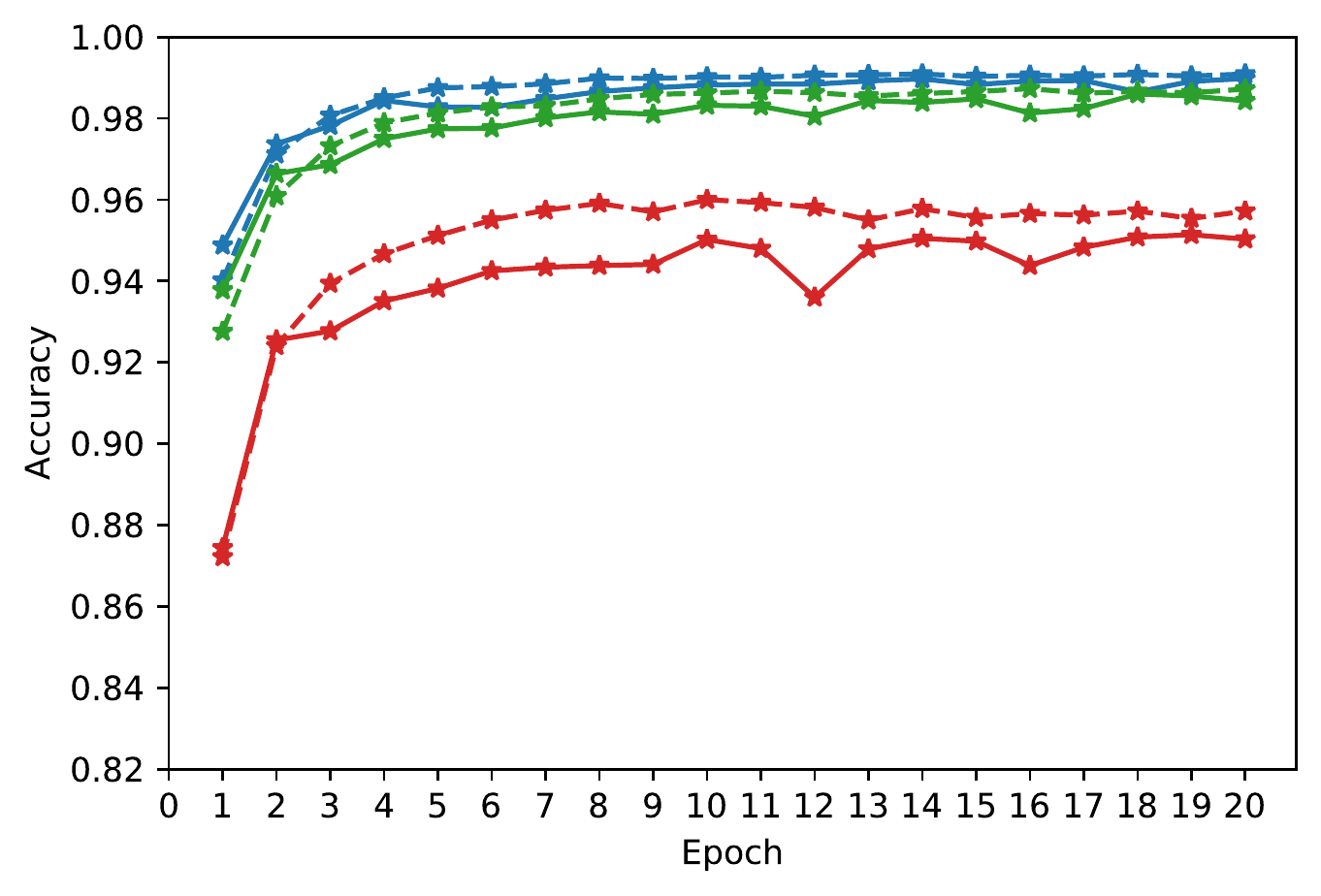}
\end{minipage}%
}%
\hspace{-0.4cm}
\subfigure[HB (zoomed in)]{
\begin{minipage}[t]{0.333\linewidth}
\centering
\includegraphics[width=2.3in]{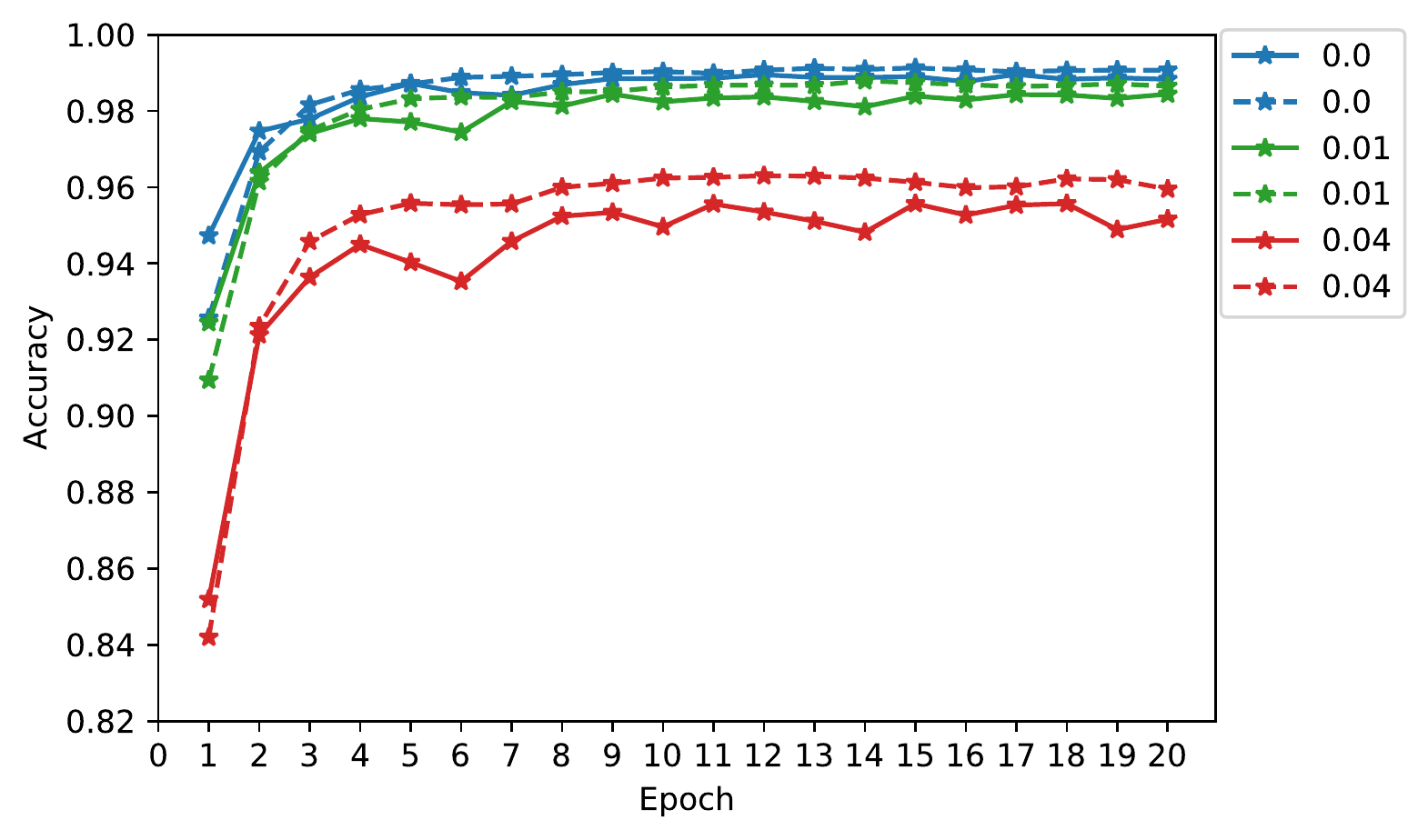}
\end{minipage}
}%
\caption{Adversarial accuracy of neural networks on MNIST under PGD, trained with different optimizers.}
\label{fig:MNIST PGD optimizer}
\end{figure}

 From \Cref{fig:MNIST FGSM optimizer} and \Cref{fig:MNIST PGD optimizer} we observe that, fixing the adversarial attack, the outer minimization optimizer does affect the adversarial robustness of neural networks. For example, when attacked by FGSM or PGD with $\epsilon=0.01$, the neural network learned by SGD, especially in the first two epochs, has a very low accuracy (from 18\% to 93\% at a slow growth rate) compared with the accuracy achieved by momentum methods, in which almost all the results are above 92\% accuracy (except for the first epoch, which is still much greater than the 18\% from the first epoch of SGD) and can be up to 96\% accuracy. On the other hand, our snapshot ensemble improves much more on momentum methods (roughly 5\%) than on SGD. In addition, different attacks seem to have small effects on the performance under varied $\epsilon$.

\subsubsection{Number of snapshots}

\begin{figure}[!htb]
\centering
\subfigure[98 snapshots]{
\begin{minipage}[t]{0.25\linewidth}
\centering
\includegraphics[width=1.5in]{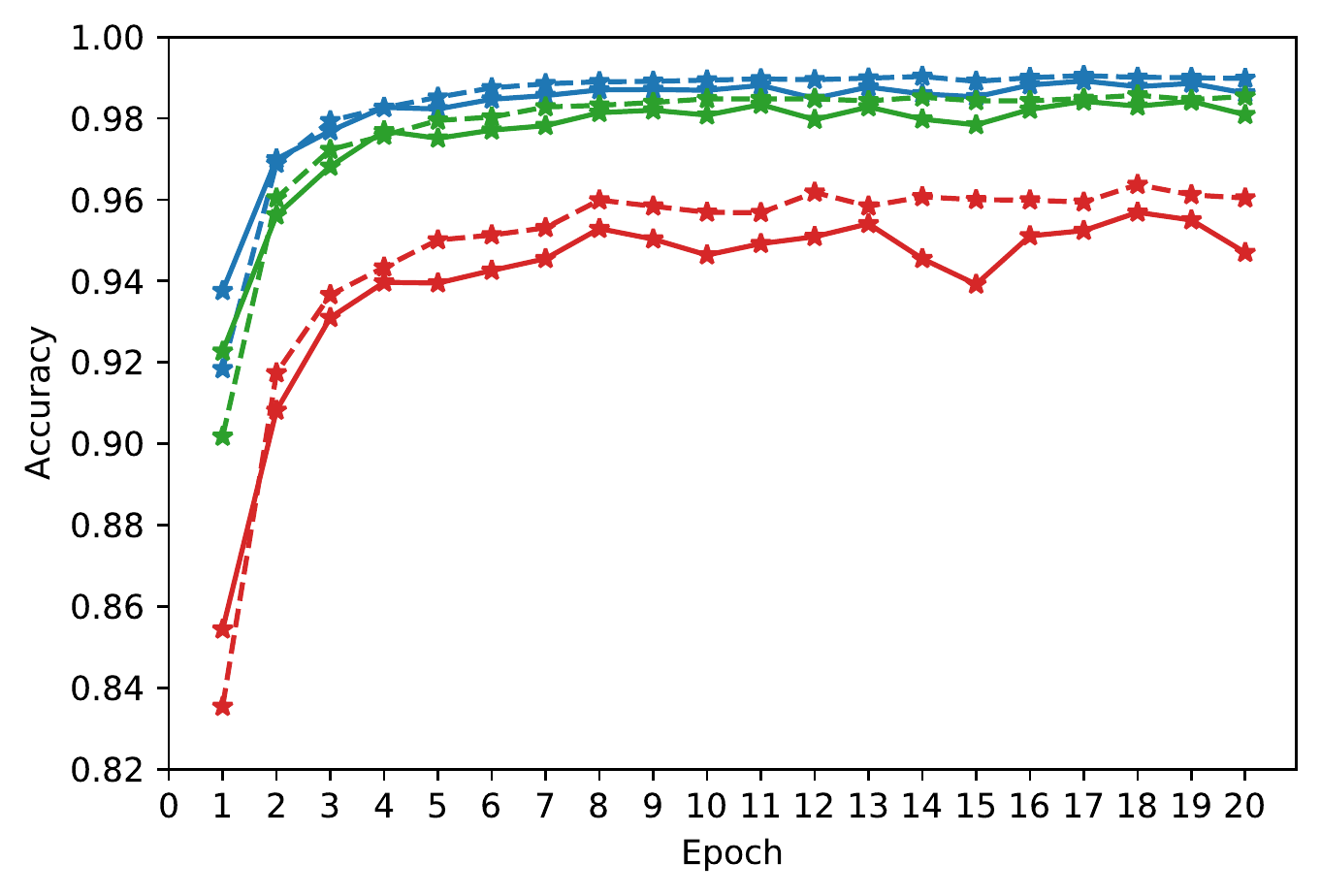}
\end{minipage}
}%
\hspace{-0.5cm}
\subfigure[40 snapshots]{
\begin{minipage}[t]{0.25\linewidth}
\centering
\includegraphics[width=1.5in]{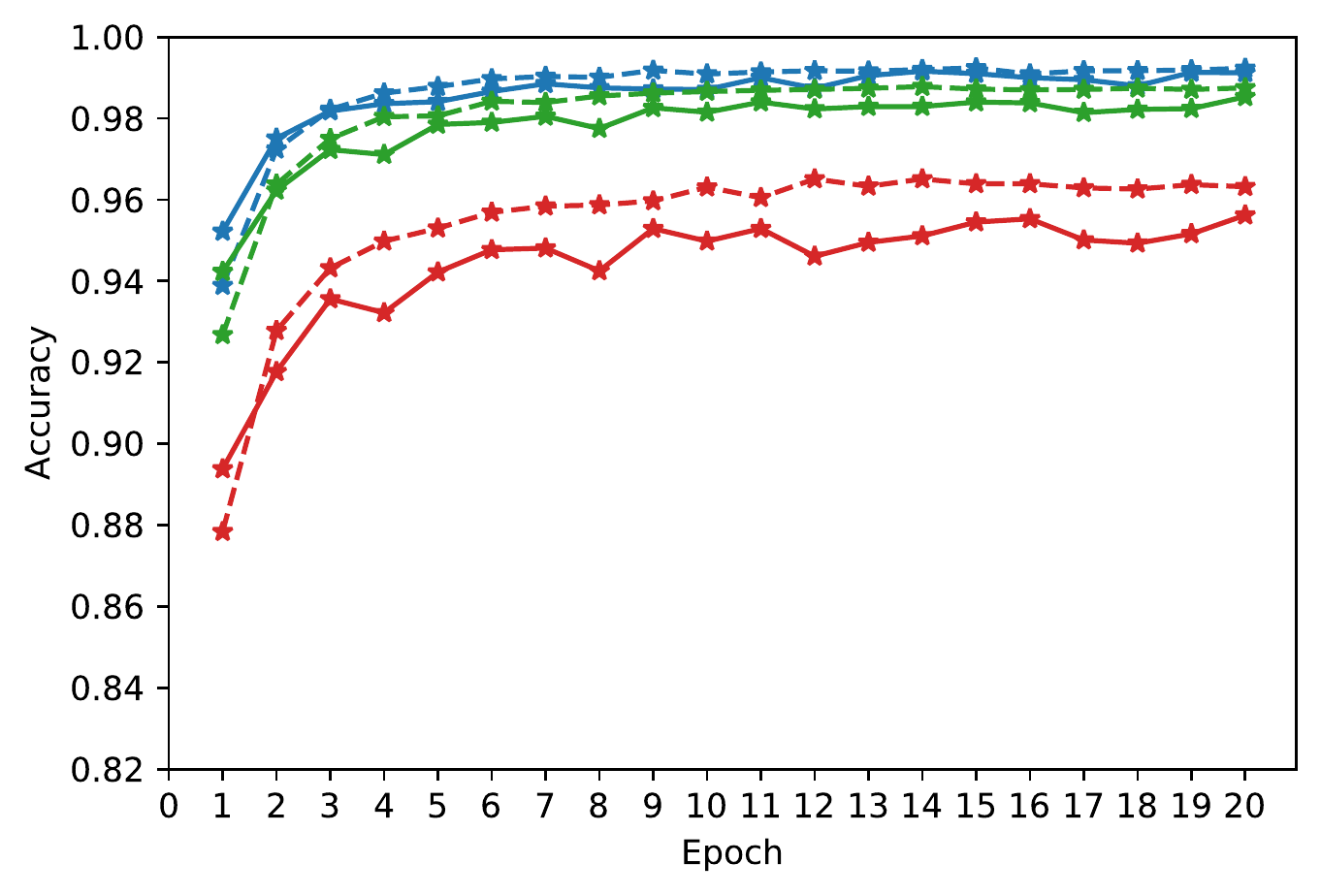}
\end{minipage}
}%
\hspace{-0.5cm}
\subfigure[20 snapshots]{
\begin{minipage}[t]{0.25\linewidth}
\centering
\includegraphics[width=1.5in]{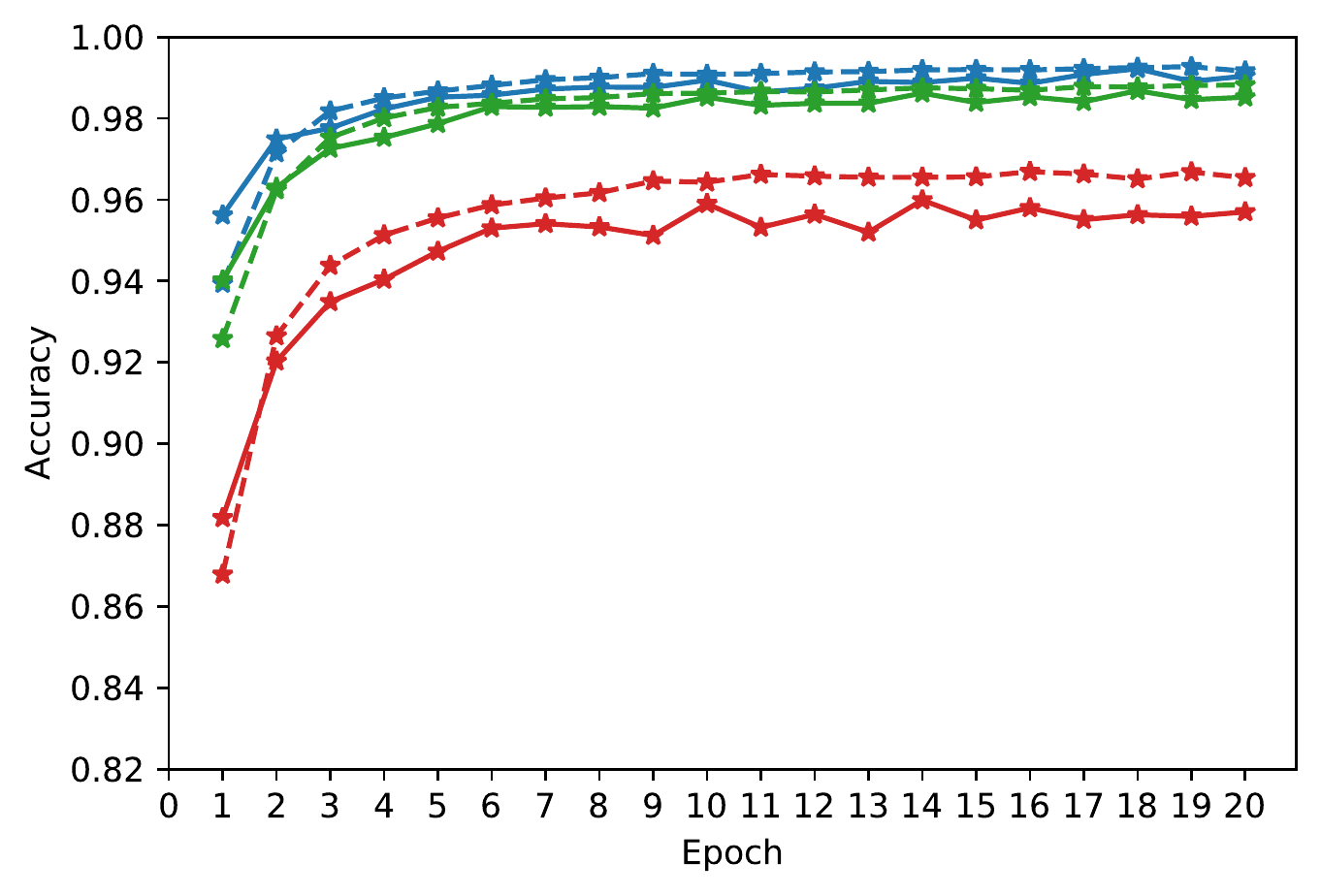}
\end{minipage}%
}%
\hspace{-0.2cm}
\subfigure[10 snapshots]{
\begin{minipage}[t]{0.25\linewidth}
\centering
\includegraphics[width=1.7in]{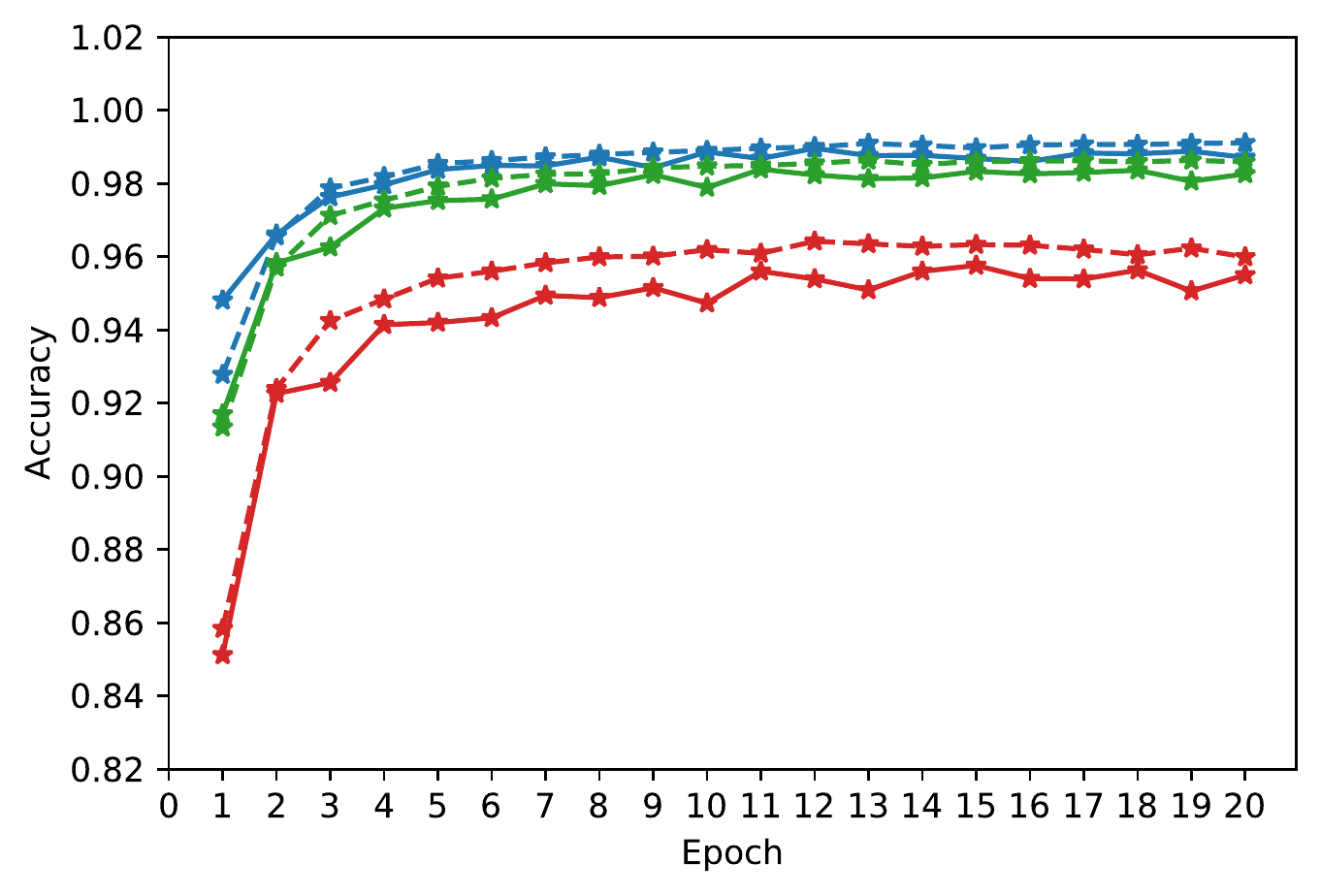}
\end{minipage}%
}%
\caption{ Adversarial accuracy of neural networks on MNIST under FGSM, ensembled with different snapshots.}
\label{fig:MNIST FGSM snapshots}
\end{figure}

\begin{figure}[!htb]
\centering
\subfigure[98 snapshots]{
\begin{minipage}[t]{0.25\linewidth}
\centering
\includegraphics[width=1.5in]{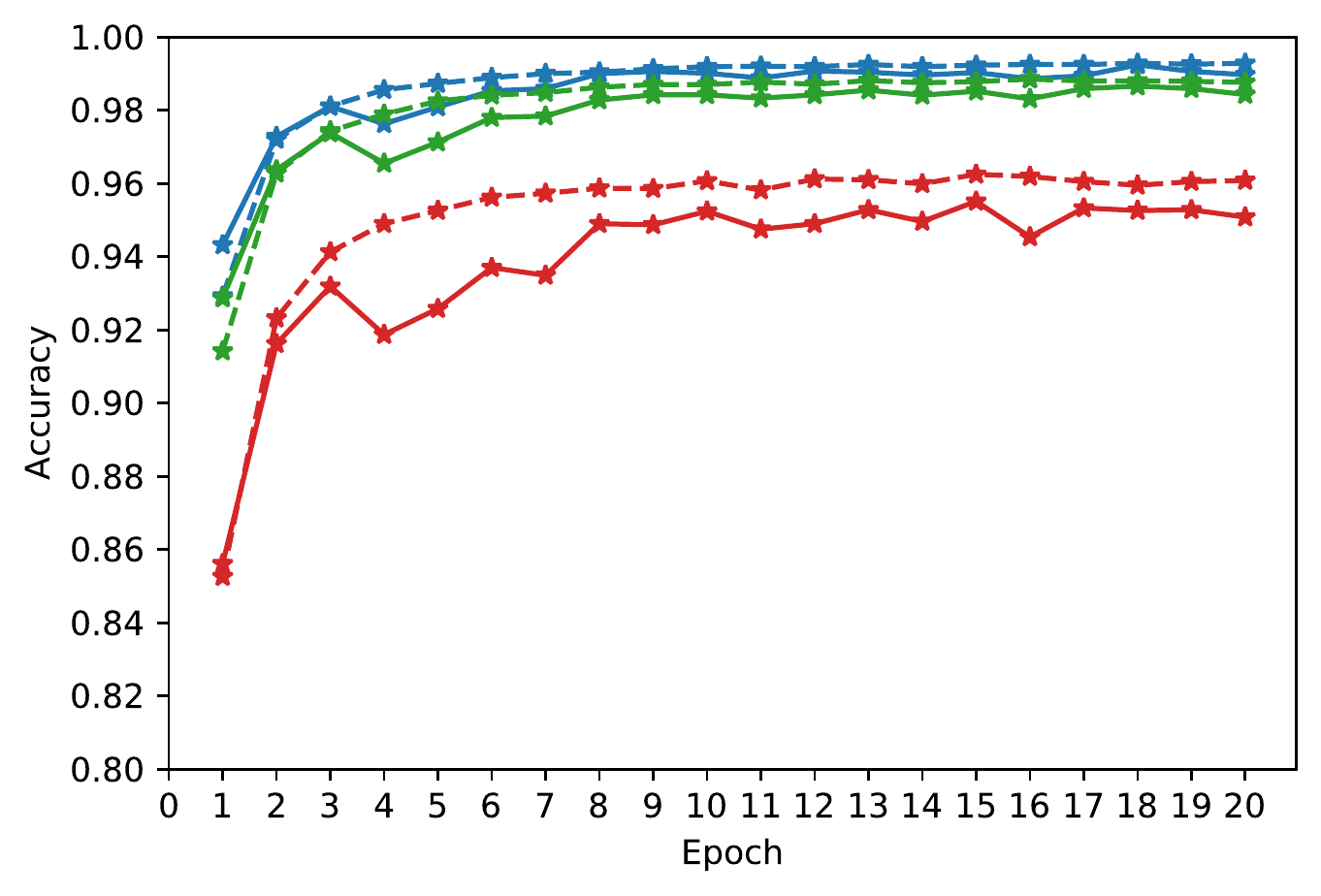}
\end{minipage}
}%
\hspace{-0.5cm}
\subfigure[40 snapshots]{
\begin{minipage}[t]{0.25\linewidth}
\centering
\includegraphics[width=1.5in]{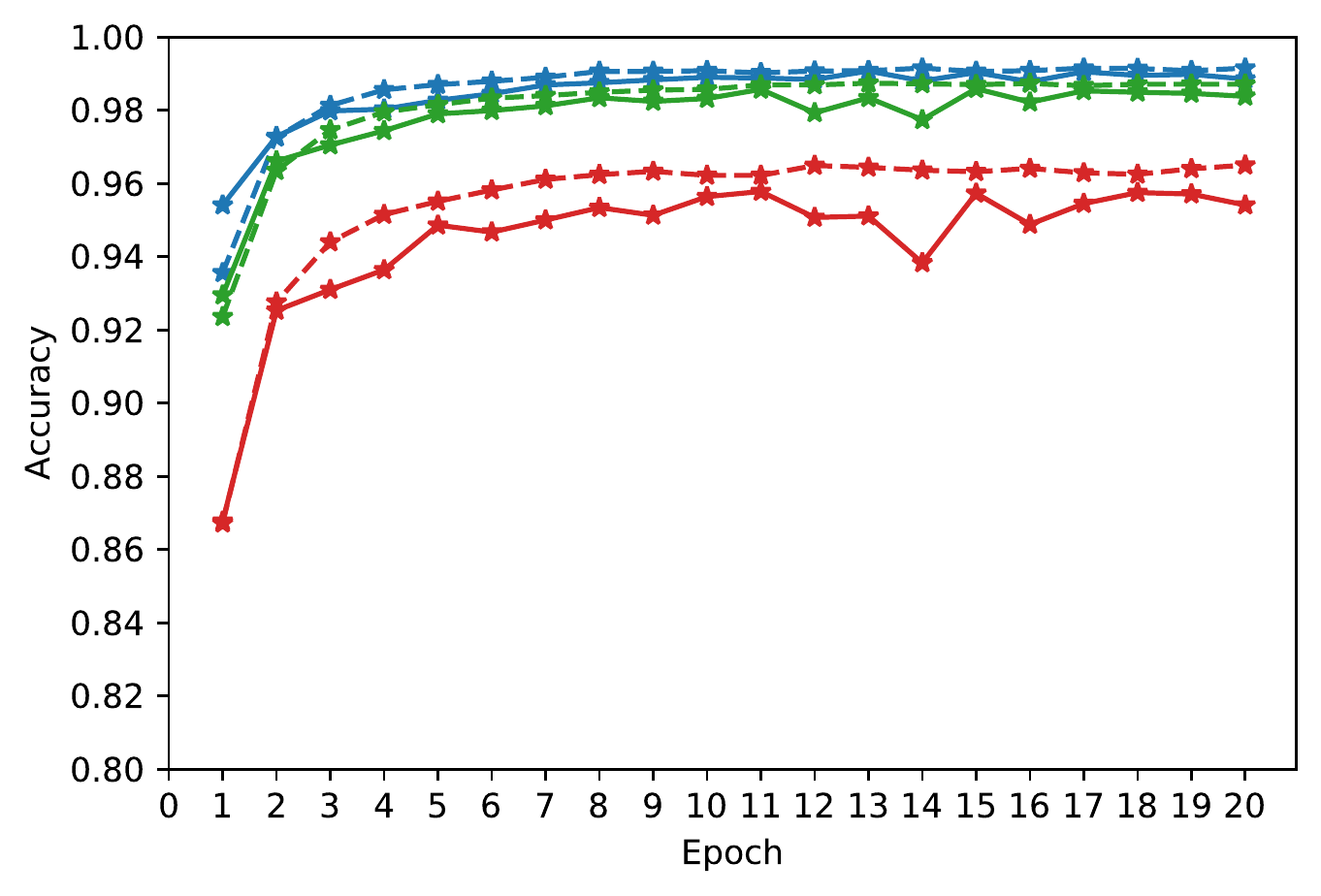}
\end{minipage}
}%
\hspace{-0.5cm}
\subfigure[20 snapshots]{
\begin{minipage}[t]{0.25\linewidth}
\centering
\includegraphics[width=1.5in]{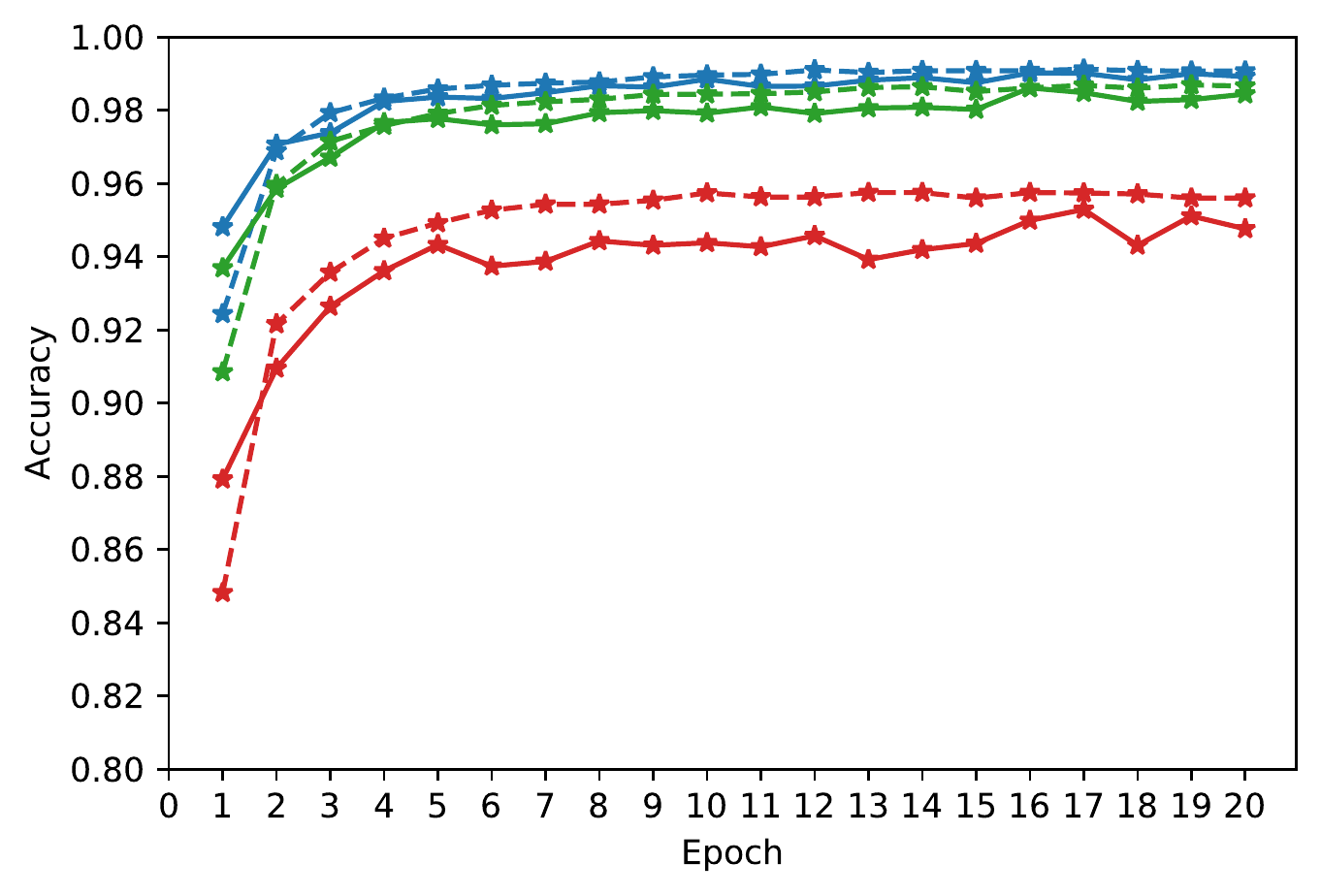}
\end{minipage}%
}%
\hspace{-0.4cm}
\subfigure[10 snapshots]{
\begin{minipage}[t]{0.25\linewidth}
\centering
\includegraphics[width=1.75in]{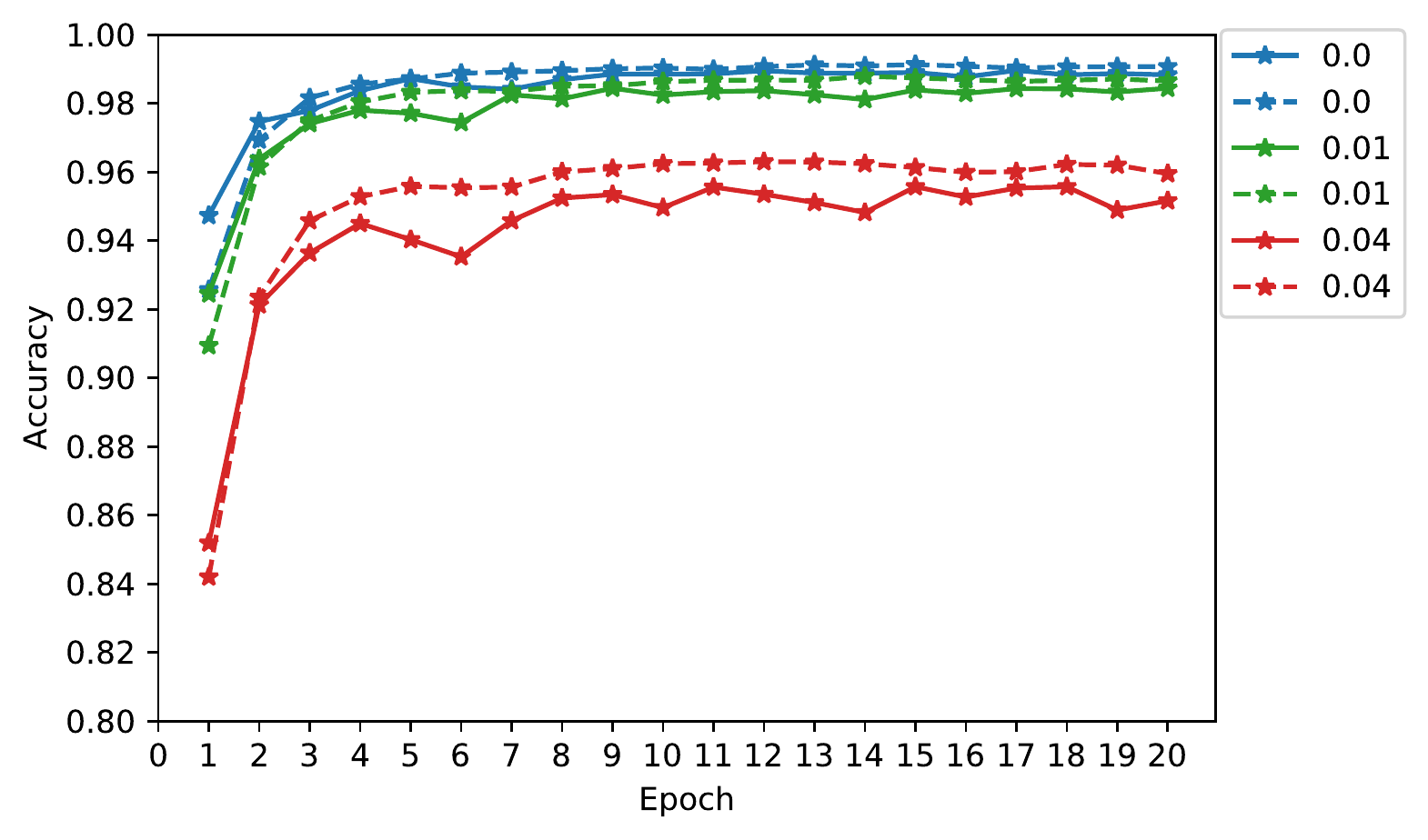}
\end{minipage}%
}%
\caption{ Adversarial accuracy of neural networks on MNIST under PGD, ensembled with different snapshots.}
\label{fig:MNIST PGD snapshots}
\end{figure}
 From \Cref{fig:MNIST FGSM snapshots} and \Cref{fig:MNIST PGD snapshots} we can see that the number of snapshots has the same ignorable influence on the adversarial accuracy in MNIST dataset. Because more snapshots consume more time spent on ensembling (in other words, prediction time) and more memory to store the historical weights from past iterations, our experiments suggest that using a small number of snapshots would be sufficient in practice.

\subsubsection{Learning rate}

\begin{figure}[!htb]
\centering
\subfigure[0.002 learning rate]{
\begin{minipage}[t]{0.25\linewidth}
\centering
\includegraphics[width=1.5in]{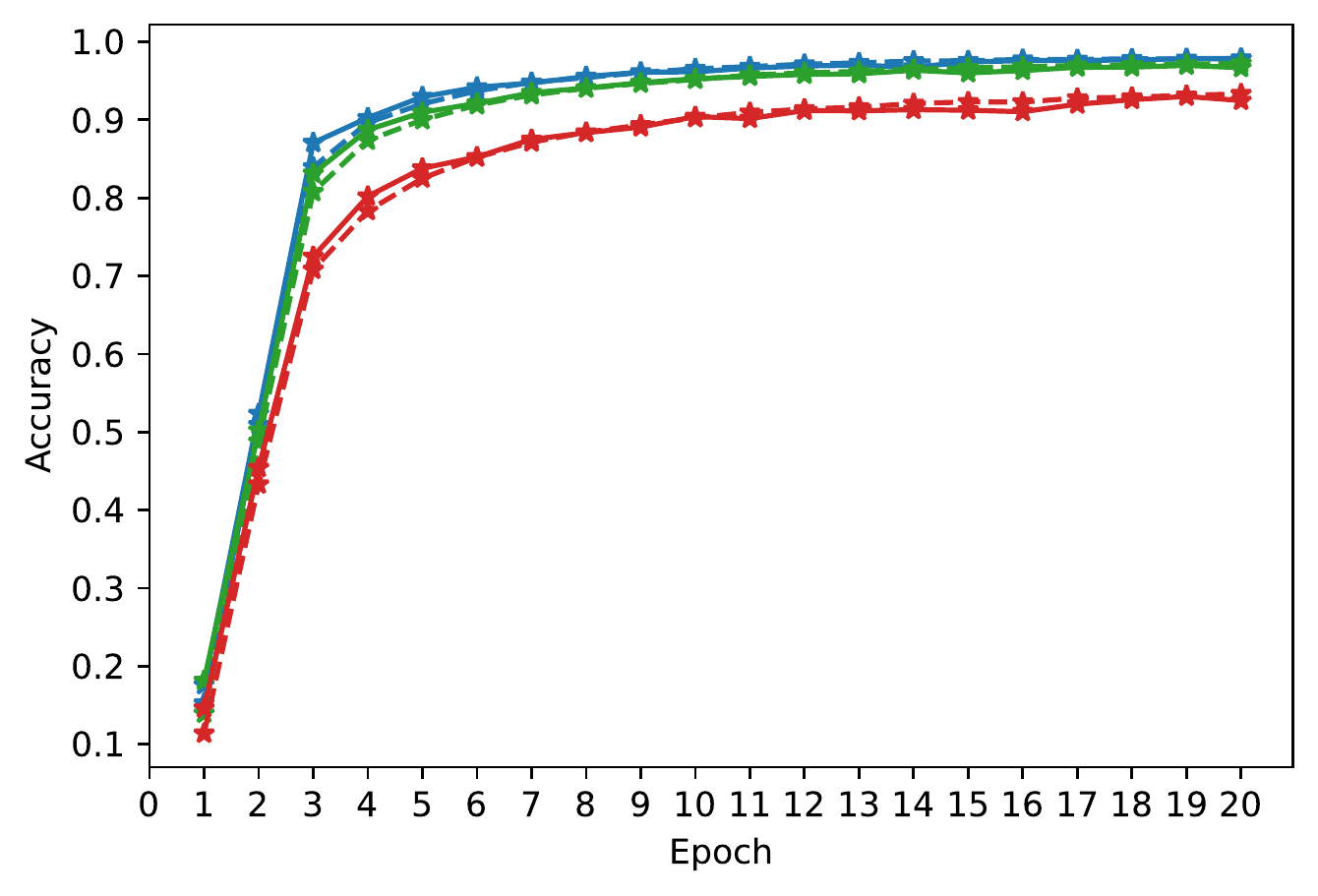}
\end{minipage}%
}%
\hspace{-0.5cm}
\subfigure[0.005 learning rate]{
\begin{minipage}[t]{0.25\linewidth}
\centering
\includegraphics[width=1.5in]{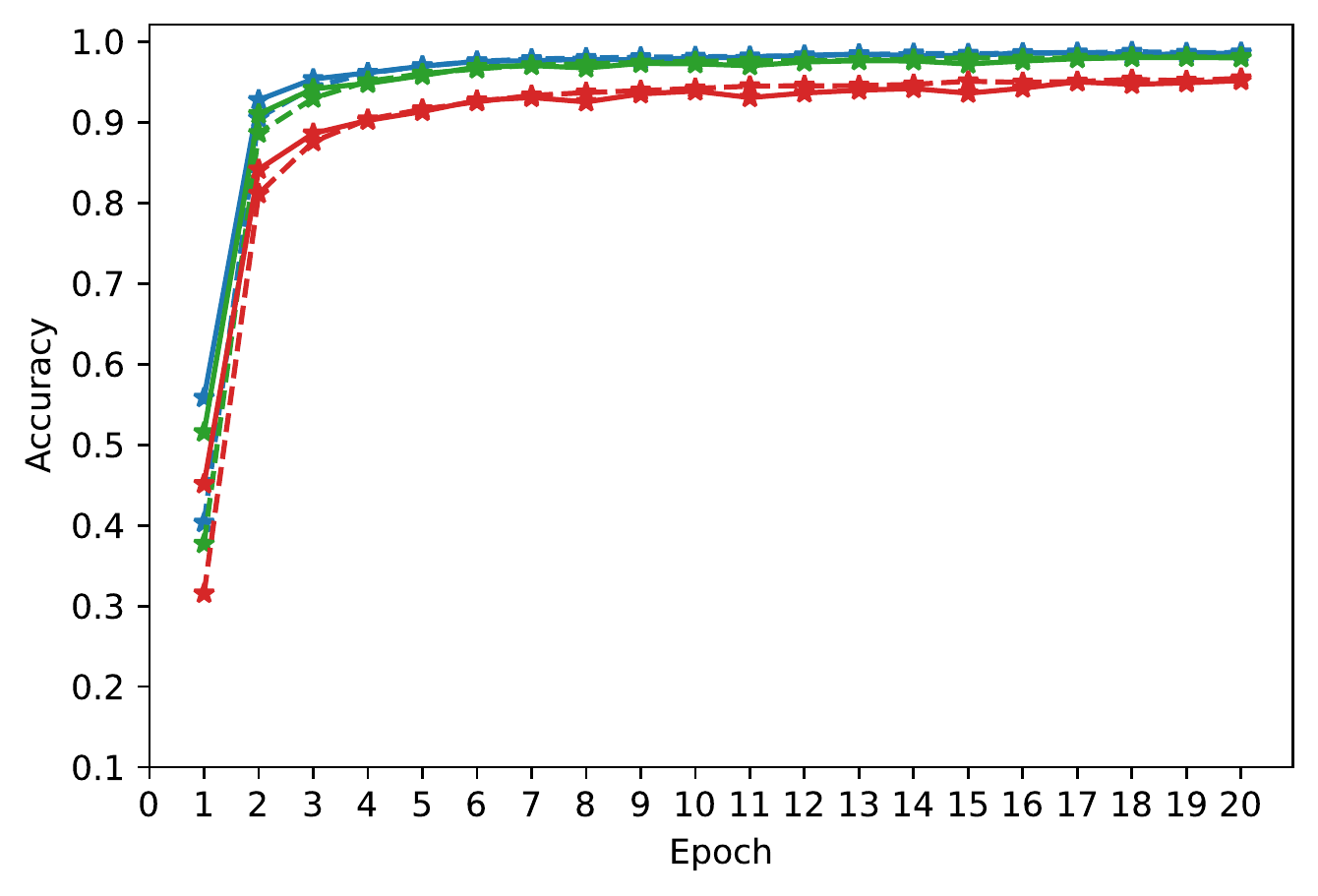}
\end{minipage}%
}%
\hspace{-0.5cm}
\subfigure[0.01 learning rate]{
\begin{minipage}[t]{0.25\linewidth}
\centering
\includegraphics[width=1.5in]{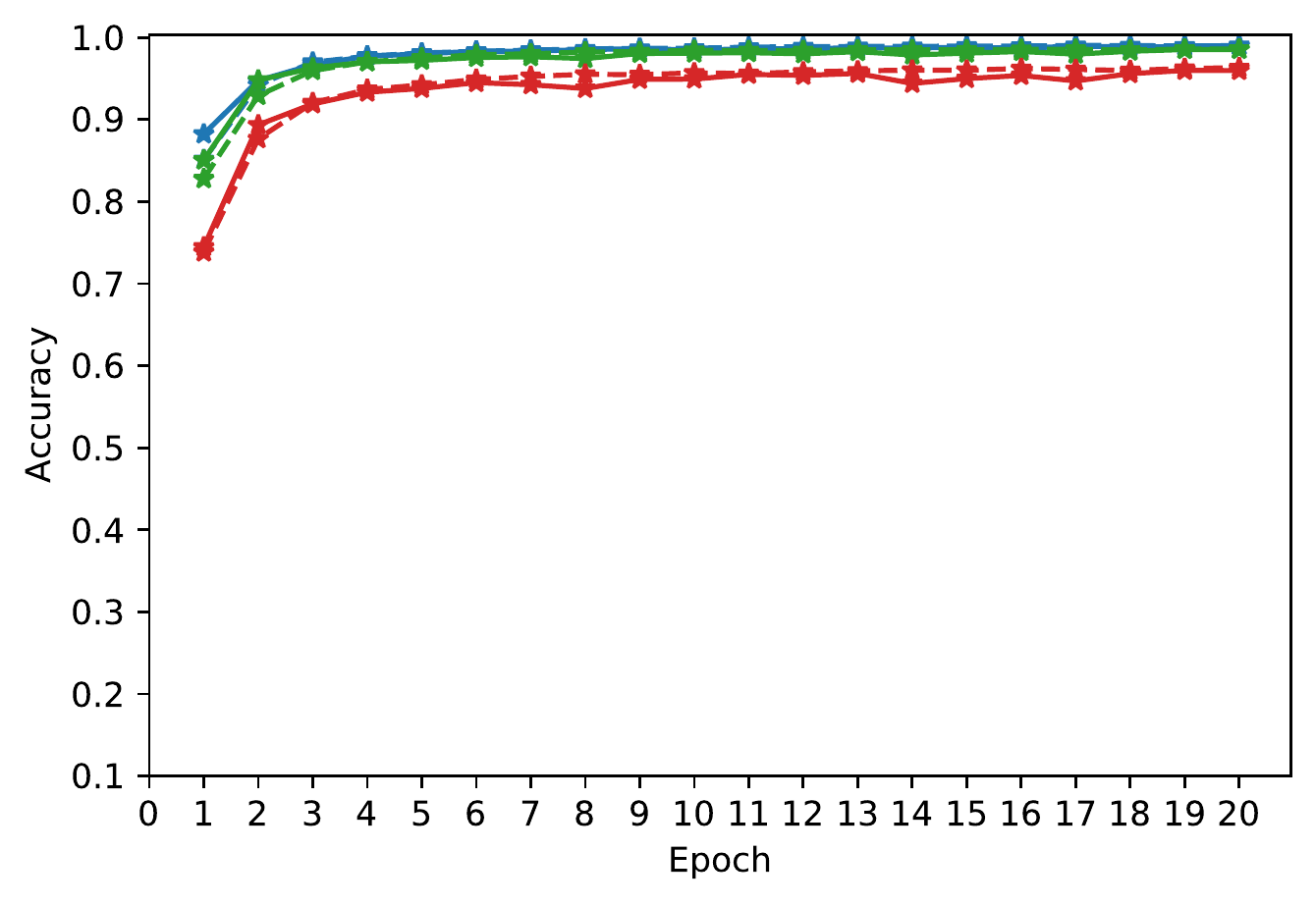}
\end{minipage}
}%
\hspace{-0.4cm}
\subfigure[0.02 learning rate]{
\begin{minipage}[t]{0.25\linewidth}
\centering
\includegraphics[width=1.75in]{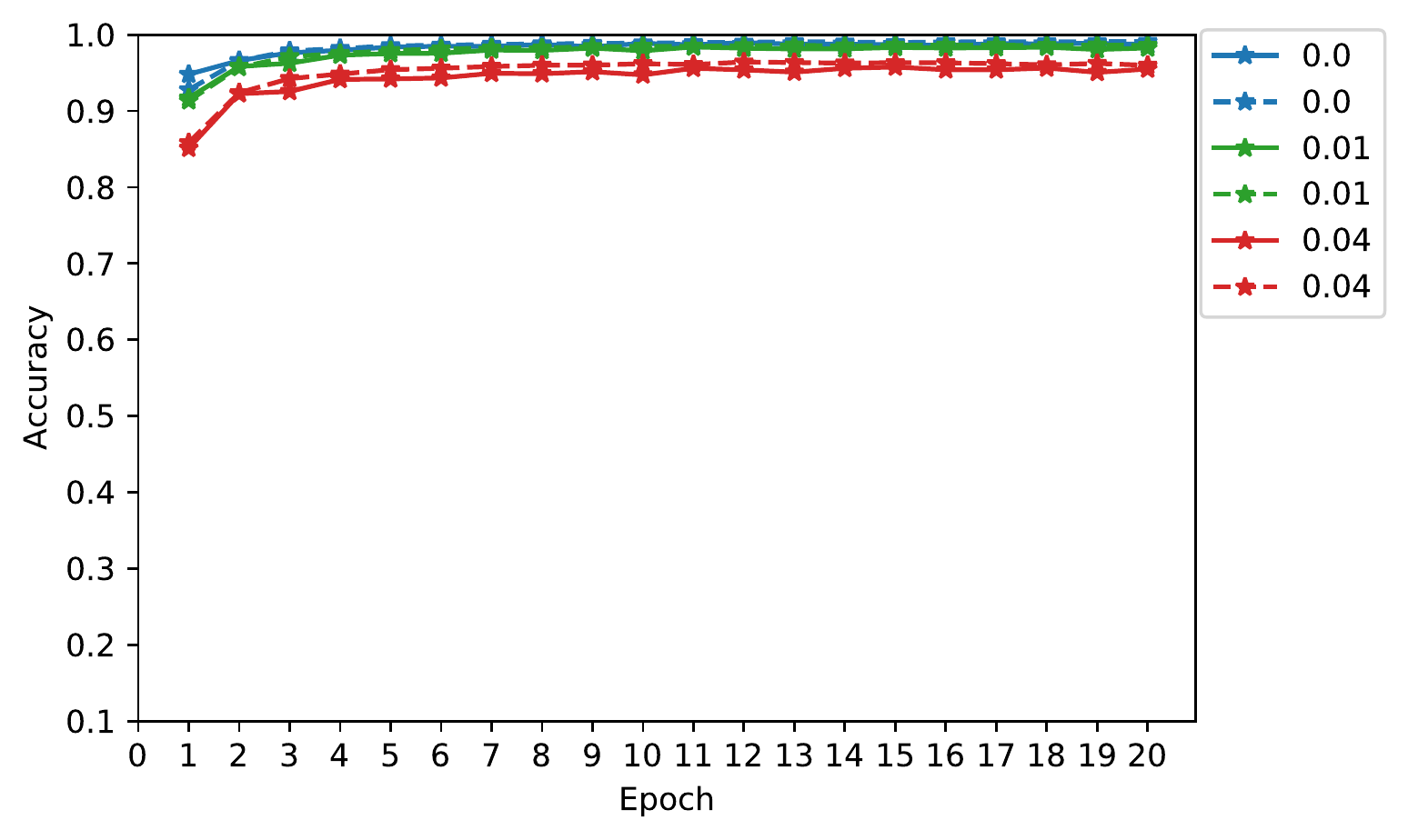}
\end{minipage}
}%

\subfigure[0.002 lr (zoomed in)]{
\begin{minipage}[t]{0.25\linewidth}
\centering
\includegraphics[width=1.5in]{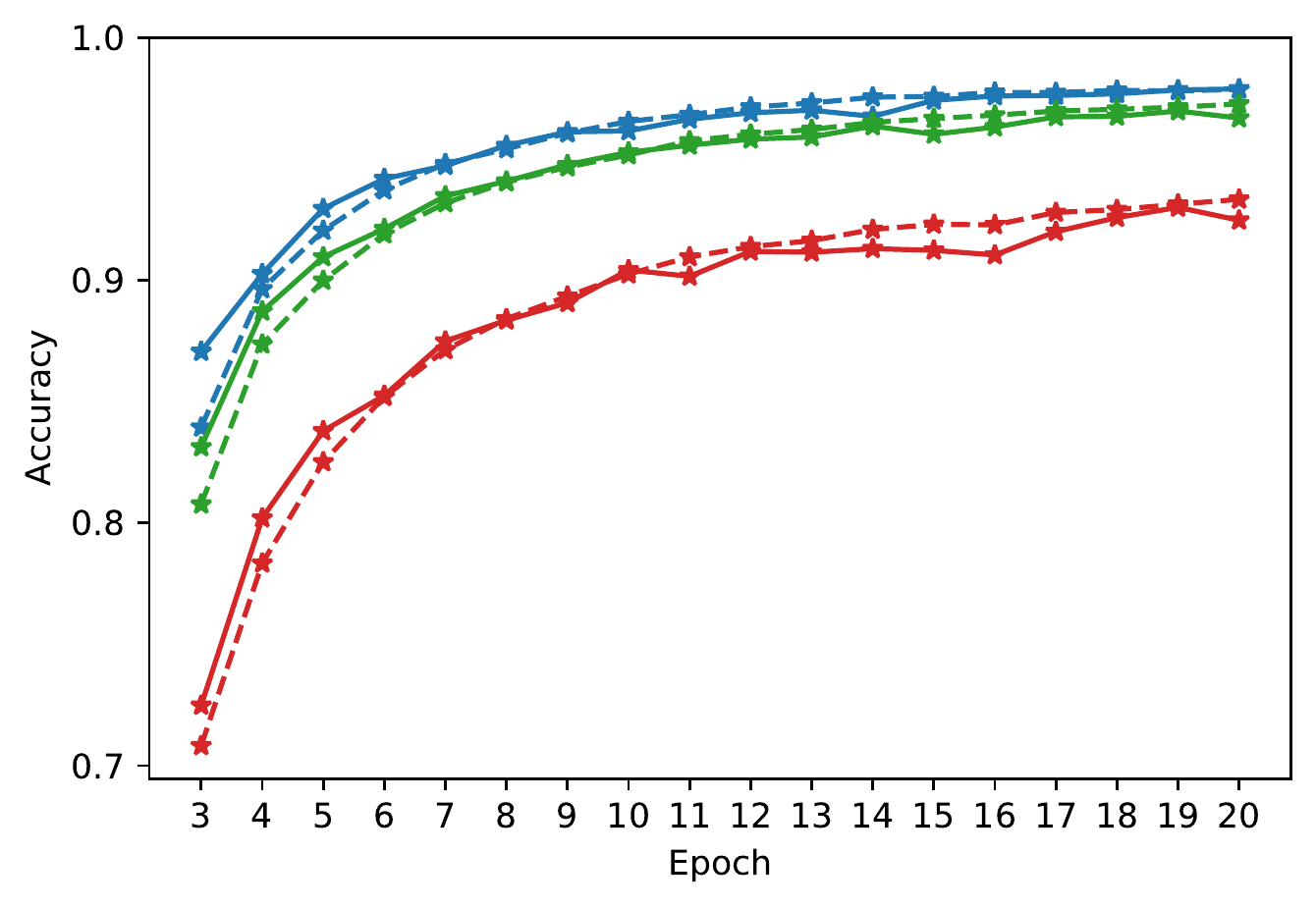}
\end{minipage}%
}%
\hspace{-0.5cm}
\subfigure[0.005 lr (zoomed in)]{
\begin{minipage}[t]{0.25\linewidth}
\centering
\includegraphics[width=1.5in]{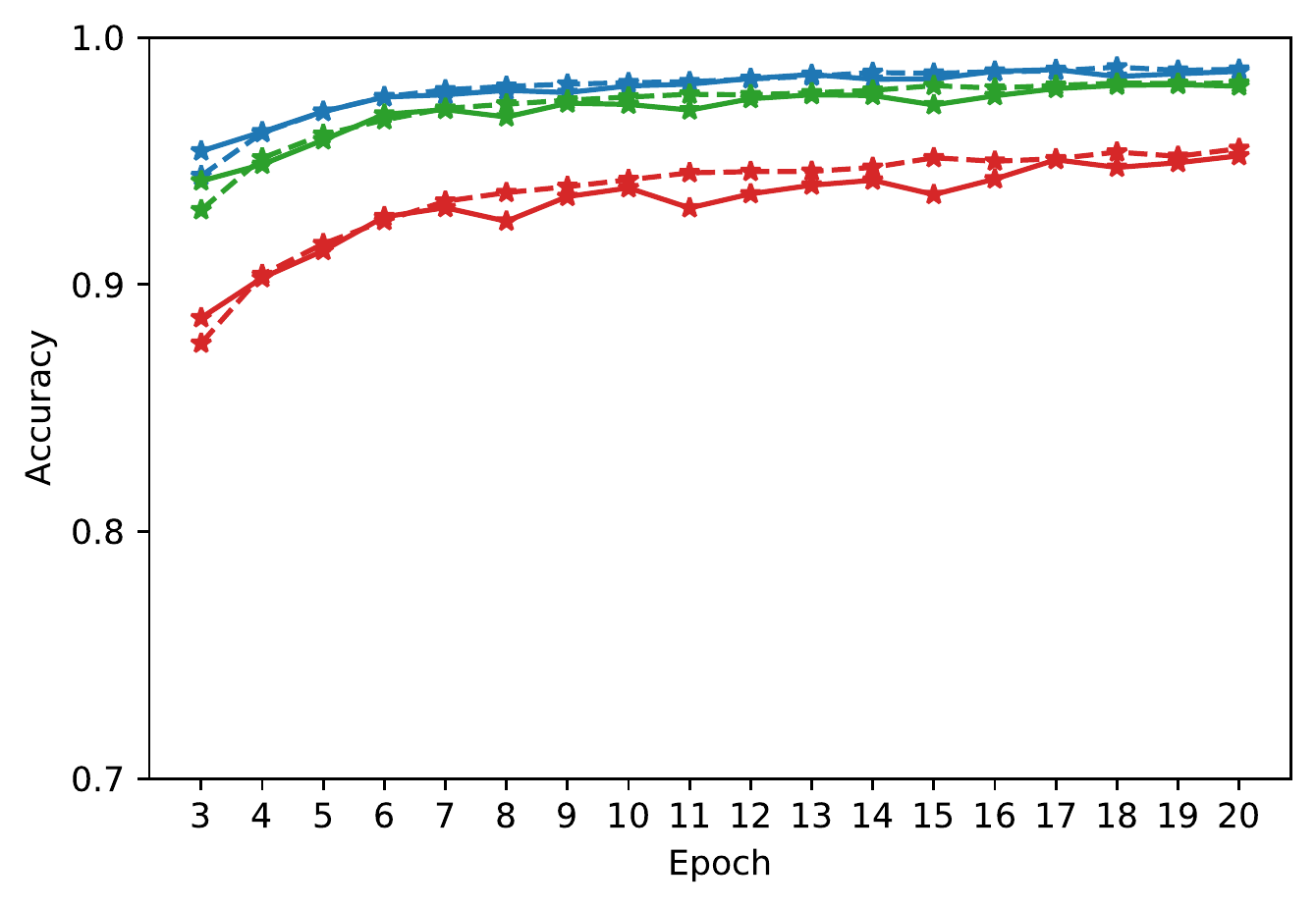}
\end{minipage}%
}%
\hspace{-0.5cm}
\subfigure[0.01 lr (zoomed in)]{
\begin{minipage}[t]{0.25\linewidth}
\centering
\includegraphics[width=1.5in]{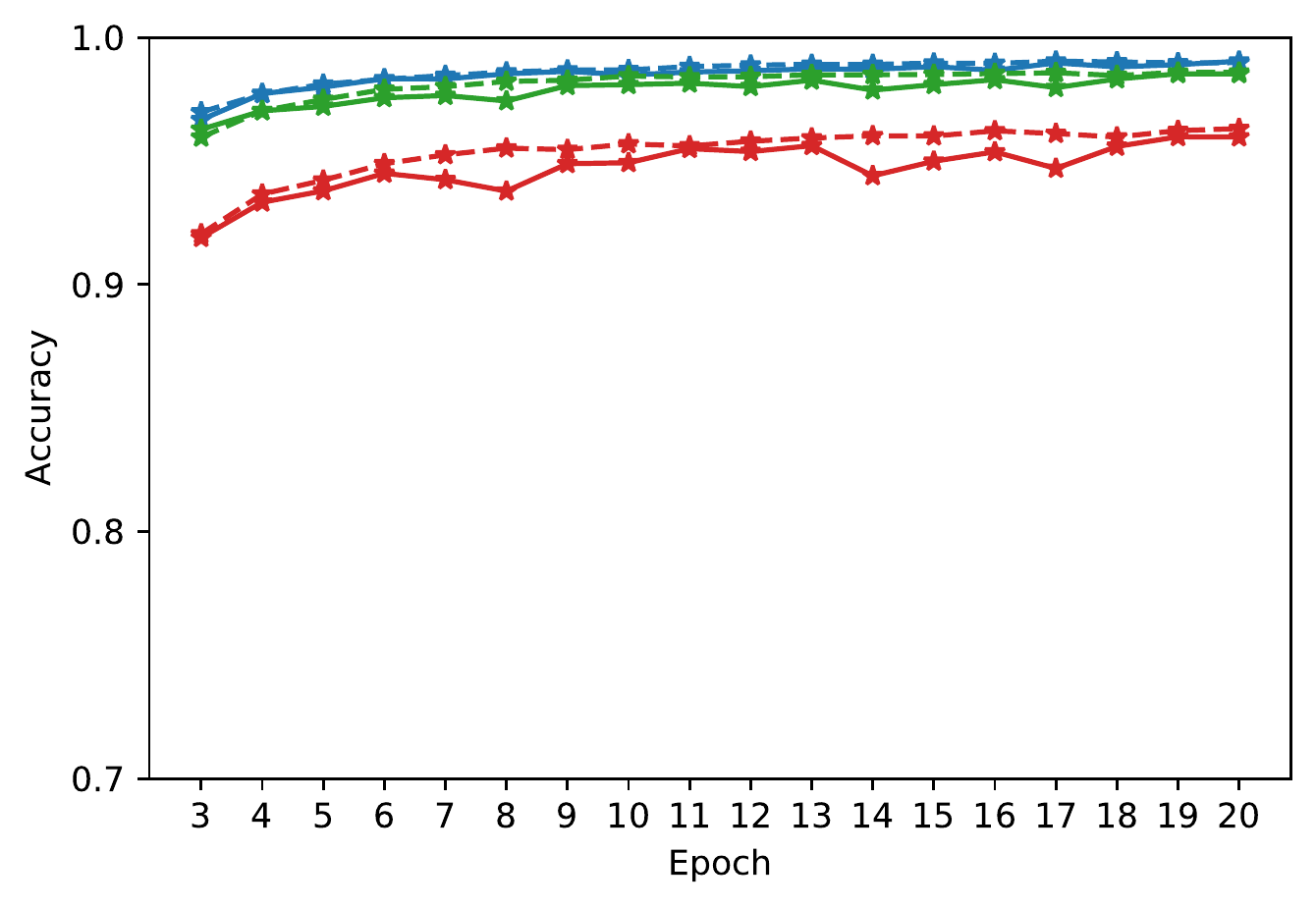}
\end{minipage}
}%
\hspace{-0.4cm}
\subfigure[0.02 lr (zoomed in)]{
\begin{minipage}[t]{0.25\linewidth}
\centering
\includegraphics[width=1.75in]{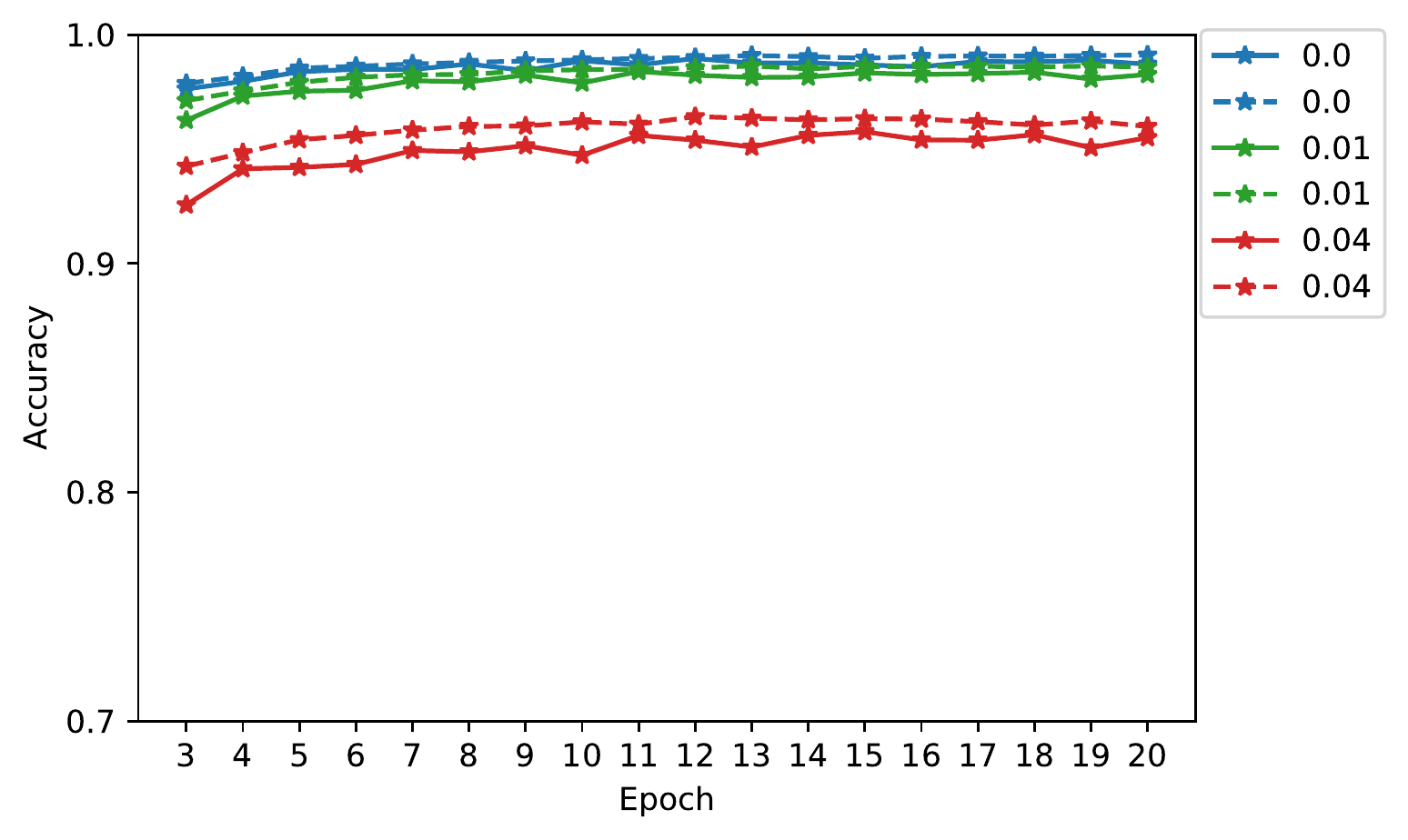}
\end{minipage}
}%
\centering
\caption{Adversarial accuracy of neural networks on MNIST under FGSM, trained with different learning rates.}
\label{fig:MNIST FGSM lr}
\end{figure}

\begin{figure}[!htb]
\centering
\subfigure[0.002 learning rate]{
\begin{minipage}[t]{0.25\linewidth}
\centering
\includegraphics[width=1.5in]{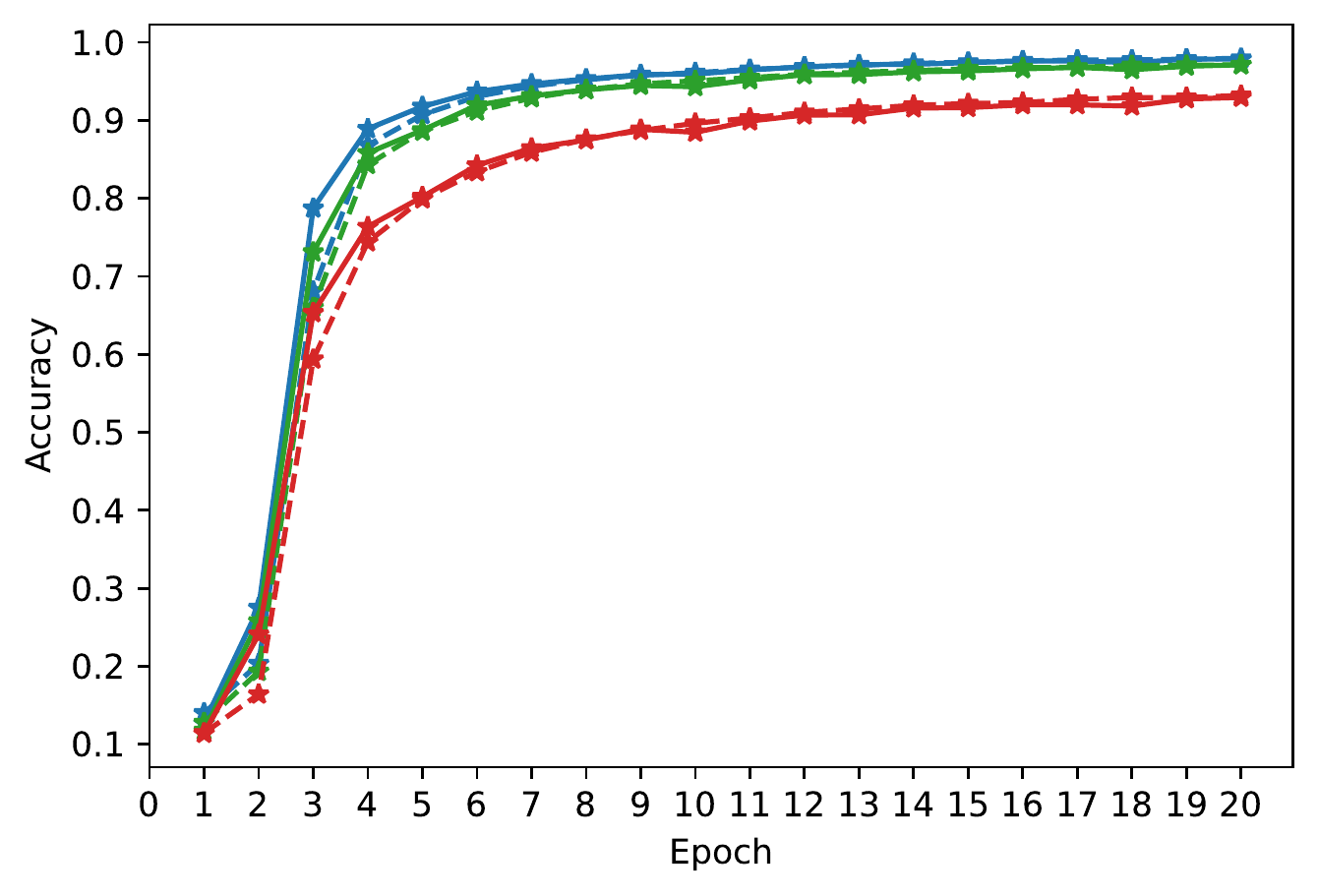}
\end{minipage}
}%
\hspace{-0.5cm}
\subfigure[0.005 learning rate]{
\begin{minipage}[t]{0.25\linewidth}
\centering
\includegraphics[width=1.5in]{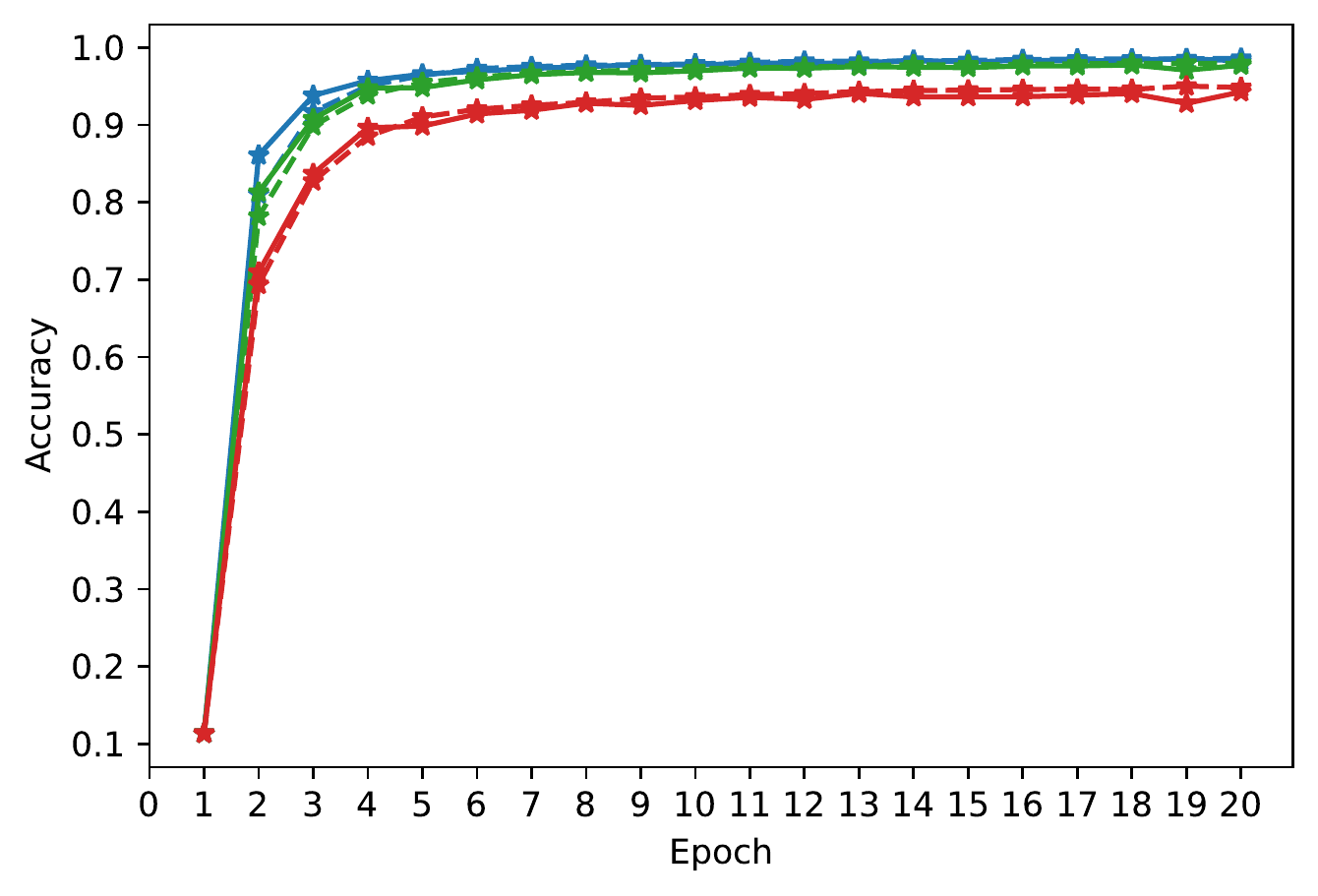}
\end{minipage}
}%
\hspace{-0.5cm}
\subfigure[0.01 learning rate]{
\begin{minipage}[t]{0.25\linewidth}
\centering
\includegraphics[width=1.5in]{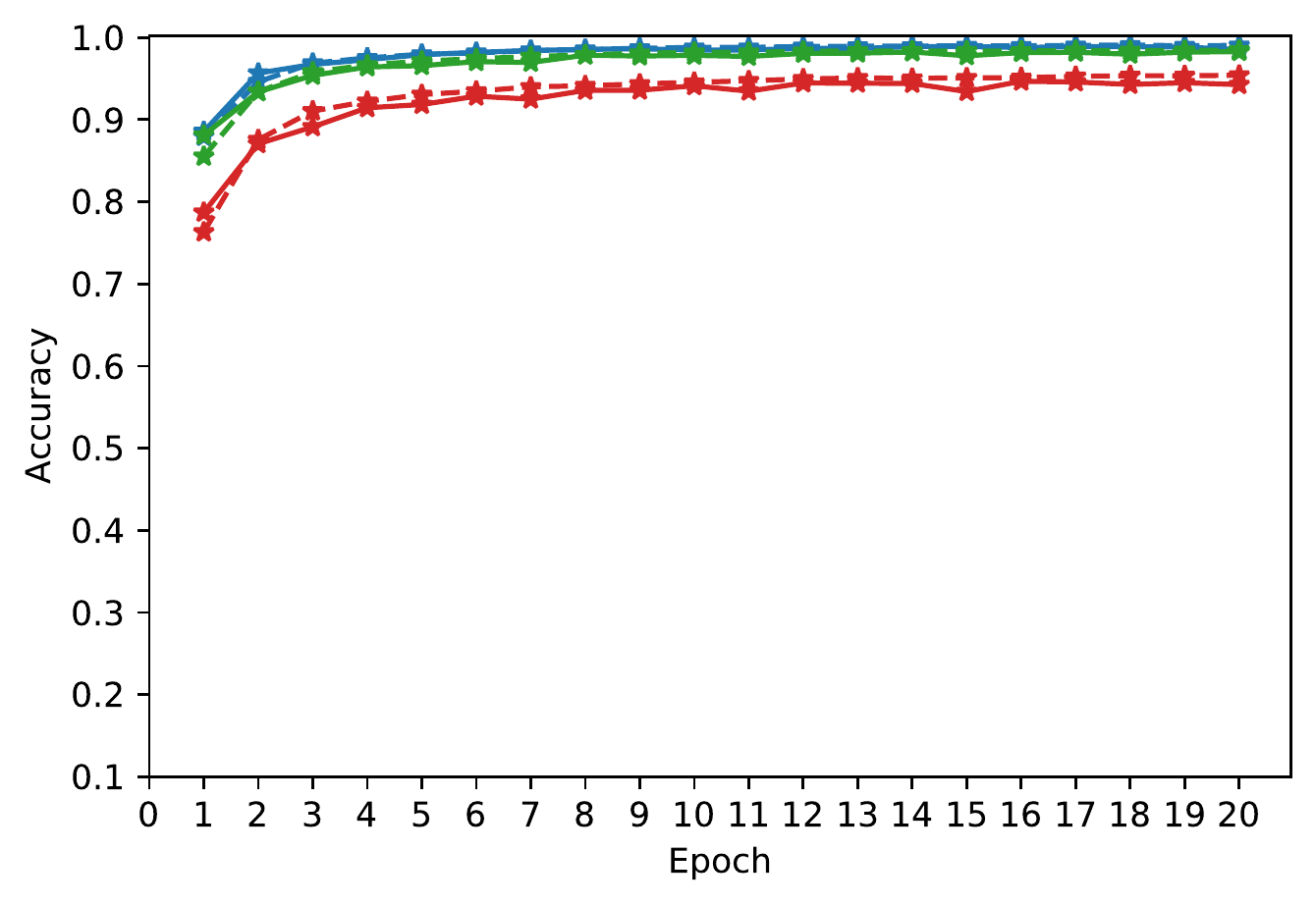}
\end{minipage}
}%
\hspace{-0.4cm}
\subfigure[0.02 learning rate]{
\begin{minipage}[t]{0.25\linewidth}
\centering
\includegraphics[width=1.75in]{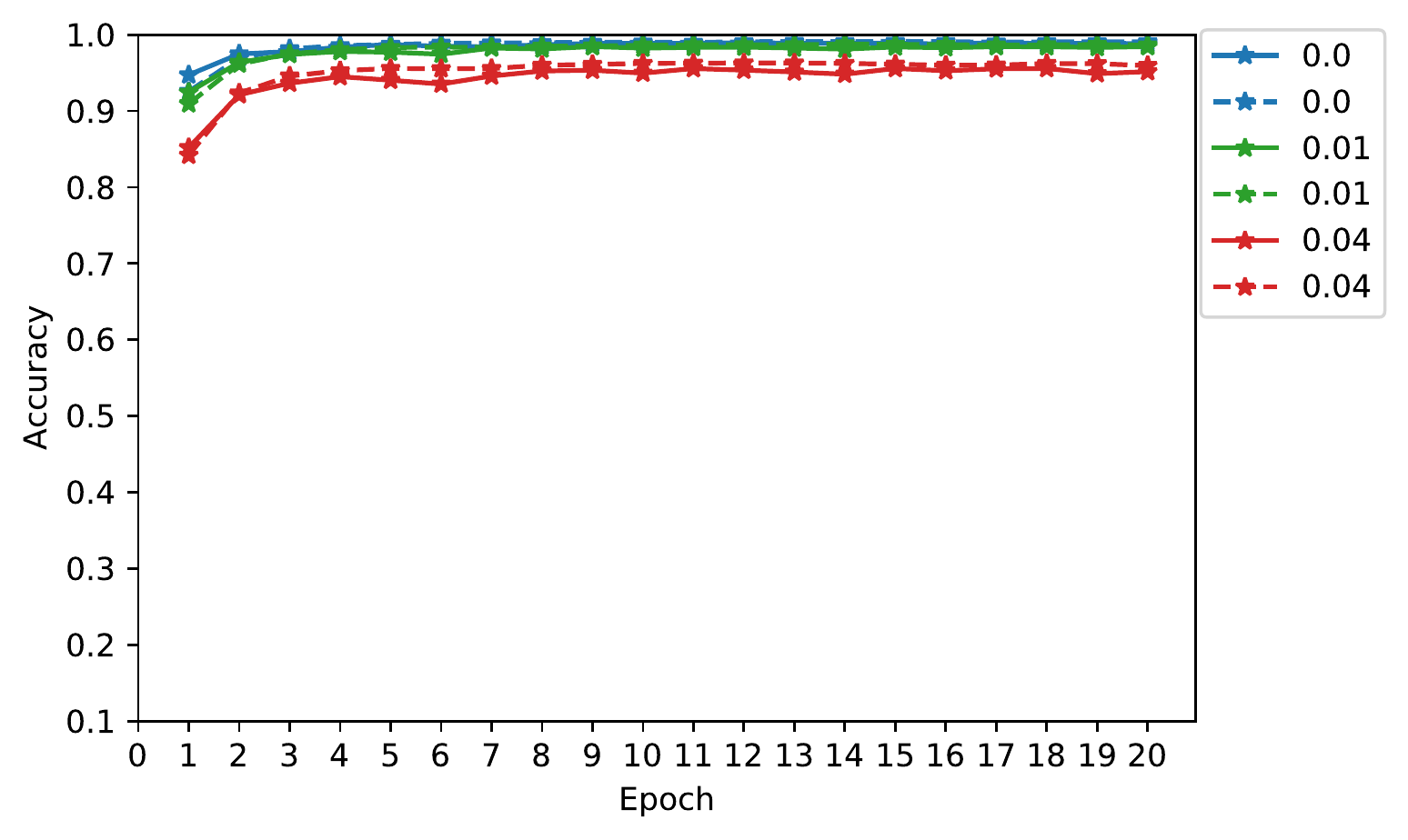}
\end{minipage}
}%

\subfigure[0.002 lr (zoomed in)]{
\begin{minipage}[t]{0.25\linewidth}
\centering
\includegraphics[width=1.5in]{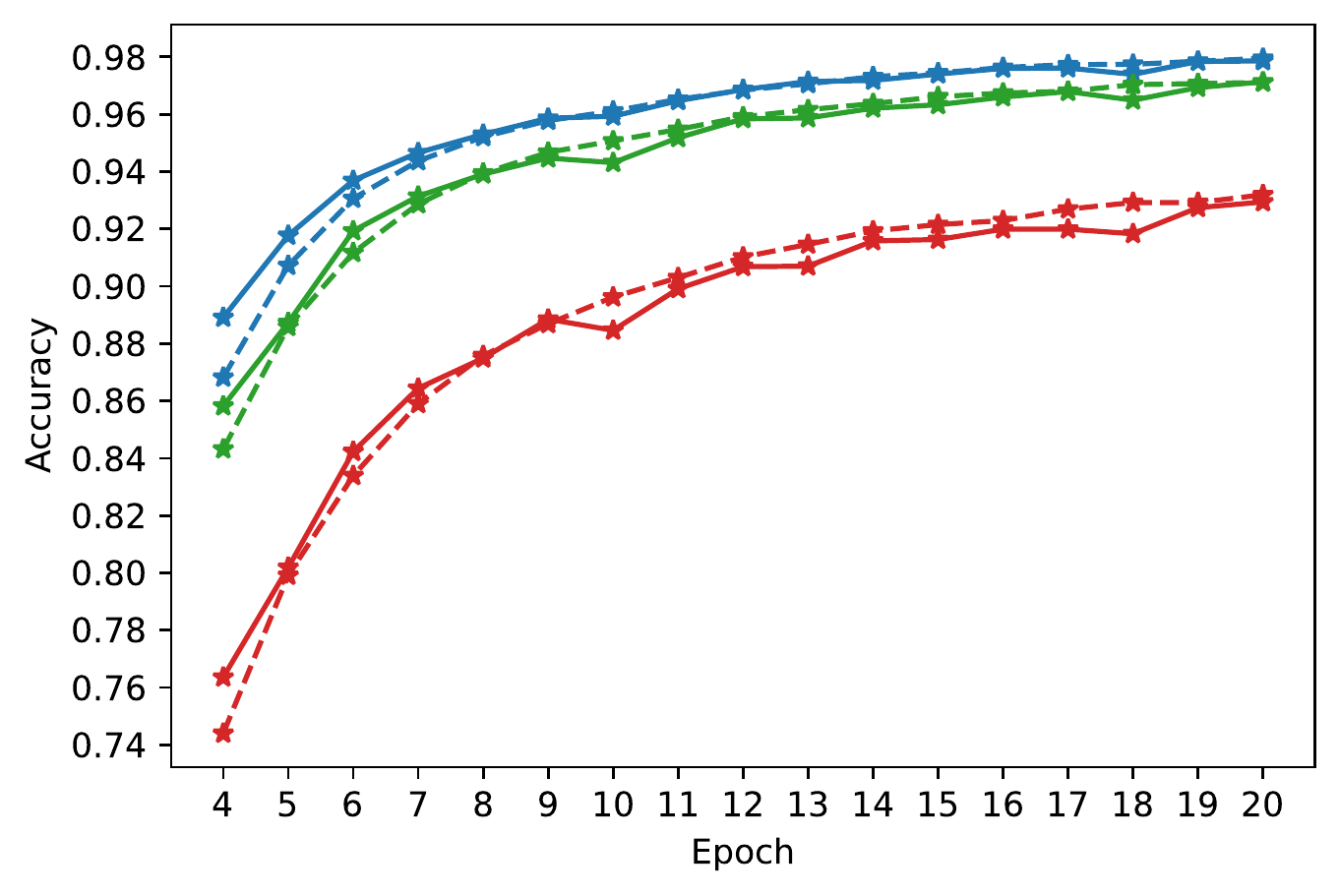}
\end{minipage}
}%
\hspace{-0.5cm}
\subfigure[0.005 lr (zoomed in)]{
\begin{minipage}[t]{0.25\linewidth}
\centering
\includegraphics[width=1.5in]{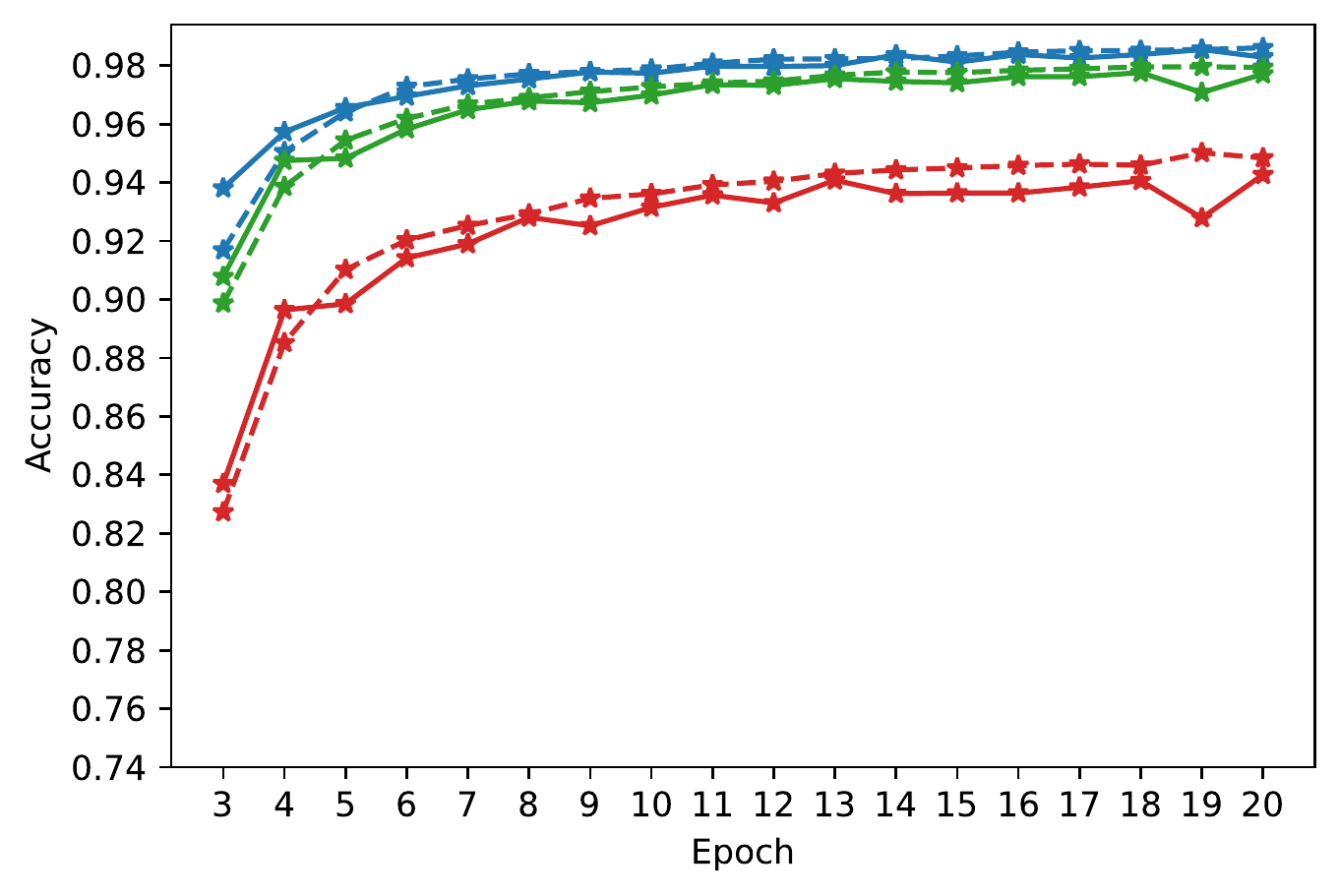}
\end{minipage}
}%
\hspace{-0.5cm}
\subfigure[0.01 lr (zoomed in)]{
\begin{minipage}[t]{0.25\linewidth}
\centering
\includegraphics[width=1.5in]{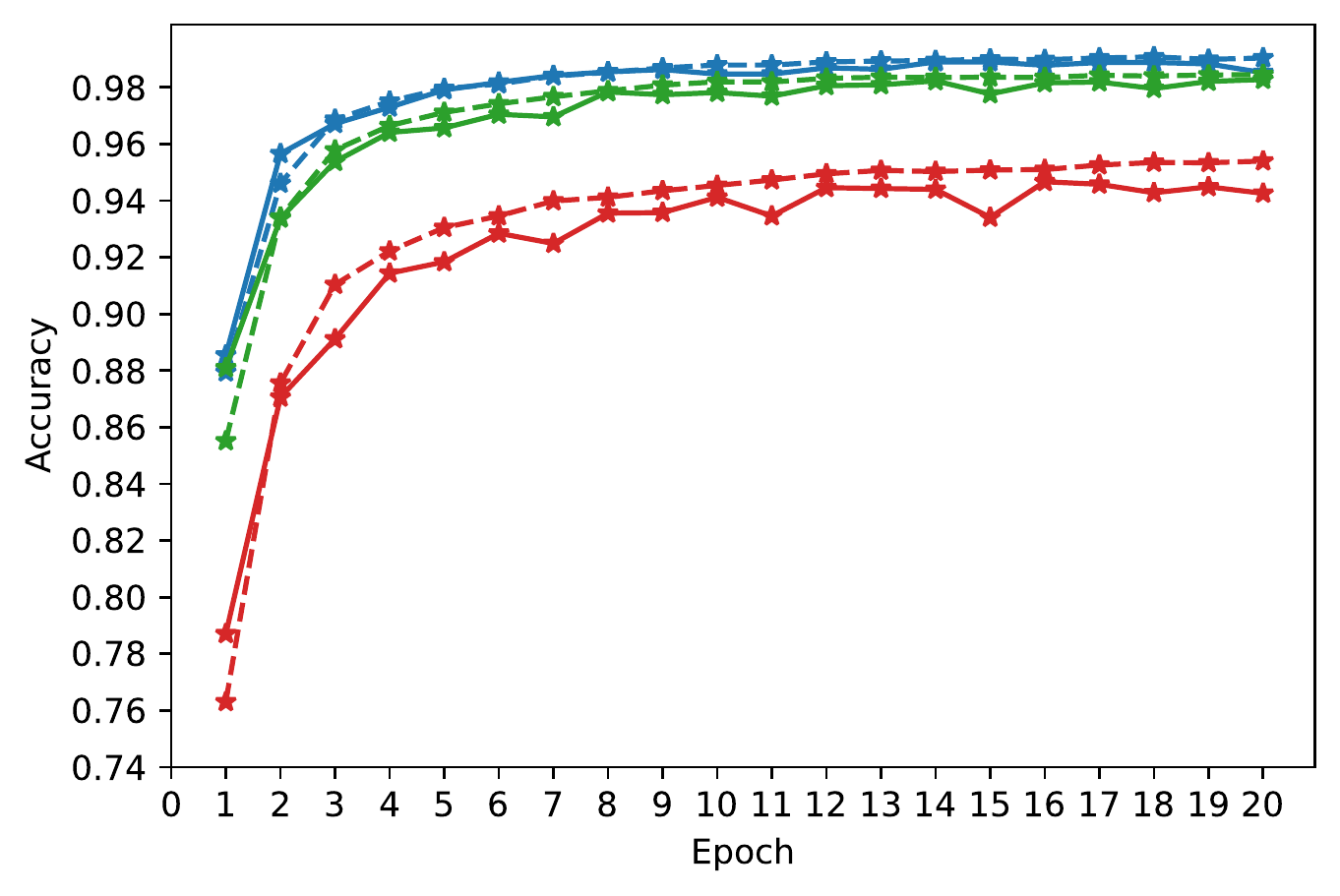}
\end{minipage}
}%
\hspace{-0.4cm}
\subfigure[0.02 lr (zoomed in)]{
\begin{minipage}[t]{0.25\linewidth}
\centering
\includegraphics[width=1.75in]{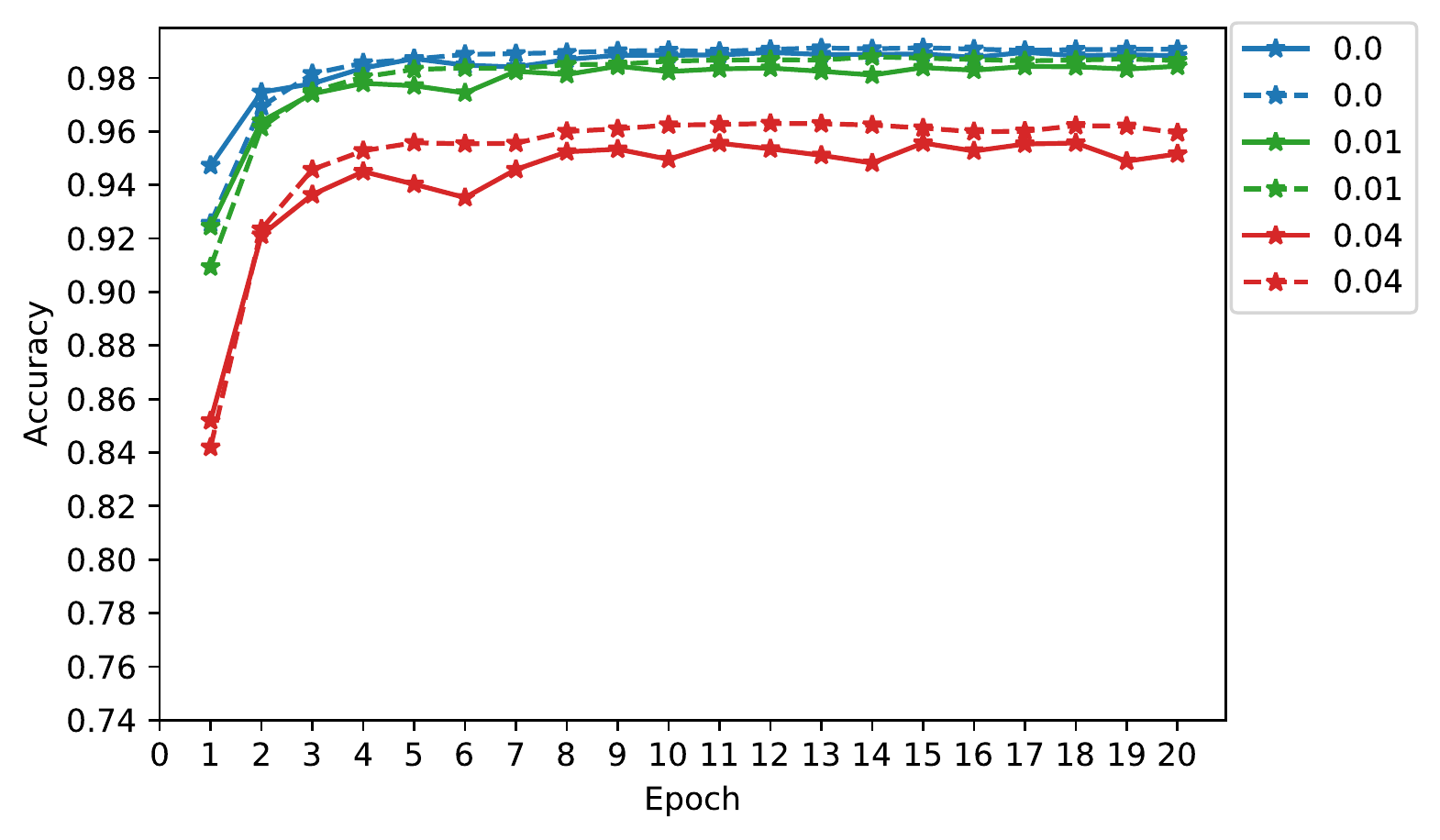}
\end{minipage}
}%
\centering
\caption{Adversarial accuracy of neural networks on MNIST under PGD, trained with different learning rates.}
\label{fig:MNIST PGD lr}
\end{figure}

From \Cref{fig:MNIST FGSM lr} and \Cref{fig:MNIST PGD lr} we can see that a larger learning rate corresponds to a higher adversarial accuracy and has a more significant effect on the performance of our snapshot ensemble especially. It is clear that a small magnitude of learning rate produces non-robust results, but an excessively large learning rate could have the same negative effects since the models may not converge. 

\subsubsection{Momentum coefficient}

\begin{figure}[!htb]
\centering
\subfigure[0.5 momentum]{
\begin{minipage}[t]{0.333\linewidth}
\centering
\includegraphics[width=2in]{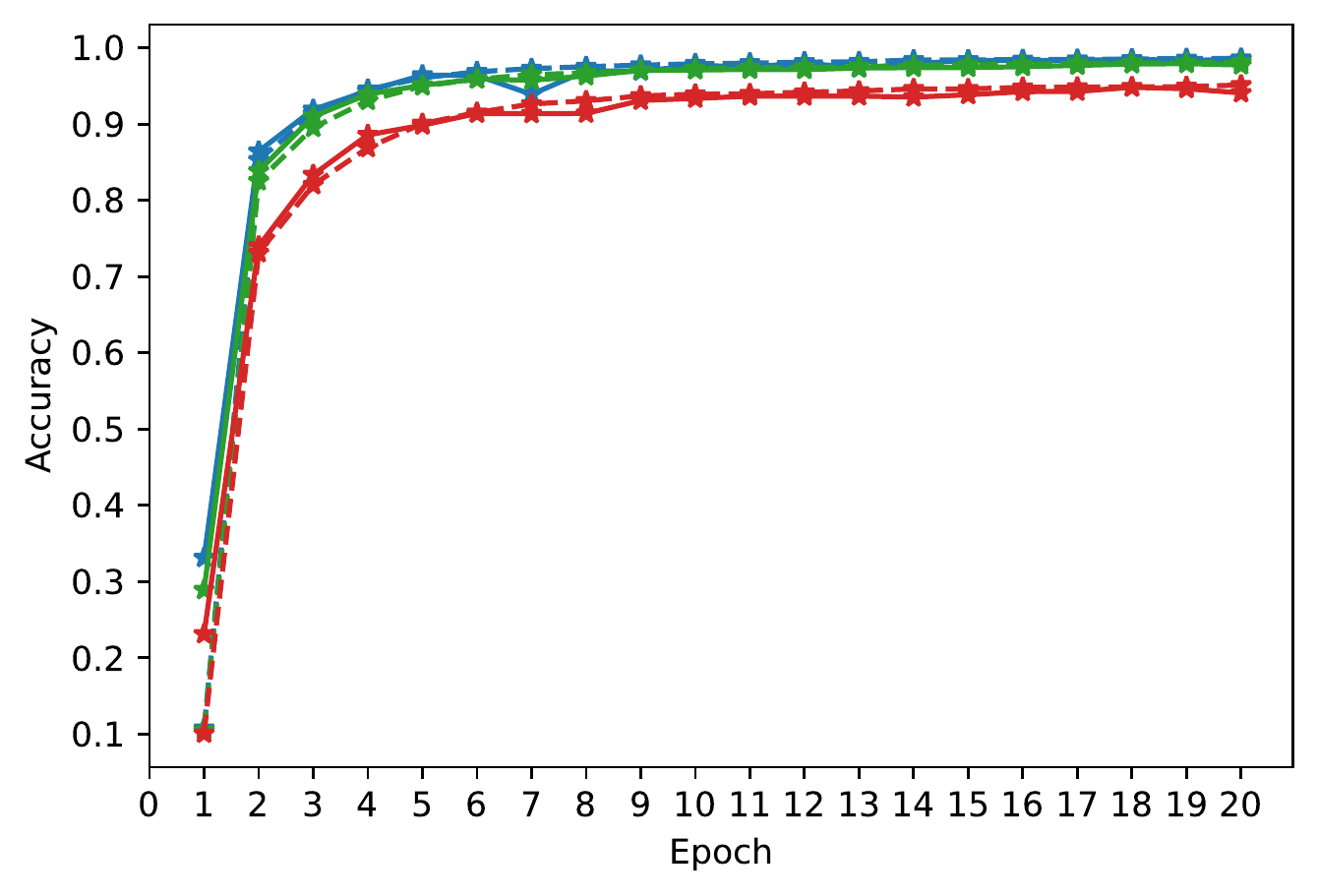}
\end{minipage}%
}%
\hspace{-0.6cm}
\subfigure[0.7 momentum]{
\begin{minipage}[t]{0.333\linewidth}
\centering
\includegraphics[width=2in]{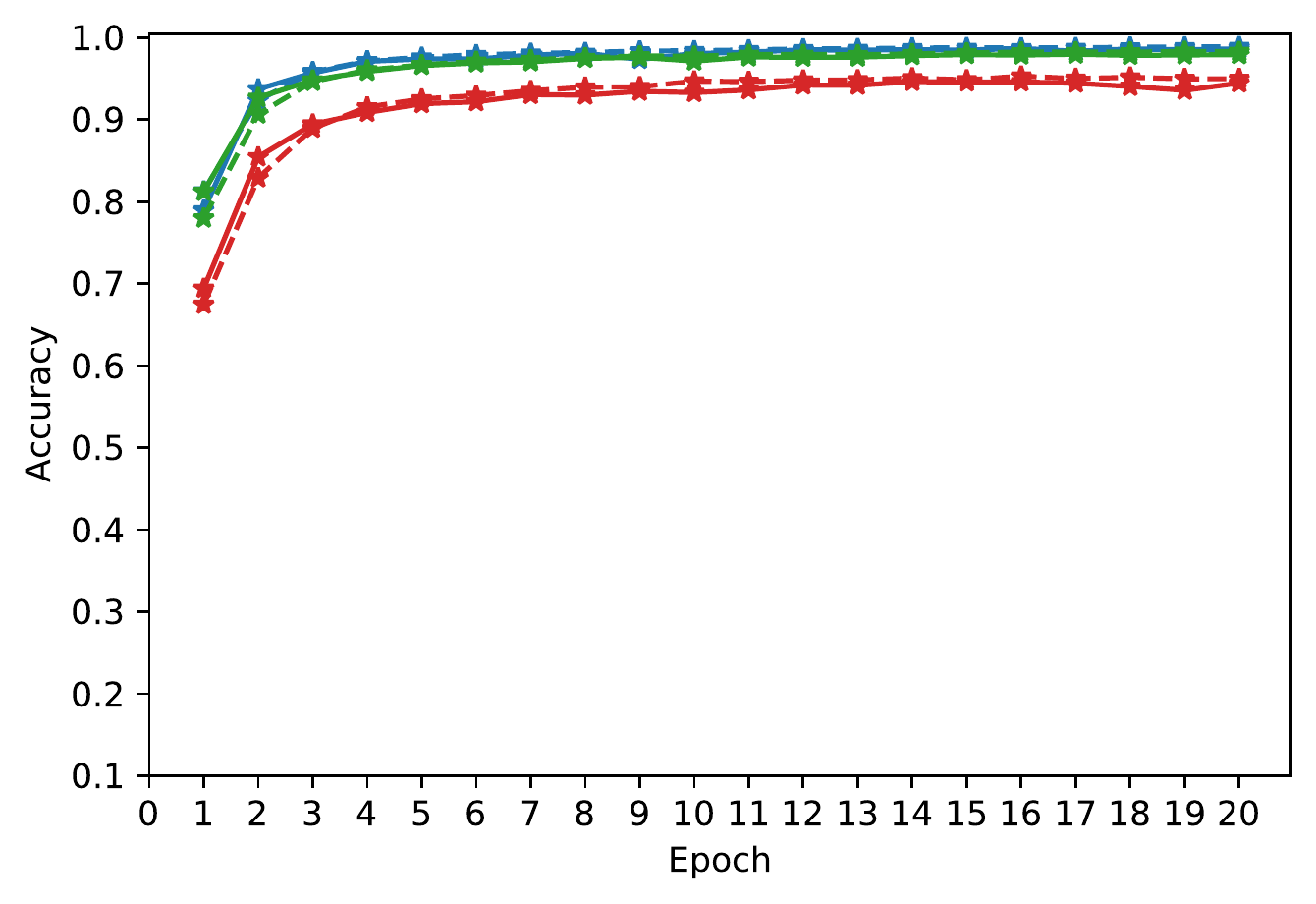}
\end{minipage}%
}%
\hspace{-0.4cm}
\subfigure[0.9 momentum]{
\begin{minipage}[t]{0.333\linewidth}
\centering
\includegraphics[width=2.3in]{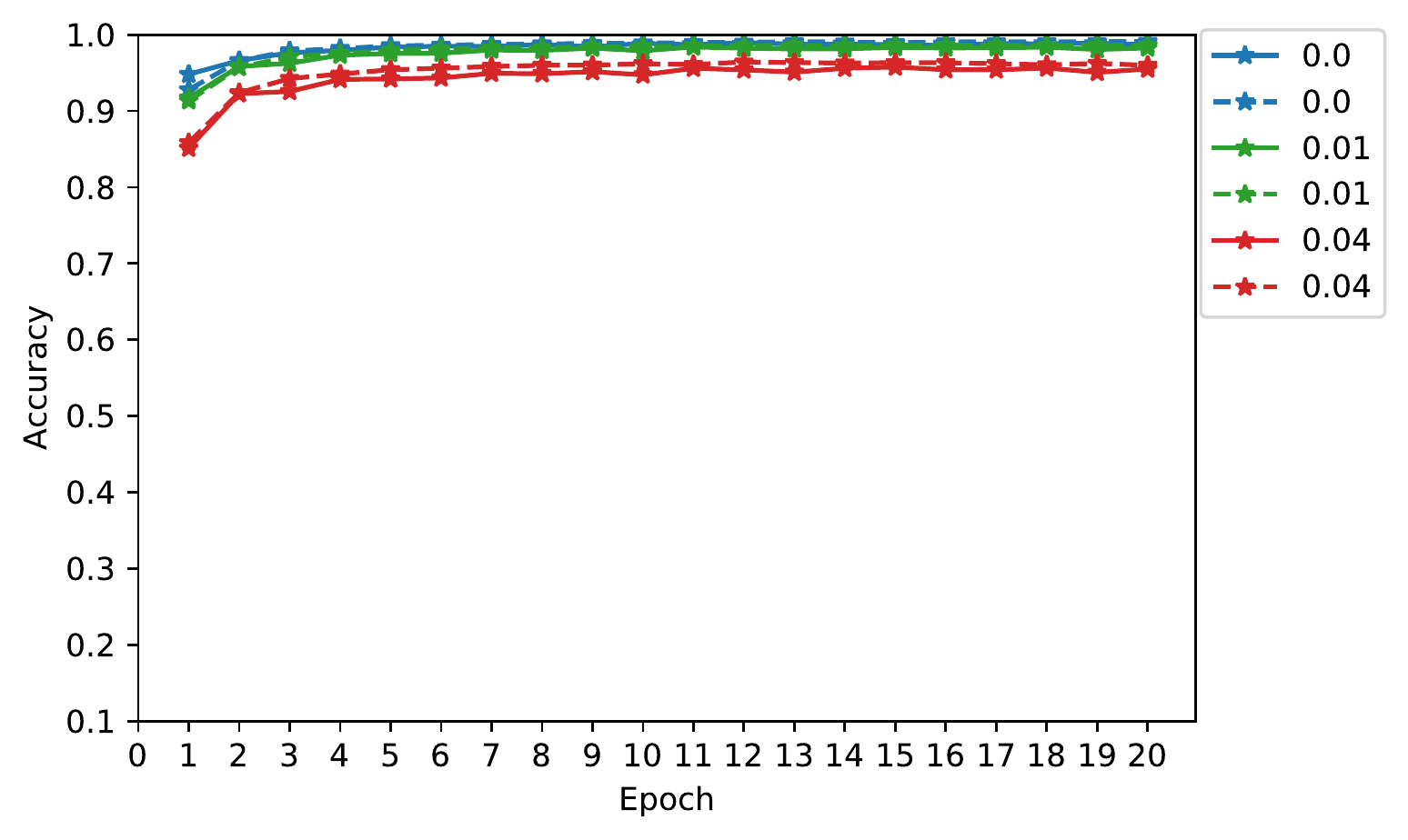}
\end{minipage}
}%

\subfigure[0.5 momentum (zoomed in)]{
\begin{minipage}[t]{0.333\linewidth}
\centering
\includegraphics[width=2in]{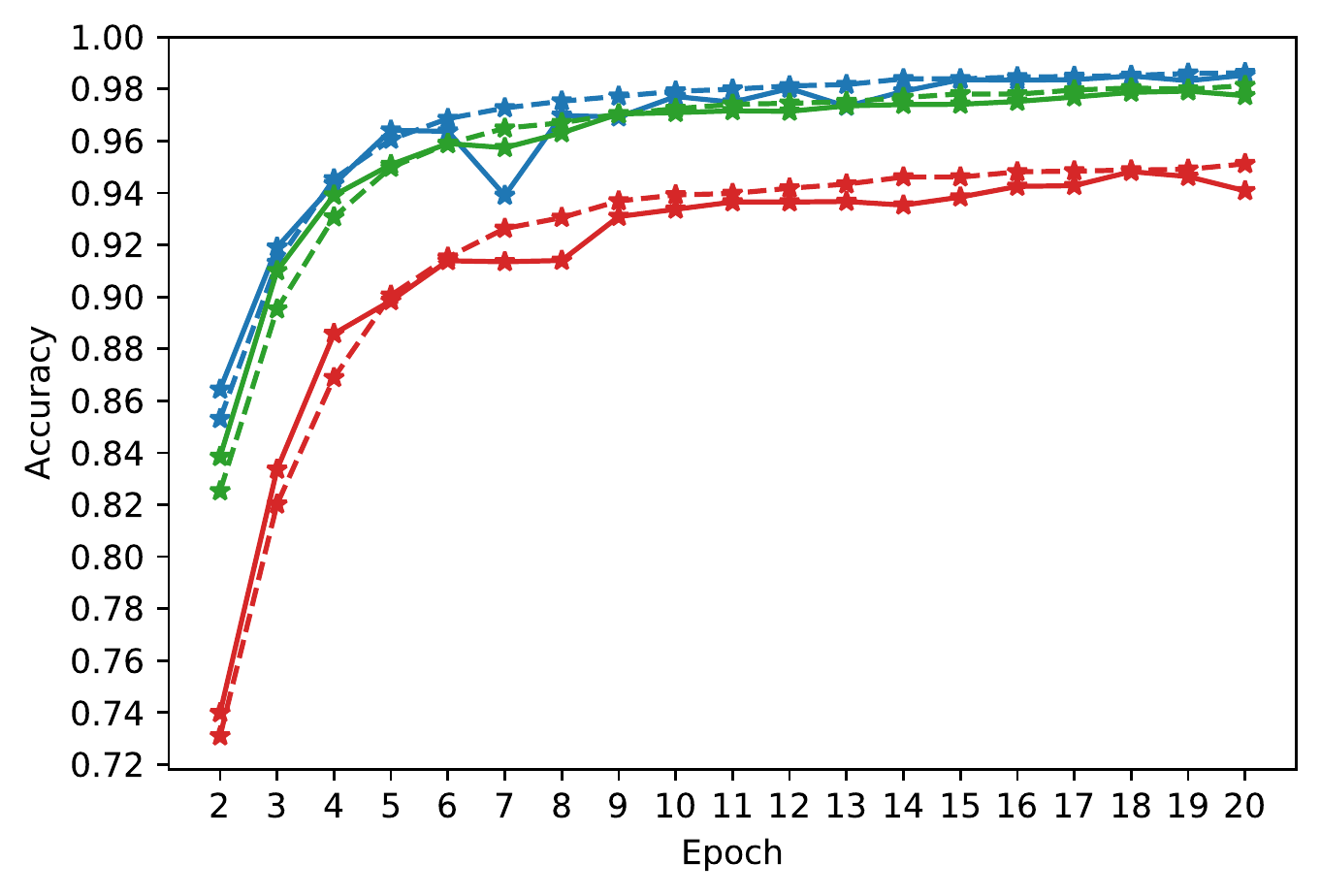}
\end{minipage}%
}%
\hspace{-0.6cm}
\subfigure[0.7 momentum (zoomed in)]{
\begin{minipage}[t]{0.333\linewidth}
\centering
\includegraphics[width=2in]{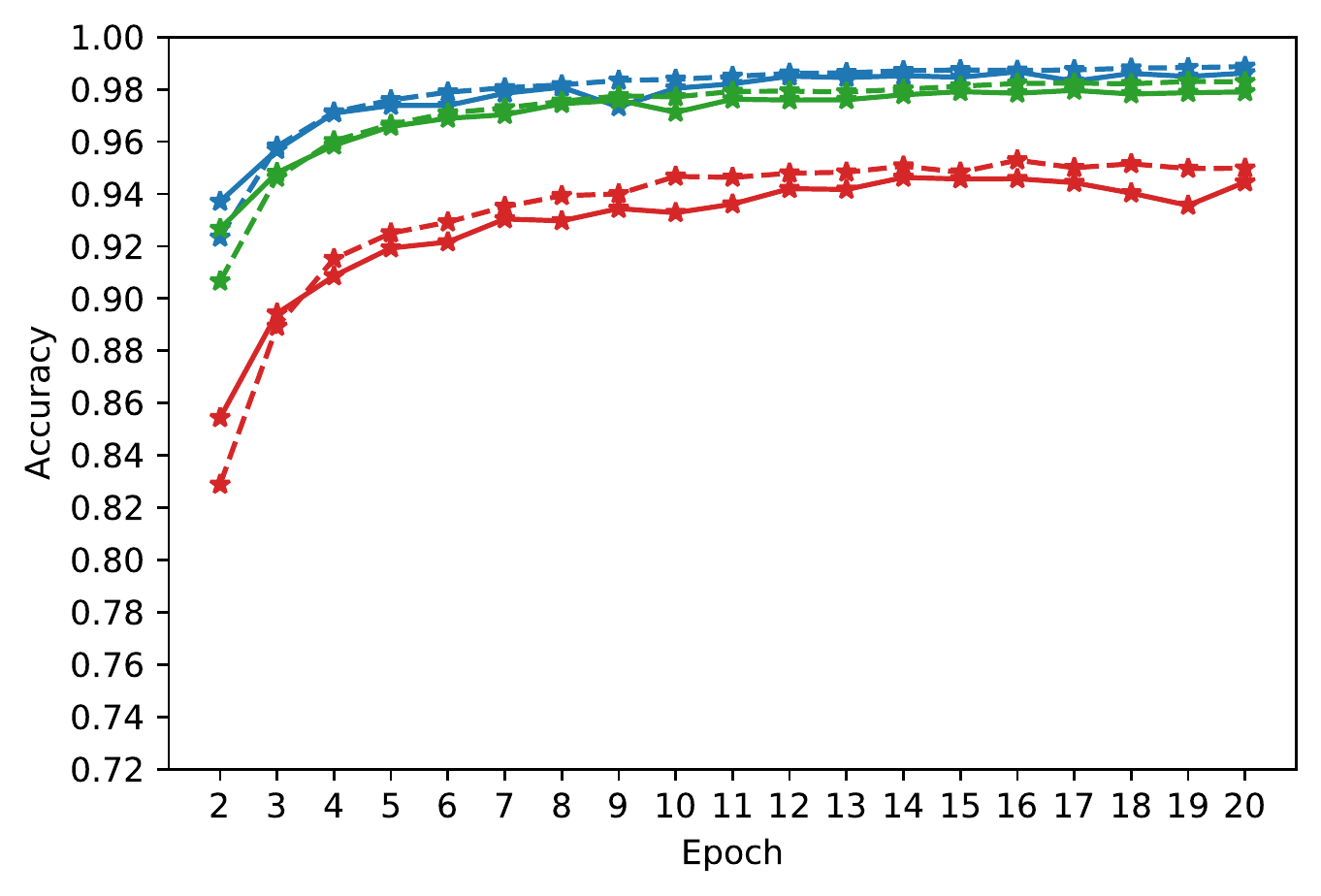}
\end{minipage}%
}%
\hspace{-0.4cm}
\subfigure[0.9 momentum (zoomed in)]{
\begin{minipage}[t]{0.333\linewidth}
\centering
\includegraphics[width=2.3in]{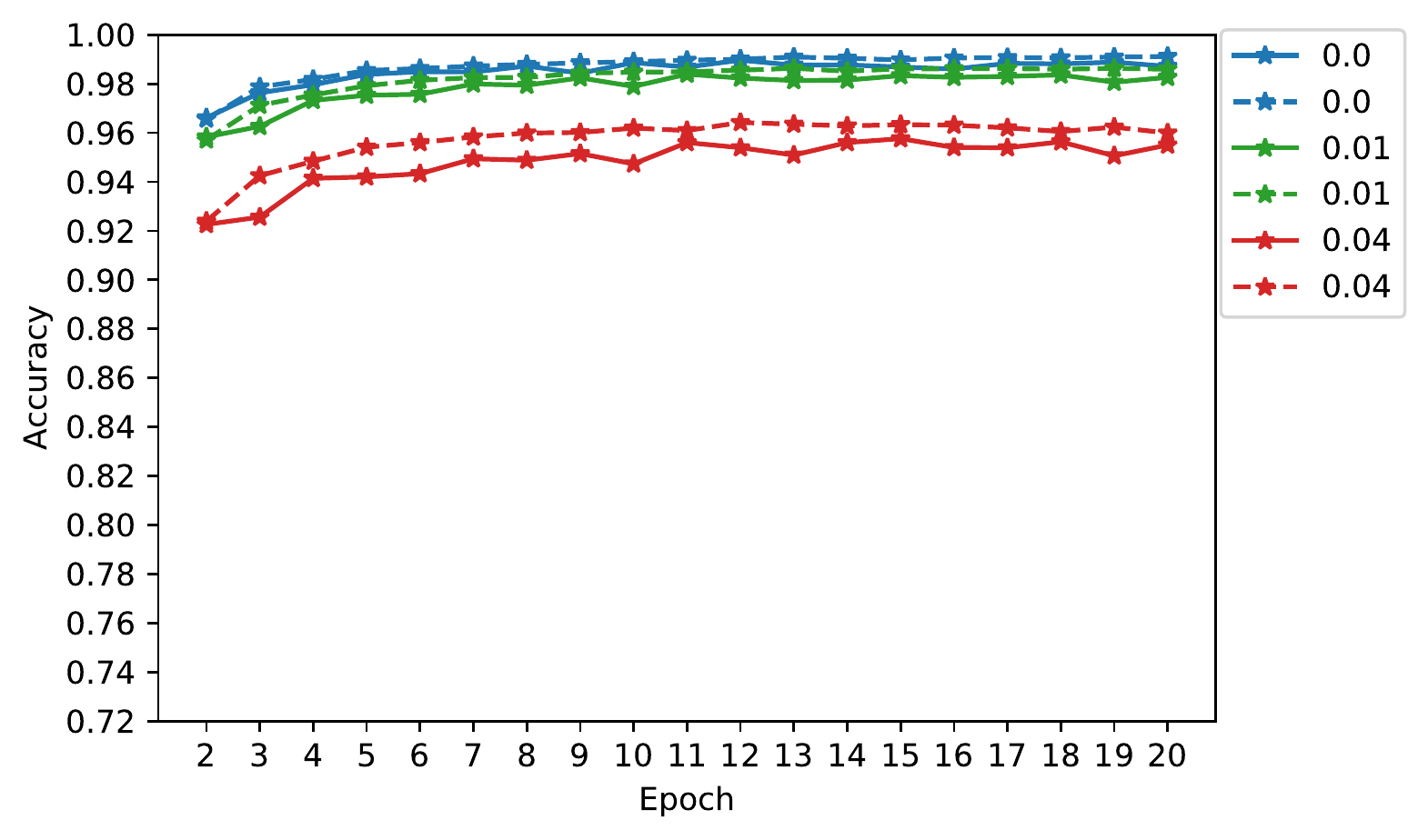}
\end{minipage}
}%
\centering
\caption{ Adversarial accuracy of neural networks on MNIST under FGSM, trained with different momentums.}
\label{fig:MNIST FGSM mm}
\end{figure}

\begin{figure}[!htb]
\centering
\subfigure[0.5 momentum]{
\begin{minipage}[t]{0.333\linewidth}
\centering
\includegraphics[width=2in]{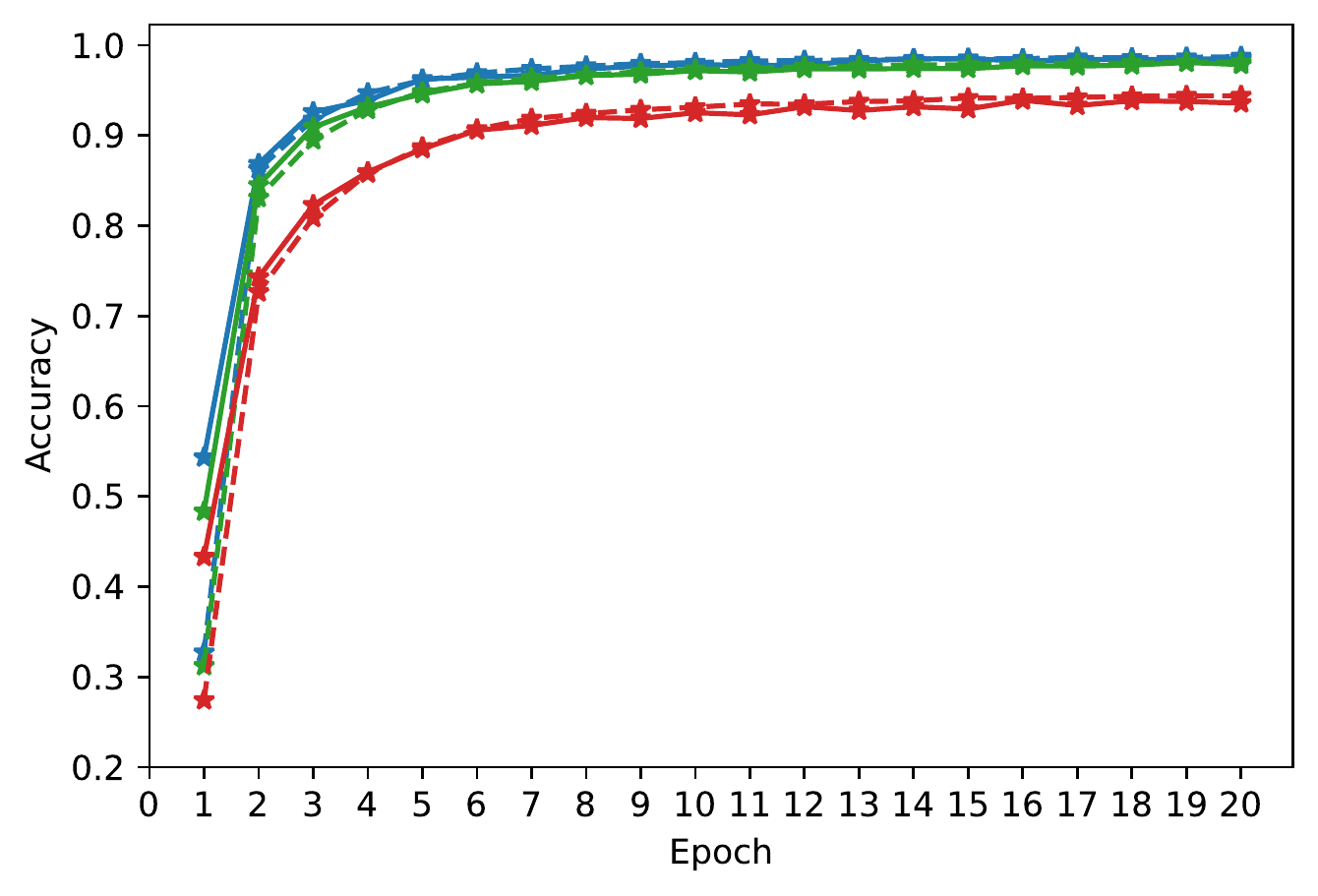}
\end{minipage}%
}%
\hspace{-0.6cm}
\subfigure[0.7 momentum]{
\begin{minipage}[t]{0.333\linewidth}
\centering
\includegraphics[width=2in]{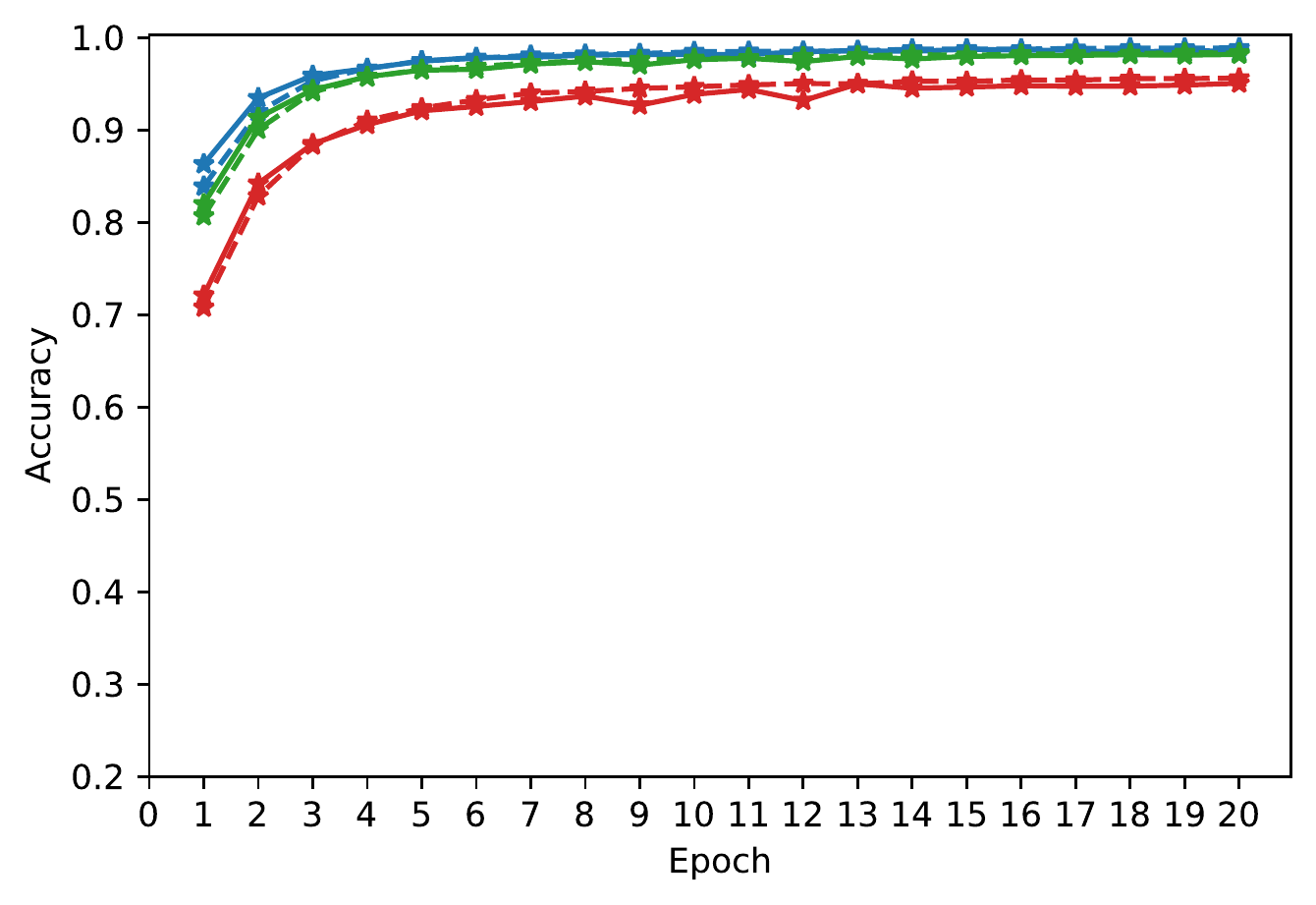}
\end{minipage}%
}%
\hspace{-0.4cm}
\subfigure[0.9 momentum]{
\begin{minipage}[t]{0.333\linewidth}
\centering
\includegraphics[width=2.3in]{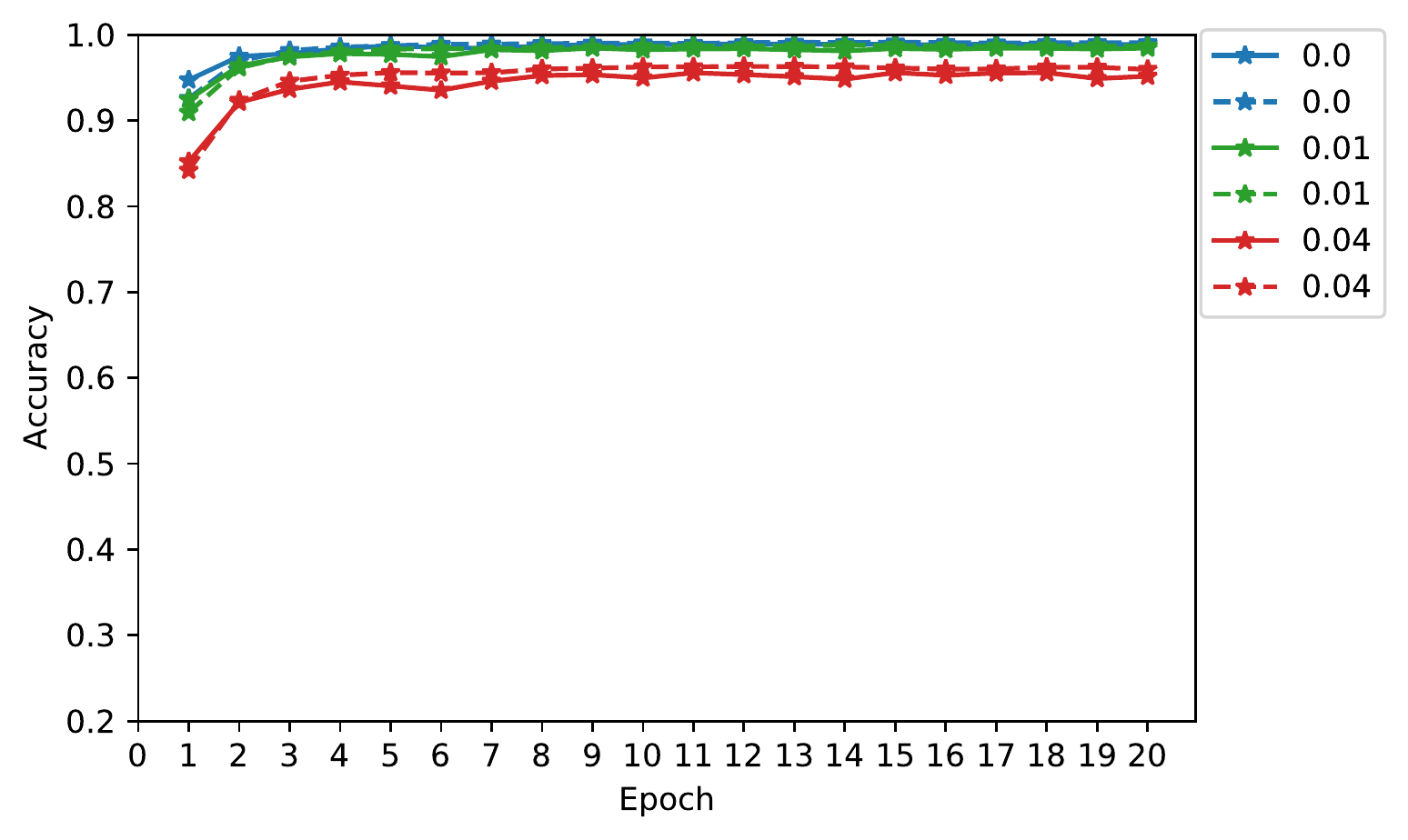}
\end{minipage}
}%

\subfigure[0.5 momentum (zoomed in)]{
\begin{minipage}[t]{0.333\linewidth}
\centering
\includegraphics[width=2in]{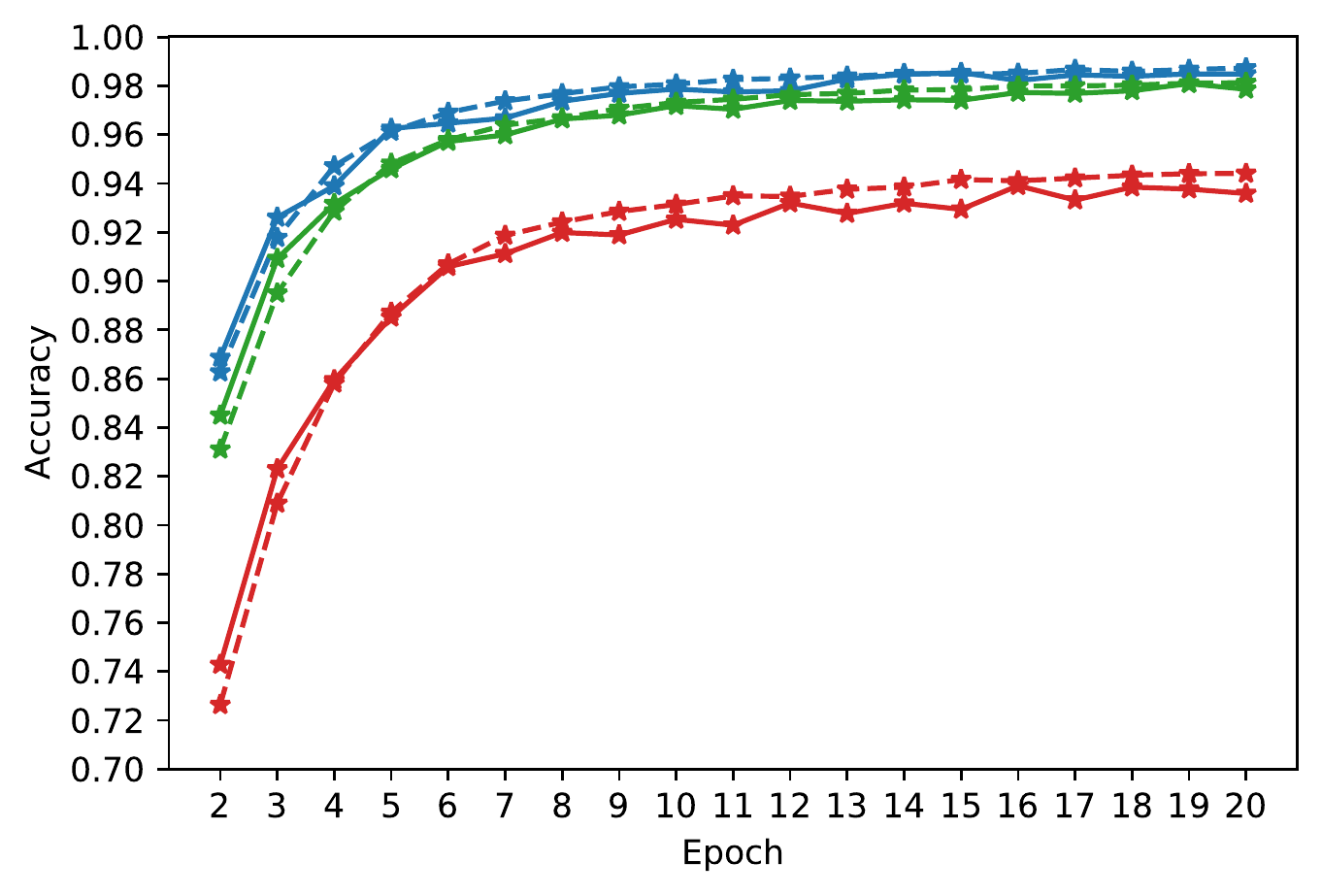}
\end{minipage}%
}%
\hspace{-0.6cm}
\subfigure[0.7 momentum (zoomed in)]{
\begin{minipage}[t]{0.333\linewidth}
\centering
\includegraphics[width=2in]{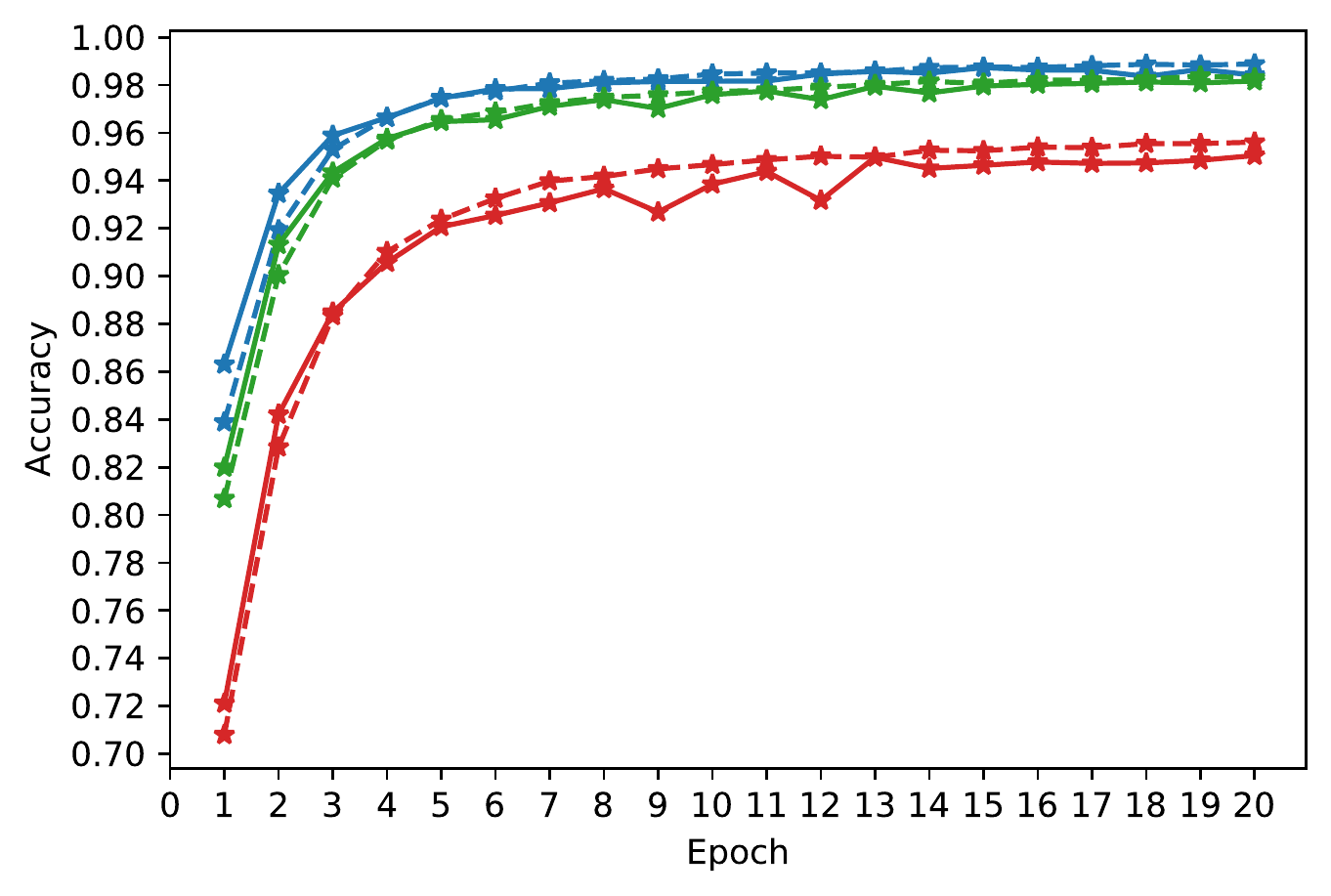}
\end{minipage}%
}%
\hspace{-0.4cm}
\subfigure[0.9 momentum (zoomed in)]{
\begin{minipage}[t]{0.333\linewidth}
\centering
\includegraphics[width=2.3in]{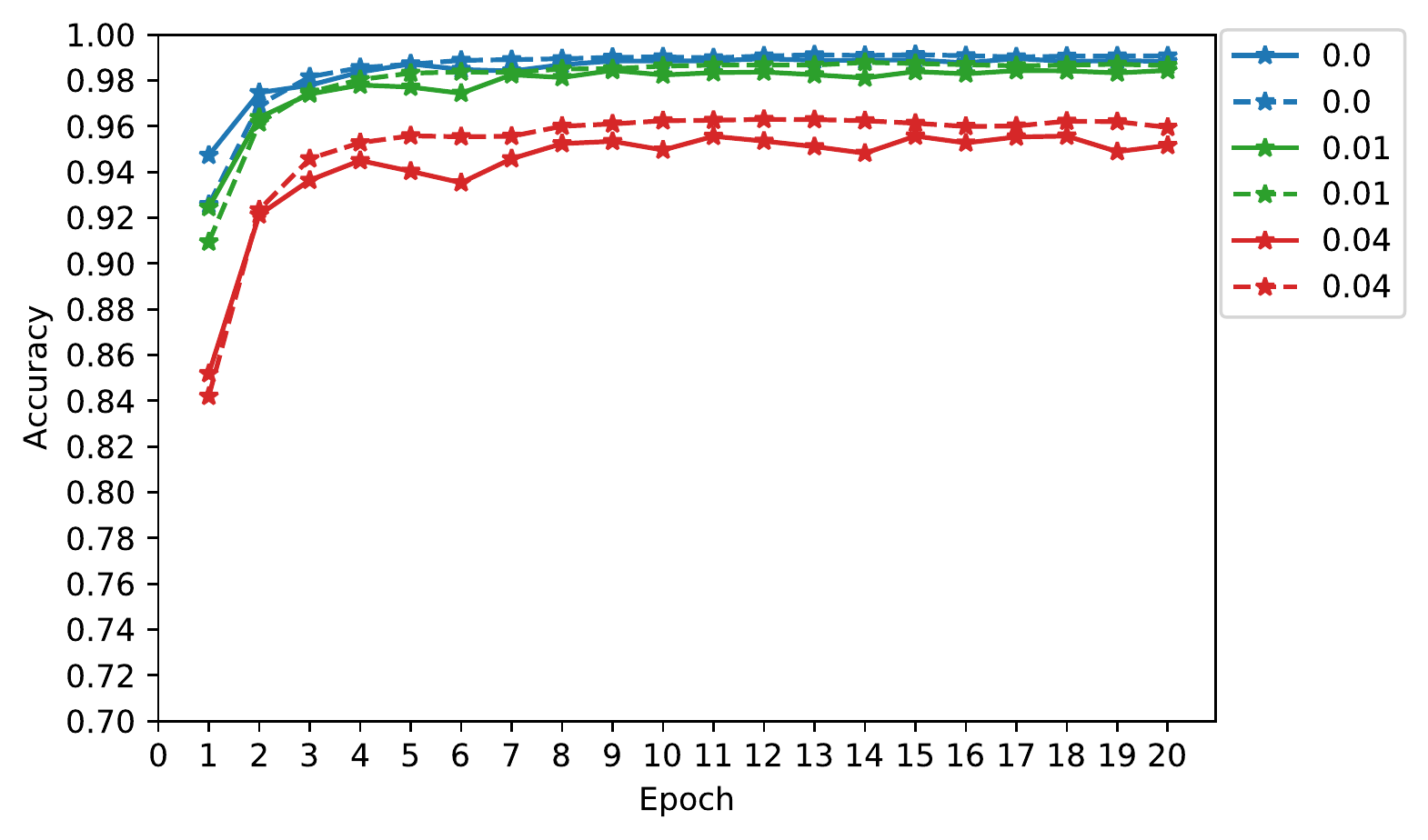}
\end{minipage}
}%
\centering
\caption{Adversarial accuracy of neural networks on MNIST under PGD, trained with different momentums.}
\label{fig:MNIST PGD mm}
\end{figure}

From \Cref{fig:MNIST FGSM mm} and \Cref{fig:MNIST PGD mm} we can observe the same phenomenon that a higher momentum corresponds to a higher adversarial accuracy and again that higher momentum with our snapshot ensemble improves the performance more significantly.

\end{document}